\documentclass[10pt,letterpaper]{article}

\usepackage[T1]{fontenc}
\usepackage[utf8]{inputenc}
\usepackage{times}
\usepackage{graphicx}
\usepackage{amsmath}
\usepackage{amssymb}
\usepackage{booktabs}
\usepackage{multirow}
\usepackage{array}
\usepackage{makecell}
\usepackage{multicol}
\newcommand{\tablefont}{\footnotesize}
\usepackage{url}
\usepackage{xurl}
\usepackage{microtype}
\usepackage[margin=1in]{geometry}
\usepackage{xcolor}
\usepackage[colorlinks=true,allcolors=blue]{hyperref}

\newcommand{\code}[1]{\texttt{#1}}
\makeatletter
\newcommand{\takeaway}[1]{\par\smallskip\noindent\colorbox{black!6}{\parbox{\dimexpr\linewidth-2\fboxsep\relax}{\textbf{Takeaway.} #1}}\par\medskip\@afterindentfalse\@afterheading}
\makeatother

\newcommand{\newtext}[1]{#1}
\newcommand{\repo}{\url{https://github.com/plant-ai-biophysics-lab/AGgentic-benchmarks}}
\newcommand{\agmlurl}{\url{https://project-agml.github.io/}}

\title{Open-weight vision-language models remain unreliable for agricultural image understanding across 372 datasets}

\author{Earl Ranario, Bharath Sridhar, Jared Smith,\\ Urmil Chandarana, Lars Lundqvist, J. Mason Earles\\[4pt]
University of California, Davis\\
{\small\texttt{\{ewranario, bsridhar, jrssmith, uchandar, llund, jmearles\}@ucdavis.edu}}}

\date{}

\begin{document}
\maketitle

\begin{abstract}
Vision-language models (VLMs) are increasingly proposed as general-purpose tools for agricultural diagnosis, but their reliability is poorly understood. We benchmark eight recent open-weight VLMs on 372 agricultural datasets from the AgML collection (\agmlurl), each model running on a single 48\,GB GPU. The datasets comprise 220 image classification datasets (1,832 classes), 105 object detection, 34 semantic segmentation and 13 visual question answering (VQA) datasets, or 32,456 items per model. Every classification image is asked in three formats, as multiple choice with distractors from the same dataset (MCQA within) or from other datasets (MCQA across), and as an open-ended question (OEQ) graded by an LLM judge. Five findings stand out. (1) No model is reliable, and none is ready to replace a supervised model. The best classification accuracy is 54.9\% (MCQA across), open-ended accuracy peaks at 35.6\%, detection and segmentation reach at most 24.7\% and 30.6\%, and no model leads every task. A supervised YOLO11 model trained on only 10 labelled images per class, which sees only the image, outperforms the best VLM on every task it covers, with 77.3\% top-1 classification accuracy and 37.5\% and 51.4\% on detection and segmentation. (2) The question format changes the result. Asking the same images open-ended instead of as multiple choice lowers accuracy by 12.0--27.7 percentage points. Drawing distractors from the same dataset instead of other datasets lowers it by 3.2--7.6 points for seven of eight models. Even which agricultural task looks hardest depends on the format. Because users naturally ask open-ended questions, multiple-choice scores overstate how models perform in practice. (3) Models rarely notice when the correct answer is missing. With plausible options from the same dataset (MCQA within), most models pick a wrong option more often than they answer ``None of the above''. Implausible options from other datasets (MCQA across) make them reject the list more often, but whether or not the answer is in it. (4) Benchmark design choices are not neutral. Multiple-choice accuracy averages two scores, the accuracy on questions whose correct class is listed and the accuracy on questions whose correct class was removed, and the share of removed-class questions sets the weights. Changing that share from 30\% to 10\% or 50\% changes a model's accuracy by up to 7.7 points and reorders the top models, although no model answers differently. Fuzzy string matching also scores the same open-ended answers 4.7--17.4 points higher than the judge. (5) Aggregate accuracy is reproducible, but individual answers are not. A model's accuracy varies by a standard deviation of at most 1.1 points across three generation seeds and 0.6 points across five option orders. Current open-weight VLMs are therefore not yet suitable as standalone agricultural diagnostic systems. Multiple-choice benchmark scores should be treated with caution and reported together with their answer format, option-set design and scoring. Code, prompts and results are available at \repo.
\end{abstract}

\section{Introduction}
\label{sec:intro}

Agriculture must adapt to climate extremes with limited technological access and little publicly available, standardized data \cite{aijaz_artificial_2025}, which calls for climate-resilient and nutrient-stable crops \cite{berlingeri_integration_2025}. Identifying the genetic and environmental drivers of disease or yield requires phenotypic data from large experiments, and collecting it remains manual and costly \cite{reynolds_what_2019}. Vision models trained for one location, however, often fail when deployed elsewhere because of domain shift \cite{hu_domain_2025}. Vision-language models (VLMs) offer a different route. A single model takes an image and a question and returns a candidate diagnosis with an explanation, and it can be redirected to a new task by changing the prompt instead of retraining. Used this way, a VLM acts as a ``virtual agronomist'' that could bring expert knowledge to growers and researchers alike.

Agricultural images are, however, harder than those in general-purpose benchmarks \cite{huang_evaluating_2025}. Disease diagnosis, pest and damage identification and fine-grained species recognition depend on subtle visual cues, high intra-class variability and context such as crop, growth stage, geography and management history. These tasks have historically been solved by supervised, task-specific models \cite{kamilaris_deep_2018}, which need substantial annotation and often fail to generalize across crops, regions and imaging conditions \cite{magistri_one_2023, zhang_unsupervised_2022, yang_enhancing_2025}. Agricultural benchmarks have begun to test VLMs and report large gaps to domain requirements \cite{joshi_standardizing_2023, arshad_leveraging_2025, shinoda_agrobench_2025}. Their results, however, vary with the evaluation protocol, prompt format and scoring, so it is hard to tell how ready these models are or how they compare.

The question that motivates this study, posed for classification in our earlier work \cite{ranario_are_2025}, is therefore whether current VLMs are ready to replace supervised, task-specific models for agricultural perception. A VLM needs no labelled images, so the fair point of comparison is not a supervised model trained on a full dataset but one trained on the few labels a practitioner could collect quickly. We therefore benchmark eight recent open-weight VLMs on 372 agricultural datasets from the AgML collection\footnote{\agmlurl} \cite{joshi_standardizing_2023}, covering image classification (220 datasets, 1,832 classes), object detection (105), semantic segmentation (34) and visual question answering (VQA, 13). Every model runs locally on a single 48\,GB GPU with the same prompts, items and scoring. Every classification image is asked in three formats, as a multiple-choice question (MCQA) with distractors from the same dataset (MCQA within) or from other datasets (MCQA across), and as an open-ended question (OEQ). We repeat the benchmark with three generation seeds and run controlled experiments on the multiple-choice option set. This separates differences between models from run-to-run noise and from artifacts of the prompt design. To answer the question directly, we also train a supervised YOLO11 model on only 10 labelled images per class and score it on the same images, with 5 and 15 images per class in the appendix. Five findings stand out.

\begin{enumerate}
    \item No open-weight VLM is reliable for agricultural perception, and none is ready to replace a supervised model (Section~\ref{sec:results_overall}). The best classification accuracy is 54.9\%, open-ended accuracy is at most 35.6\%, and detection and segmentation accuracy are at most 24.7\% and 30.6\%. A supervised YOLO11 model trained on only 10 images per class reaches 77.3\% top-1 classification accuracy, above every VLM in every question format and agricultural task, and 37.5\% and 51.4\% on detection and segmentation. No model is best at everything either, and the model that is best at open-ended answers is among the worst at detection.
    \item How the question is asked changes accuracy and which task looks hard (Section~\ref{sec:results_format}). Accuracy falls by 12.0--27.7 percentage points when the same images are asked open-ended instead of as multiple choice, and by 3.2--7.6 points when the distractors come from the same dataset instead of other datasets. Species identification and quality grading swap places between the easiest and hardest task when the format changes. Because users naturally ask open-ended questions, multiple-choice scores overstate how models perform in practice.
    \item Models rarely recognize that the correct answer is missing (Section~\ref{sec:results_nota}). When the true class is withheld from plausible options (MCQA within), most models pick a wrong option more often than they answer ``None of the above''. Implausible options (MCQA across) make models reject the list more often, but they do so whether or not the answer is in it, so they do not help models notice what is missing.
    \item Benchmark design and scoring choices shift accuracy and reorder models (Section~\ref{sec:results_design}). Multiple-choice accuracy averages two scores, the accuracy on questions whose true class is listed and the accuracy on questions whose true class was removed, where the correct answer is ``None of the above''. The share of removed-class questions sets the weights. Moving that share from the default 30\% to 10\% or 50\% changes a model's accuracy by up to 7.7 points and reorders the top models, even though no model answers differently. Removing the ``I don't know'' option changes accuracy by up to 2.7 points, and fuzzy string matching scores the same open-ended answers 4.7--17.4 points higher than a large language model (LLM) used as a judge.
    \item Aggregate scores are reproducible, but individual answers are not (Section~\ref{sec:results_stability}). The standard deviation of a model's accuracy is at most 1.1 points across three generation seeds and 0.6 points across five option orders, yet only 27--64\% of multiple-choice items receive the same answer in all five option orders.
\end{enumerate}

The answer to our question is therefore no. These findings suggest that VLMs are not yet drop-in replacements for supervised models, even small ones trained on a handful of labels, and that multiple-choice benchmarks overstate how they perform when asked open-ended questions, as users naturally do. Benchmarks that do use multiple choice need to control and report how the options are built. To support such work, we release the benchmark code, the exact sampled items, the prompts, the judge and all reported scores and statistics at \repo, and Appendix~\ref{app:repro} gives the command that produces every table and figure.

\section{Background}
\label{sec:background}

\subsection{From contrastive pretraining to instruction-tuned VLMs}
\label{sec:background_history}
Modern VLMs combine two lines of work. The first is foundation models, which are pretrained on broad, largely uncurated data and then adapted to many downstream tasks rather than trained from scratch for each one \cite{bommasani_opportunities_2021}. Contrastive image-text models such as CLIP and SigLIP learn a shared embedding space for images and text from hundreds of millions of web image-caption pairs \cite{radford_learning_2021, zhai_sigmoid_2023}. An image can then be matched to the class whose description it resembles most, without labeled examples of that class. This enables zero-shot classification, but these models are discriminative, scoring a fixed set of labels without generating text or explaining a prediction.

The second line is instruction-tuned large language models (LLMs). GPT-3 showed that a model trained on next-token prediction at scale can solve unseen tasks from a natural-language description and a few examples in the prompt, without gradient updates \cite{brown_language_2020}. LLaVA extended this to images by connecting a pretrained vision encoder to an LLM through a projection layer and instruction-tuning the combination on multimodal conversations, so that it can answer open-ended questions about an image \cite{liu_visual_2023}. LLaVA-NeXT adds higher-resolution and interleaved image-text inputs \cite{li_llava-next-interleave_2024}, and Qwen-VL and its successors scale the recipe with stronger grounding and multi-image reasoning \cite{bai_qwen-vl_2023, bai_qwen3-vl_2025}. All eight models evaluated here belong to this second family, and most can produce a reasoning trace before answering. Our three prompt types retrace this history. Multiple-choice prompts sit close to the discriminative, CLIP-style setting, open-ended prompts test free-form generation, and grounding prompts ask the model to localize objects. Section~\ref{sec:results} shows that a model's ranking in one regime does not predict its ranking in another.

\subsection{Zero-shot prompting and option-set design}
\label{sec:background_prompting}
Zero-shot learning means performing a task or recognizing a class without labeled training examples of it, typically by relating it to knowledge acquired in pretraining such as a class name \cite{xian_zero-shot_2019}. In VLMs it is done either by listing candidate classes in the prompt or by asking an open-ended question and mapping the answer back to a label. Few-shot in-context learning adds labeled examples to the prompt without changing the model's parameters \cite{brown_language_2020, gu_systematic_2023}, and prompt engineering more broadly designs the instruction, format and content of the input to steer a frozen model \cite{sahoo_systematic_2024, schulhoff_prompt_2024}. All evaluations in this paper are zero-shot.

Open-ended prompting places no constraint on the answer. It is closest to how assistants are used, but a free-form answer must be parsed into a label, and a correct answer phrased differently from the reference can be missed. Constrained prompting restricts the model to a listed set of candidate classes, a label ontology, which removes most of this ambiguity but introduces new design choices, chief among them which incorrect options (distractors) to show. LLMs answering multiple-choice questions are sensitive to the composition and order of the options, not only to the content of the question \cite{pezeshkpour_large_2023}, so distractor choice can raise or lower accuracy independently of a model's ability. General-purpose benchmarks such as MMLU made multiple-choice evaluation a standard \cite{hendrycks_measuring_2021}, but usually treat the option set as fixed. We instead treat the distractor source, the option order, the rate of ``None of the above'' answers and the presence of an ``I don't know'' option as experimental variables.

\subsection{Agricultural benchmarks for vision and VLMs}
\label{sec:background_ag_benchmarks}
Agricultural images combine high intra-class variability, fine-grained differences between classes and sensitivity to growth stage, lighting and geography \cite{min_computer_2025, upadhyay_deep_2025}. iNatAg curates millions of crop and weed images for fine-grained species recognition at several taxonomic levels \cite{jain_inatag_nodate}, and AgML standardizes datasets for disease, pest, abiotic stress and other tasks across many crops \cite{joshi_standardizing_2023}. For VLMs, AgEval defines 12 plant stress phenotyping tasks and finds that few-shot prompting often helps but performance varies widely across stress types \cite{arshad_leveraging_2025}. AgroBench provides expert-annotated questions over hundreds of crop and disease categories and finds that VLMs still struggle with fine-grained tasks such as weed and disease identification \cite{shinoda_agrobench_2025}. Relative to these efforts, our benchmark (i) covers classification, detection, segmentation and VQA under one protocol; (ii) repeats every evaluation with independent generation seeds and reports dataset-bootstrap confidence intervals and paired tests; and (iii) varies the multiple-choice option set as an experimental factor.

\section{Experimental Setup}
\label{sec:experiment}

\subsection{Models}
\label{sec:models}

We benchmarked eight open-weight VLMs from Hugging Face \cite{wolf_huggingfaces_2020} (Table~\ref{tab:models}). Seven were served with vLLM \cite{kwon2023efficient}. NVIDIA-Nemotron-3.5-Bolt was served with llama.cpp\footnote{\url{https://github.com/ggml-org/llama.cpp}} from its 24\,GB GGUF release (llama.cpp's quantized model format), because its vLLM checkpoint of the same size requires a newer GPU architecture. Every run used one NVIDIA RTX 6000 Ada GPU (48\,GB). Sampling parameters follow each model's card. Every model reasons at a low effort setting before answering, and the reasoning trace is removed before scoring. Serving and reasoning settings are detailed in Appendix~\ref{app:inference}.

\begin{table}[!tbp]
\centering
\caption{Models evaluated. The number in each VLM name is the total parameter count; ``A3B'' and ``A9B'' mark mixture-of-experts models with about 3B and 9B active parameters. All VLMs were run with their reasoning trace at low effort (Appendix~\ref{app:inference}). The supervised baseline, YOLO11s, is one model per task, initialized from the Ultralytics pretrained weights and trained on 10 images per class (Section~\ref{sec:baselines}). It has no reasoning trace. $^\dagger$Gemma-4-31B is also the LLM judge, so its judged scores are self-judged.}
\label{tab:models}
\renewcommand{\arraystretch}{1.15}
\tablefont
\setlength{\tabcolsep}{5pt}
\begin{tabular}{llll}
\toprule
\textbf{Model} & \textbf{Repository (Hugging Face unless noted)} & \textbf{Weights} & \makecell[l]{\textbf{Reasoning}\\\textbf{(low effort)}} \\
\midrule
Qwen3.6-35B-A3B & \code{Qwen/Qwen3.6-35B-A3B-FP8} & FP8 & 1,024-token cap \\
Qwen3.8-27B & \code{Qwen/Qwen3.8-27B-FP8} & FP8 & native low \\
Gemma-4-31B$^\dagger$ & \code{RedHatAI/gemma-4-31B-it-FP8-dynamic} & FP8 & 1,024-token cap \\
Gemma-4-12B & \code{google/gemma-4-12B-it} & BF16 & 1,024-token cap \\
Muse-Glimmer-30B & \code{RedHatAI/Muse-Glimmer-30B-FP8-block} & FP8 & native low \\
Nemotron-3.5-Bolt-74B-A9B & \code{nvidia/NVIDIA-Nemotron-3.5-Bolt-74B-A9B-24GB-GGUF} & GGUF & native medium \\
MiniCPM-V-4.7-35B-A3B & \code{openbmb/MiniCPM-V-4.7-35B-A3B-FP8} & FP8 & 1,024-token cap \\
Ministral-3-14B & \code{mistralai/Ministral-3-14B-Instruct-2512} & FP8 & prompted (\code{<think>}) \\
\midrule
\makecell[l]{YOLO11s (supervised\\baseline, one per task)} & \makecell[l]{Ultralytics \code{yolo11s-cls.pt}, \code{yolo11s.pt},\\\code{yolo11s-seg.pt}, fine-tuned} & FP32 & --- \\
\bottomrule
\end{tabular}
\end{table}

\subsection{Datasets}
\label{sec:data}
\begin{figure*}[!tp]
    \centering
    \includegraphics[width=0.88\textwidth]{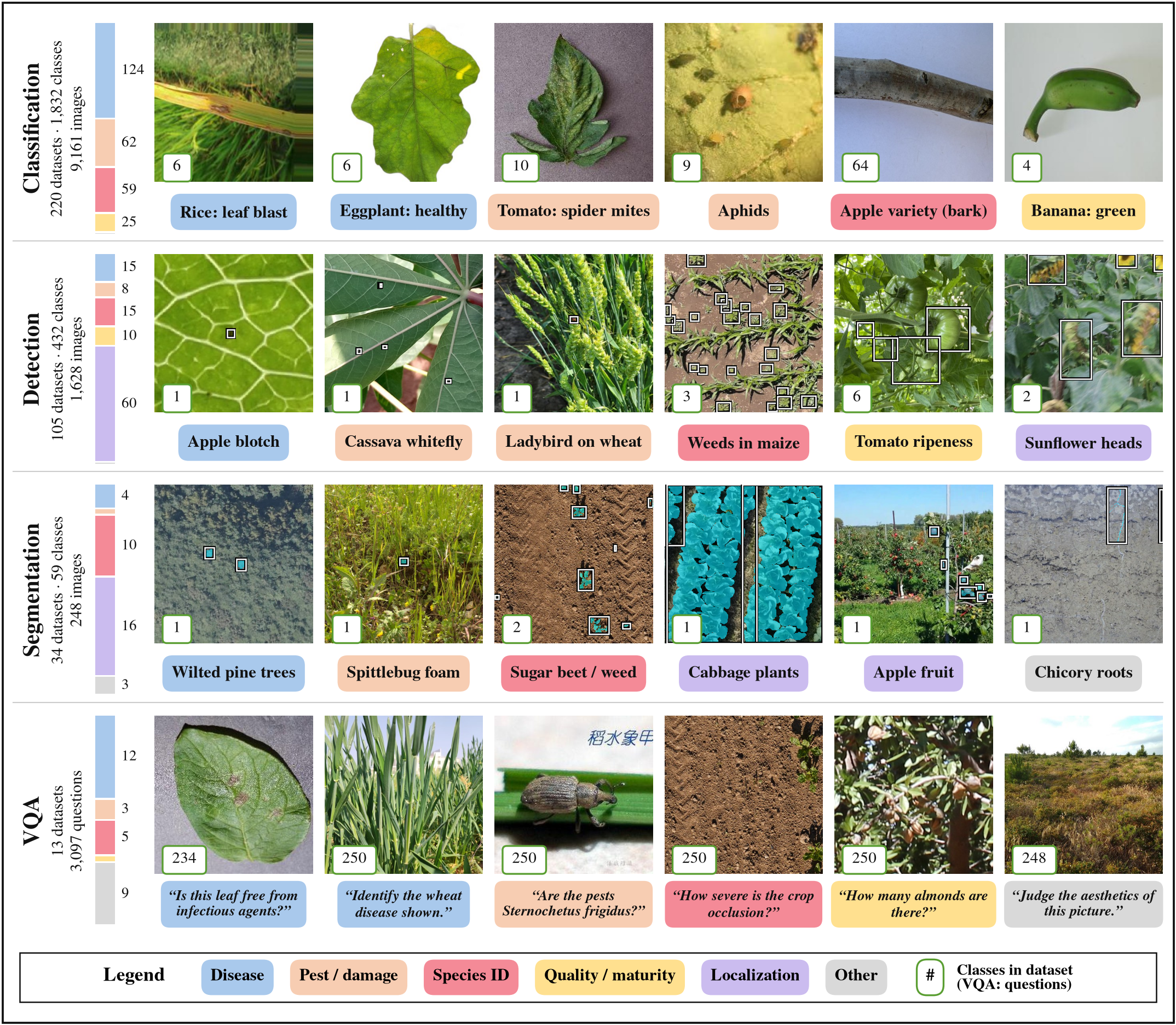}
    \caption{Benchmark overview. Each row is one task, with its dataset, class and image counts and a vertical bar giving the number of datasets per agricultural task (a dataset whose classes span several tasks counts in each). The six example images come from different datasets; the label colour gives the agricultural task and the badge the number of classes in that dataset (for VQA, the number of questions). Detection and segmentation examples show the ground-truth boxes, and segmentation examples also show the ground-truth mask (cyan), which is scored through its boxes. VQA labels show an abridged question.}
    \label{fig:dataset_overview}
\end{figure*}

We use every eligible RGB dataset in the \href{https://project-agml.github.io/}{AgML catalog} \cite{joshi_standardizing_2023} published for image classification, object detection, semantic segmentation or image-text-to-text (VQA), 372 datasets in total (Table~\ref{tab:data}, Figure~\ref{fig:dataset_overview}). Appendix~\ref{app:datasets} lists every dataset with its class and image counts. Each dataset is subsampled with a fixed seed (Appendix~\ref{app:sampling}). Classification keeps up to five images per class. Detection and segmentation keep enough images for every class to appear at least five times. VQA keeps up to 250 questions per dataset, stratified by the dataset's own question types and by answer format. This gives 14,629 images and 32,456 evaluation items per model. The VLMs are evaluated zero-shot and never fine-tuned. Only the supervised baseline (Section~\ref{sec:baselines}) is trained, on separately drawn images that never overlap the evaluated ones.

Every classification class is assigned to one of four agricultural tasks, namely disease, pest/damage (including abiotic stress), species identification, or quality/maturity. Disease classes occur in 124 classification datasets, pest/damage classes in 62, species classes in 59 and quality/maturity classes in 25 (3,605, 1,521, 3,520 and 485 images). The crop is named in the prompt when it is known, except for species identification, where prompts always say ``plant'' so that the crop name cannot reveal the answer.

\begin{table}[!tbp]
\centering
\caption{Benchmark composition. Images are subsampled per dataset (Section~\ref{sec:data}). Each classification image is asked in three formats, so classification contributes 27,483 items. VQA items are questions, some of which show several images; the 3,097 questions comprise 1,053 multiple-choice, 204 yes/no, 236 numeric and 1,604 open-ended questions.}
\label{tab:data}
\renewcommand{\arraystretch}{1.15}
\tablefont
\begin{tabular}{lrrrrl}
\toprule
\textbf{Task} & \textbf{Datasets} & \textbf{Classes} & \textbf{Images} & \textbf{Items} & \textbf{Scored by} \\
\midrule
Classification & 220 & 1,832 & 9,161 & 27,483 & option match (MCQA); LLM judge (OEQ) \\
Detection & 105 & 432 & 1,628 & 1,628 & box matching at IoU $\geq$ 0.5 \\
Segmentation & 34 & 59 & 248 & 248 & box matching at IoU $\geq$ 0.5 \\
VQA & 13 & --- & 3,592 & 3,097 & answer parsing; LLM judge (open-ended) \\
\midrule
Total & 372 & & 14,629 & 32,456 & \\
\bottomrule
\end{tabular}
\end{table}

\subsection{Prompt formats}
\label{sec:prompt_methodology}
\begin{figure*}[!tp]
    \centering
    \includegraphics[width=0.9\textwidth]{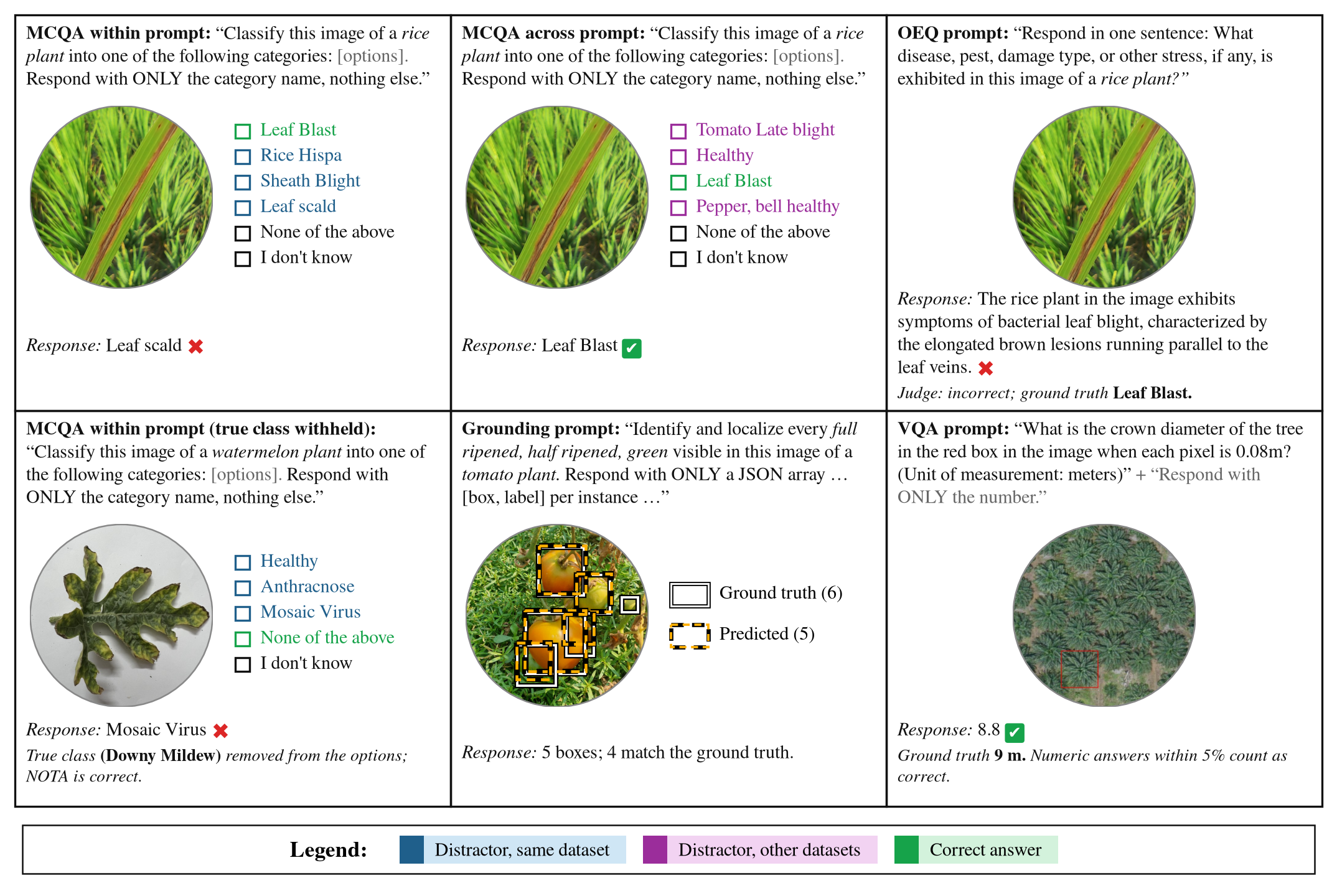}
    \caption{Prompt formats with real responses from Gemma-4-31B (seed 0). The top row shows one image under MCQA within, MCQA across and OEQ. The model picks a same-dataset distractor under MCQA within, answers correctly under MCQA across, and names a different disease under OEQ, which the LLM judge marks incorrect. The bottom row shows an MCQA item whose true class is withheld, where the model picks a listed class instead of ``None of the above''; a detection item (ground-truth boxes solid, predicted boxes dashed); and a numeric VQA question on an aerial image, which asks for a measurement rather than a class. Blue options are distractors from the same dataset, purple options are distractors from other datasets with the same agricultural task, and green marks the correct answer. In the prompts, [options] stands for the options listed beside each image, which are inserted into the prompt in that order. Full prompts are given in Table~\ref{tab:prompts}.}
    \label{fig:prompts_overview}
\end{figure*}

Table~\ref{tab:prompts} lists the prompts, and Figure~\ref{fig:prompts_overview} shows examples with real model answers. Each prompt manipulates one named variable, following the pattern-based approach of White et al.\ \cite{white_prompt_2023} and the terminology of Schulhoff et al.\ \cite{schulhoff_prompt_2024}. All templates were fixed in advance and are identical across models. The only per-model difference is the box coordinate order in grounding prompts, which follows each model's own format. No per-model prompt tuning was done, so differences between models reflect the models, not prompt optimization.

\begin{table*}[!tp]
\centering
\caption{Prompts. \{a crop\} is the crop phrase (e.g., ``a tomato plant''; always ``a plant'' for species identification). MCQA options list up to four classes in a seeded random order, followed by ``None of the above'' and ``I don't know''. In 30\% of MCQA items the true class is withheld and ``None of the above'' is correct. The OEQ question is chosen by the item's agricultural task. VQA questions are the datasets' own, with a format-specific answer instruction appended; ``None of the above'' and ``I don't know'' are appended to their multiple-choice options.}
\label{tab:prompts}
\renewcommand{\arraystretch}{1.3}
\tablefont
\begin{tabular}{p{2.0cm} p{2.8cm} p{10.4cm}}
\hline
\textbf{Format} & \textbf{Tasks} & \textbf{Prompt} \\
\hline
MCQA within & Classification & ``Classify this image of \{a crop\} into one of the following categories: \{up to four classes from the same dataset\}, None of the above, I don't know. Respond with ONLY the category name, nothing else.'' \\
\hline
MCQA across & Classification & Same as MCQA within, but the distractor classes are drawn from other datasets with the same agricultural task. \\
\hline
OEQ & Disease, Pest/Damage & ``Respond in one sentence: What disease, pest, damage type, or other stress, if any, is exhibited in this image of \{a crop\}?'' \\
\hline
OEQ & Species & ``Respond in one sentence: What plant or weed species, if any, is shown in this image?'' \\
\hline
OEQ & Quality/Maturity & ``Respond in one sentence: What quality, ripeness, or condition is exhibited in this image of \{a crop\}?'' \\
\hline
Grounding & Detection, Segmentation & ``Identify and localize every \{class names\} visible in this image of \{a crop\}. Respond with ONLY a JSON array, one object per instance found, each with exactly these two keys: "box\_2d": [y\_min, x\_min, y\_max, x\_max] (or [x\_min, y\_min, x\_max, y\_max]) --- coordinates on a 0-1000 normalized scale (top-left corner of the image is [0, 0]), in that exact order. "label": the class name for that instance, chosen from: \{class names\}. If none are visible, respond with exactly: []'' \\
\hline
VQA & Multiple choice / yes--no / numeric & \{native question and options\} + ``Respond with ONLY the letter of the correct option (or letters, comma-separated, if more than one is correct).'' / ``Respond with ONLY yes or no.'' / ``Respond with ONLY the number.'' \\
\hline
VQA & Open-ended & \{native question\} \\
\hline
\end{tabular}
\end{table*}

Every multiple-choice prompt lists the candidate classes and ends with an instruction to answer with only the category name or option letter. This output-template pattern \cite{white_prompt_2023} fixes the form of the answer across model families with very different verbosity. The open-ended (OEQ) prompts ask about the same images without a list and request a one-sentence answer. Because both formats use the same images, the gap between them shows how much of the multiple-choice score depends on the option list rather than on what the model can say about the image unaided, which is how users naturally query it.

MCQA within draws its distractors from the same dataset, e.g., other diseases of the same crop. MCQA across draws them from other datasets with the same agricultural task, e.g., a disease of a different crop, which can often be ruled out on categorical grounds alone. Both formats use the same images and the same withheld-class items, so comparing them separates confusion caused by visual similarity from the effect of an implausible option set \cite{pezeshkpour_large_2023}. We therefore also refer to the MCQA-within options as plausible and to the MCQA-across options as implausible.

Every MCQA question offers ``None of the above'' (NOTA) and ``I don't know'' (IDK) after the class options. A deployed system must recognize when none of its candidate labels applies, so in 30\% of MCQA items the true class is withheld from the options and NOTA is the correct answer. Without such items a model could score well by always choosing from the list. IDK lets a model abstain and is always scored as incorrect. The 30\% rate, the presence of IDK and their position at the end of the list are design choices, so each is varied in an experiment (Section~\ref{sec:stat_experiments}).

\subsection{Scoring}
\label{sec:metrics}
Accuracy is the share of correct items in a dataset,
\begin{equation}
  \text{Accuracy} = \frac{1}{N} \sum_{i=1}^{N} \mathbf{1}[\hat{y}_i = y_i],
\end{equation}
where $\mathbf{1}[\cdot]$ is the indicator function. For detection and segmentation, an image's score is $\text{TP}/(\text{TP}+\text{FP}+\text{FN})$. True positives (TP) are predicted boxes that match a ground-truth box, false positives (FP) are the remaining predicted boxes and false negatives (FN) are the ground-truth boxes left unmatched. A predicted box matches an unmatched ground-truth box when their intersection over union (IoU) is at least 0.5 and, for datasets with more than one class, their labels agree. The VLMs answer with boxes, not masks, so segmentation is evaluated in the same way. Each segmentation mask is converted into ground-truth boxes, one around every connected region of a class that covers at least 0.02\% of the image, and the model is asked to localize every instance with a box, as in detection. Segmentation is therefore evaluated as instance localization, and the quality of a mask outline is not measured (Appendix~\ref{app:matching}). We report the macro average, the mean of per-dataset accuracies, so that every dataset counts equally. Differences between accuracies are given in percentage points (``points''), always relative to a stated reference: another format, a default setting or another scoring of the same answers. Multiple-choice answers are matched to an option by exact, then longest-substring, then fuzzy match. Fuzzy matching lowercases both strings and replaces punctuation and underscores with spaces. It then counts a match when the reference appears as a whole word in the answer, or when the highest of three similarity scores (character-level ratio, token-set ratio and partial ratio, computed with RapidFuzz 3.14.6 \cite{bachmann_rapidfuzz_2026}) is at least 0.6. VQA answers are parsed as option letters, yes/no or numbers (Appendix~\ref{app:matching}). Empty answers and answers cut off at the token limit are wrong.

Besides accuracy, MCQA is described by five measures. Two describe the items whose true class is withheld. \emph{NOTA recall} is the share of these items answered NOTA, and the \emph{over-commit rate} is the share answered with one of the listed, all incorrect, classes. Two describe the items whose true class is listed. \emph{Listed accuracy} is the share of these items answered with the true class, any other answer, including NOTA and IDK, counting as wrong, and the \emph{false-NOTA rate} is the share answered NOTA. The \emph{IDK rate} is the share of all items answered IDK.

\begin{figure*}[!tp]
    \centering
    \includegraphics[width=0.85\textwidth]{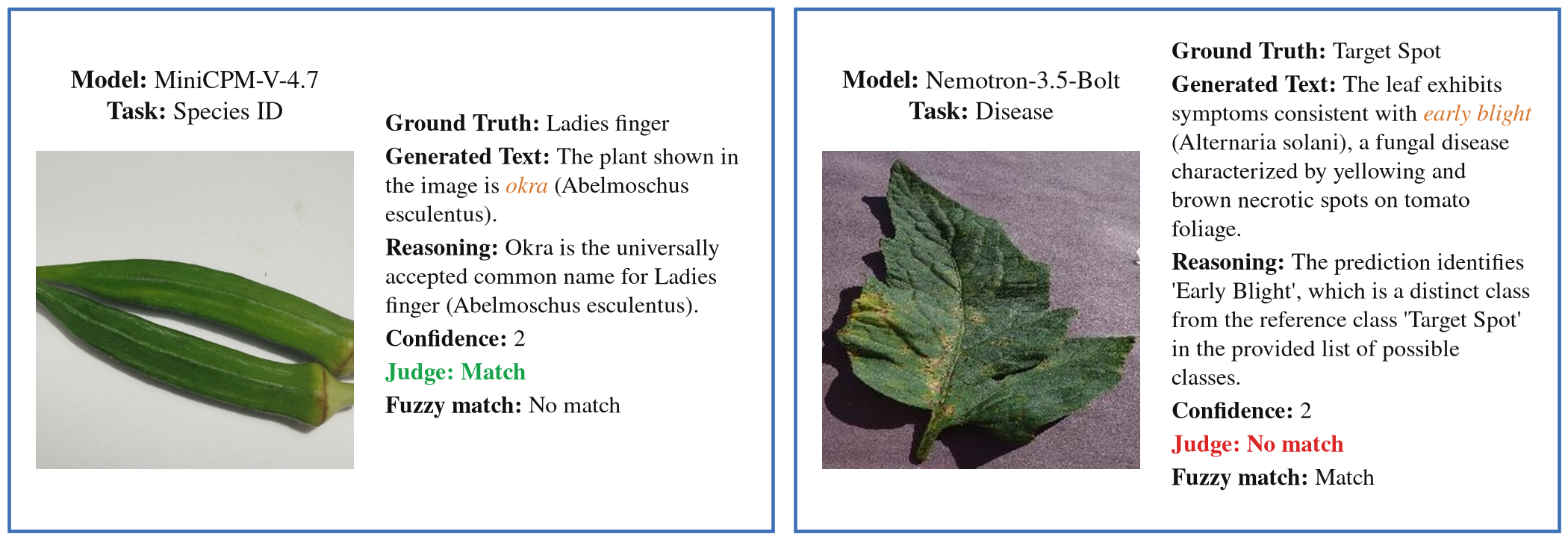}
    \caption{Example LLM-judge verdicts on OEQ answers (seed 0), with the condition each answer names in orange. On the left, a common name (okra) for the reference class ``Ladies finger'' is credited by the judge but missed by fuzzy matching. On the right, a different disease of the same crop (early blight for Target Spot) passes fuzzy matching but is rejected by the judge. Reasoning and confidence are the judge's own output.}
    \label{fig:judge_samples}
\end{figure*}

Open-ended answers (OEQ and open-ended VQA) are graded by an LLM judge, Gemma-4-31B. The judge runs with its reasoning trace turned off, so it answers directly without first generating step-by-step reasoning, and with temperature 0. It returns a JSON verdict containing whether the answer matches the reference (\code{is\_match}), a short written justification and a confidence score from 0 to 2. For OEQ, it sees the dataset's full class list and accepts an answer only if the answer identifies the reference class specifically enough to rule out every other listed class. Synonyms and common names count, but a generic label, a neighbouring condition or a related species does not (Figure~\ref{fig:judge_samples}). The judge prompt was refined until its verdicts matched a set of reference (oracle) judgments, and the full rules and prompt are in Appendix~\ref{app:judge}. \newtext{On a labelled, non-random subsample of answers from an earlier run, the final prompt agrees with a human annotator on 79\% of the answers on which earlier prompt versions disagreed and on 95\% of a sample on which they agreed (Appendix~\ref{app:judge}, Table~\ref{tab:judge_validation}).} For open-ended VQA, the answer's key facts and conclusion must agree with the reference answer. Because the judge is also an evaluated model, Gemma-4-31B's judged scores are self-judged and marked $\dagger$ throughout. For comparison, we also score each OEQ answer by fuzzy matching against its reference class, as defined above, and each open-ended VQA answer by ROUGE-L, the F1 score of the longest common word subsequence between answer and reference \cite{lin_rouge_2004}. Neither of these uses the judge.

Uncertainty is a 95\% bootstrap confidence interval \cite{efron_introduction_1993} over datasets (10,000 resamples), which asks whether a result would hold on a different set of datasets of the same kind. Differences between two models are tested with a paired bootstrap over datasets. For each dataset the two models' accuracies are subtracted, the datasets are resampled with replacement 10,000 times, and the p-value is twice the share of resamples in which the mean difference changes sign. Because both models answer the same items, a dataset's difficulty cancels in the difference, so the test can detect differences that the per-model confidence intervals hide. Comparing many pairs at once makes some differences look significant by chance, so the p-values are corrected for the 28 model pairs of each format with the Holm method \cite{holm_simple_1979}, which keeps the probability of any false positive among them at 5\%. It multiplies the smallest p-value by 28, the next by 27 and so on, and $p_{\text{Holm}}$ denotes this corrected p-value. The YOLO11 baseline is compared with each of the eight VLMs in the same way, with the correction for eight comparisons per task. Position preference under option-order randomization (Section~\ref{sec:results_stability}) is reported as an effect size, the difference between how often a position is chosen and how often it holds the correct answer, without a significance test, because the five orders of one question are not independent answers.

Figures~\ref{fig:main_results_chart} and~\ref{fig:agtask_chart} and Tables~\ref{tab:itt}, \ref{tab:main_results} and~\ref{tab:agtask} report means over the three generation seeds of the main benchmark. The NOTA analysis (Figure~\ref{fig:behaviour}, Table~\ref{tab:behaviour}), the comparison of fuzzy matching with the judge (Table~\ref{tab:judge}) and Experiments~2--4 use seed~0.

\subsection{Experiments}
\label{sec:stat_experiments}
We ran four experiments (Table~\ref{tab:experiments}, Appendix~\ref{app:methods}). Each changes one variable and keeps the evaluation items fixed (same images, distractors and withheld-class items).
\begin{itemize}
    \item In Experiment~1 (generation seeds), the full benchmark (all formats, judged) is run with three sampling seeds per model.
    \item In Experiment~2 (option order), all multiple-choice questions (MCQA within, MCQA across and VQA multiple choice) are re-asked under five seeded random option orders, with NOTA and IDK also placed at random positions instead of at the end.
    \item In Experiment~3 (NOTA rate), the share of MCQA items whose true class is withheld is changed from 30\% to 10\% and to 50\%.
    \item In Experiment~4 (IDK removed), all multiple-choice questions are re-asked without the IDK option.
\end{itemize}
Experiments 2--4 cover seven models (Nemotron was excluded for compute cost), use seed 0 and are compared item by item with the seed-0 run of Experiment 1. Generation is not bit-reproducible even with a fixed seed. Batched GPU inference makes each request's numerics depend on its batch, so two runs with the same seed produce the same output text for only about half of the items. We therefore treat seed runs as independent replicates and judge the effects in Experiments 2--4 against the seed-to-seed variation of Experiment 1.

\subsection{Baselines}
\label{sec:baselines}
Two baselines put the VLM scores in context. The chance baseline never looks at the image and is scored by the same grader as the VLMs.
It answers at random. It picks one of the listed options for multiple-choice questions (including NOTA and IDK), one of the dataset's classes for OEQ, and yes or no or one of the dataset's numeric answers for the corresponding VQA questions. For detection and segmentation it predicts a single box covering the whole image with a random class. It is averaged over three seeds. It answers OEQ with an exact class name, which is scored by exact match instead of the judge, so it knows the dataset's class list, which the VLMs never see in OEQ. It does not answer open-ended VQA questions.

The supervised baseline answers the study's central question, whether a VLM can replace a supervised model. Because a VLM is used without labelled images, we compare it with a supervised model that needs as few labels as possible, a few images per class that a practitioner could annotate quickly. YOLO11 \cite{khanam_yolov11_2024} suits this role because one widely used, lightweight model family and training recipe covers classification, detection and segmentation. We use Ultralytics 8.4.210 with model size s, train one model per task and, as in \cite{ranario_are_2025}, score YOLO on the image alone. For each of classification, detection and segmentation, a single model is trained from the pretrained YOLO11s weights for 100 epochs on 10 images per class from every dataset of that task, and, for Appendix~\ref{app:yolo}, also on 5 and 15. The training images are drawn with a fixed seed from the images that are not evaluated, preferring each dataset's training split. There are three independent subsets, and within a subset the smaller sets are contained in the larger ones. At 10 images per class, the training sets hold about 18,250 classification, 3,150 detection and 470 segmentation images.

Classes of different datasets share a label when their names are identical after normalization (lowercase, with punctuation and underscores replaced by spaces), so that, for example, powdery mildew from several crops is one class. Names without letters, such as the numeric codes of some detection datasets, stay separate for each dataset. This rule merges the 1,832 classification classes into 1,333, the 432 detection classes into 327 and the 59 segmentation classes into 43. At test time, a prediction is limited to the classes of the image's own dataset, the classes the VLMs are asked about. Classification uses the YOLO11 classification model (YOLO11-cls) and is scored by top-1 accuracy, whether the most probable of the dataset's classes is the true class. YOLO receives no question or options, so it has one accuracy per image, which is compared with every question format of the VLMs. Like the chance baseline in OEQ, it knows the dataset's class list. Detection trains on the ground-truth boxes, and segmentation trains the YOLO11 segmentation model (YOLO11-seg) on the mask components, scored through their boxes like the VLMs. VQA has no supervised baseline. The main results use 10 images per class, averaged over the three subsets like the generation seeds of the VLMs. Results at 5 and 15 images per class (Appendix~\ref{app:yolo}) lead to the same conclusions.

\section{Results}
\label{sec:results}

\subsection{Overall accuracy}
\label{sec:results_overall}

\takeaway{No model is reliable on any task, and none is best at everything. A supervised YOLO11 model trained on only 10 images per class outperforms the best VLM on every task it covers. Even the strongest answer only about half of the multiple-choice questions and a third of the open-ended ones correctly, and localization is weaker still.}

\begin{figure*}[!tp]
    \centering
    \includegraphics[width=0.88\linewidth]{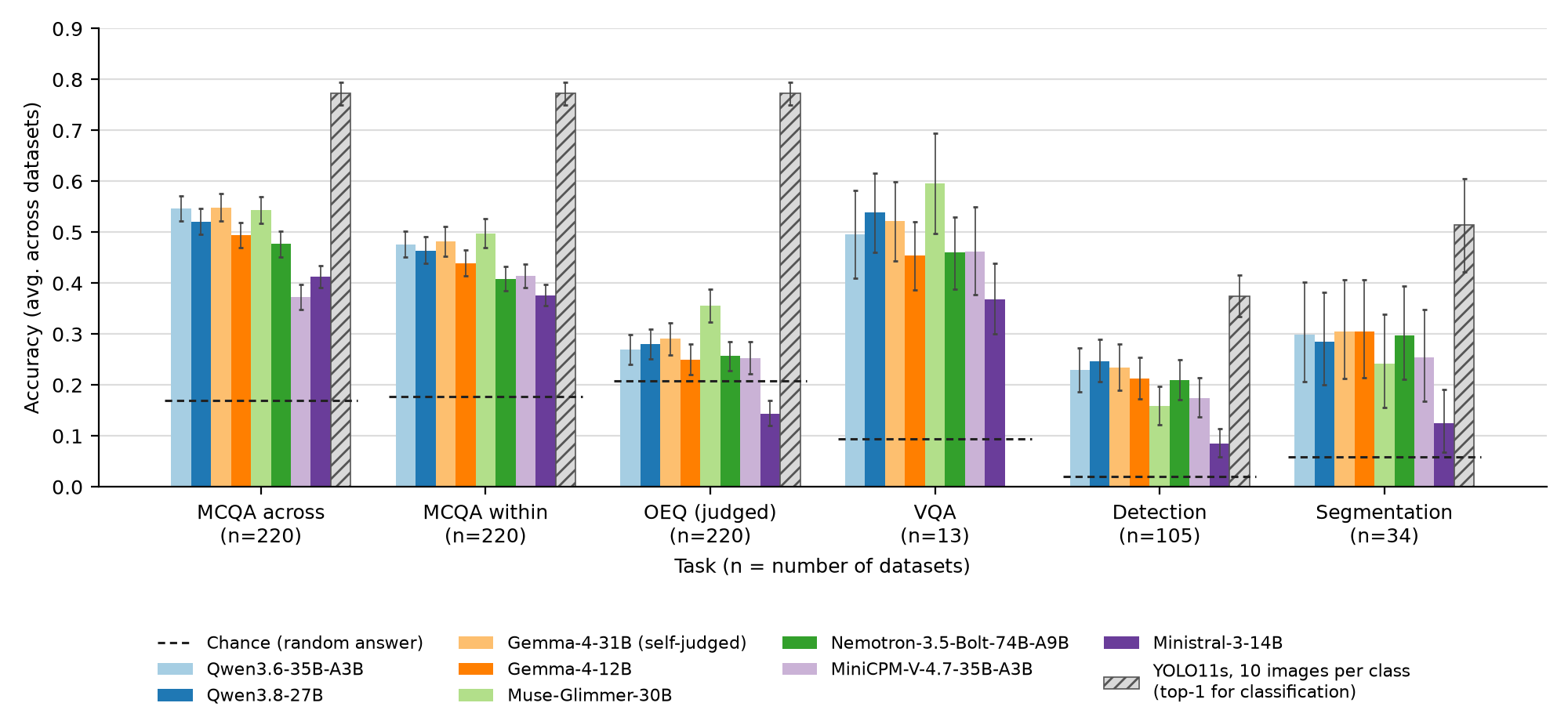}
    \caption{Accuracy by task for the eight models, mean over generation seeds (exact values in Table~\ref{tab:main_results}, Appendix~\ref{app:tables}). Error bars are 95\% bootstrap confidence intervals over datasets. Dashed lines mark the chance baseline (Section~\ref{sec:baselines}). The grey hatched bar is YOLO11 trained on 10 images per class, averaged over three training subsets (Section~\ref{sec:baselines}). It sees only the image, so its classification bar is the same top-1 accuracy in the three classification columns. VQA has no YOLO bar. $n$ is the number of datasets per task.}
    \label{fig:main_results_chart}
\end{figure*}

Figure~\ref{fig:main_results_chart} summarizes the eight models, with exact values and confidence intervals in Table~\ref{tab:main_results} (Appendix~\ref{app:tables}). In multiple-choice classification no model exceeds 55\%. Under MCQA across, Gemma-4-31B (54.9\%), Qwen3.6-35B-A3B (54.7\%), Muse-Glimmer-30B (54.4\%) and Qwen3.8-27B (52.1\%) lead, and none of the six pairwise differences among them is significant after Holm correction (all $p_{\text{Holm}} \geq 0.40$). Under MCQA within, Muse-Glimmer-30B is highest at 49.7\%. Its leads over Qwen3.6 (2.6 points) and Qwen3.8 (3.4 points) are significant ($p_{\text{Holm}} \leq 0.007$), but its lead over Gemma-4-31B is not ($p_{\text{Holm}} = 0.23$).

Open-ended classification is considerably harder. Judged OEQ accuracy ranges from 14.4\% (Ministral-3-14B) to 35.6\% (Muse-Glimmer-30B), and Muse-Glimmer leads every other model by 6.0--21.0 points ($p_{\text{Holm}} < 0.001$ for each pair). Gemma-4-31B, Qwen3.8 and Qwen3.6 follow at 27.0--29.1\%, and Nemotron, MiniCPM and Gemma-4-12B at 25.0--25.7\%.

Localization is the weakest capability. Detection accuracy ranges from 8.5\% (Ministral) to 24.7\% (Qwen3.8) and segmentation accuracy from 12.5\% to 30.6\%. On detection, the five strongest models do not differ significantly from one another, and Muse-Glimmer-30B, the strongest open-ended model, is 5.4--8.6 points below each of them ($p_{\text{Holm}} < 0.001$). On segmentation, which has only 34 datasets, Ministral is the only model that differs significantly from the others. VQA accuracy over the 13 datasets ranges from 36.9\% (Ministral) to 59.7\% (Muse-Glimmer), with wide confidence intervals (about $\pm$8 points) because there are only 13 datasets.

Every model exceeds the chance baseline on MCQA, detection, segmentation and VQA (Table~\ref{tab:main_results}). A random option scores 16.9\% on MCQA across and 17.8\% on MCQA within. On OEQ, naming a random class of the dataset scores 20.8\%, which is above Ministral (14.4\%) and within 5 points of Nemotron, MiniCPM and Gemma-4-12B (25.0--25.7\%). This comparison favours the baseline, which knows the dataset's class list, but it shows how little most models' open-ended answers add beyond knowing which classes exist.

No model leads every task. Muse-Glimmer-30B is best at open-ended answers (OEQ and VQA) but second-lowest on both detection and segmentation. Gemma-4-31B and the two Qwen models are the most balanced. Ministral-3-14B, the only model without native reasoning, is last on every task except MCQA across, where MiniCPM-V-4.7 is lower (37.3\% vs.\ 41.2\%).

The few-shot supervised baseline outperforms every VLM on every task it covers (Figure~\ref{fig:main_results_chart}, Table~\ref{tab:main_results}). Trained on 10 images per class, YOLO11 reaches a top-1 classification accuracy of 77.3\%, against 54.9\% for the best VLM under MCQA across and 35.6\% under OEQ. Its top-1 accuracy ranges from 72.9\% (pest/damage) to 79.5\% (species) across agricultural tasks, whereas no VLM exceeds 58.6\% on any agricultural task in any format (Figure~\ref{fig:agtask_chart}). On detection and segmentation it reaches 37.5\% and 51.4\%, against 24.7\% and 30.6\%. Its advantage over every VLM is significant on every task (paired bootstrap over datasets, $p_{\text{Holm}} < 0.001$ for each of the eight models), and the smallest margin, over Qwen3.8 on detection, is 12.8 points (95\% CI 9.0--16.5). YOLO sees no options, so this comparison is not tied to a question format. It does know the dataset's class list, which favours it over the VLMs in OEQ but not in MCQA, where the VLMs are given the candidate classes. The conclusion does not depend on the choice of 10 images per class. With 5 images per class YOLO already outperforms the best VLM on every task, and with 15 it gains further (Appendix~\ref{app:yolo}).

\subsection{Question format}
\label{sec:results_format}

\takeaway{Much of the accuracy comes from the option list itself. Asking the same images open-ended, as users naturally would, lowers accuracy by 12--28 points. Distractors from the same dataset lower it by 3--8 points compared with distractors from other datasets. Whether a task looks easy or hard also depends on the format.}

\begin{figure*}[!tp]
    \centering
    \includegraphics[width=0.88\linewidth]{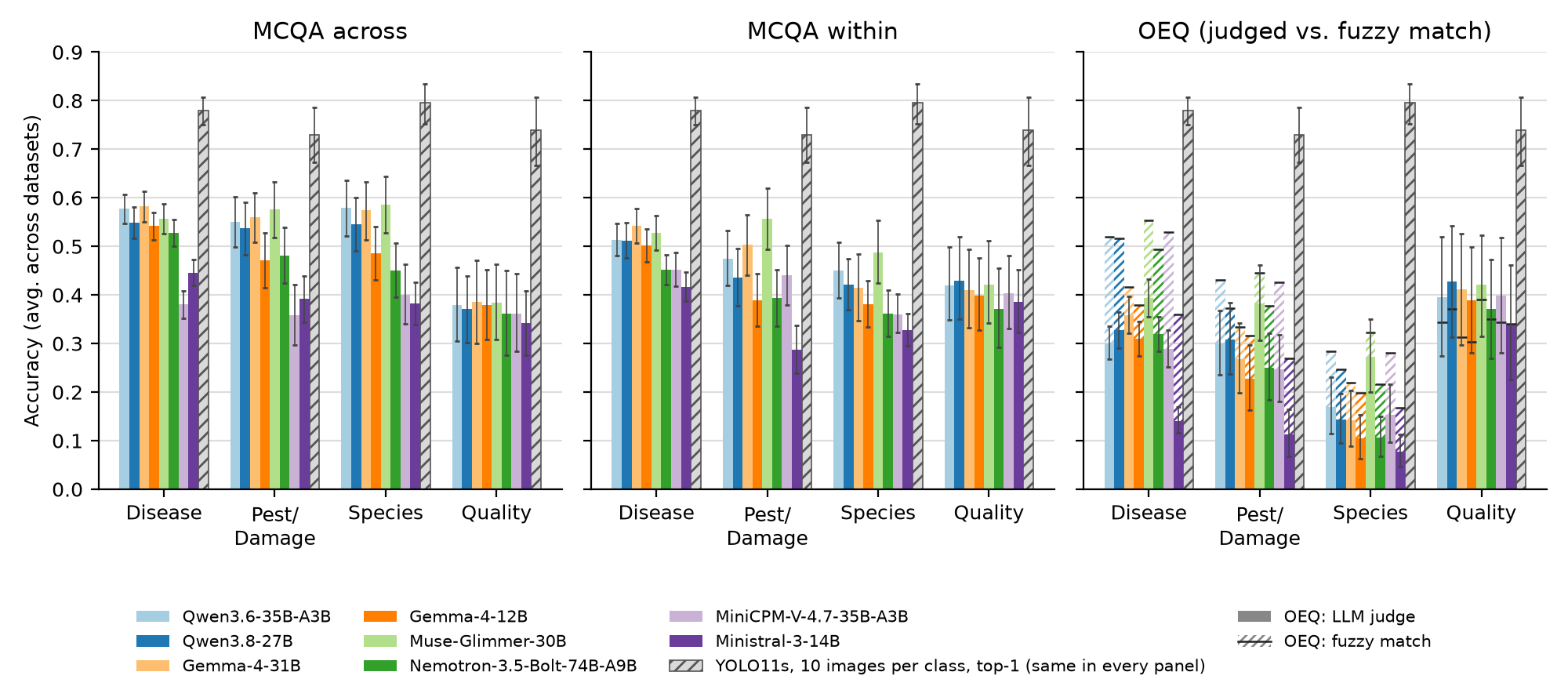}
    \caption{Classification accuracy by agricultural task for each format, mean over generation seeds of the macro average over datasets (exact values in Table~\ref{tab:agtask}, Appendix~\ref{app:tables}). Error bars are 95\% bootstrap confidence intervals over datasets. In the OEQ panel, solid bars (with error bars) are judged accuracy; the hatched bar behind each and its dark tick show the same answers scored by fuzzy string matching. A tick below the top of a solid bar means the judge is more lenient than fuzzy matching (Section~\ref{sec:results_design}; per-model values in Table~\ref{tab:judge}). The grey hatched bar is the top-1 accuracy of YOLO11 trained on 10 images per class (mean over three training subsets, Section~\ref{sec:baselines}). It sees only the image, so the same bar appears in every panel.}
    \label{fig:agtask_chart}
\end{figure*}

Removing the option list has the largest effect. On the same images, every model loses between 12.0 (MiniCPM) and 27.7 (Qwen3.6) points when it must name the class instead of choosing it from a list (MCQA across vs.\ OEQ). Part of this gap may come from answers that describe the condition without naming it the way the dataset does, but the open-ended score is the one that reflects how models are actually queried.

MCQA within is harder than MCQA across for seven of eight models, by 3.2 (Ministral) to 7.6 points (Qwen3.6), all with $p < 0.005$. MiniCPM-V-4.7 is the exception and scores 4.0 points higher within than across. Section~\ref{sec:results_nota} shows that the gap is driven mainly by how often models reject the whole list.

Averaged over models (Figure~\ref{fig:agtask_chart}; Table~\ref{tab:agtask}), disease is the easiest task under both multiple-choice formats (52.0\% across, 49.0\% within) and second easiest under OEQ (30.5\%), with pest and damage close behind (49.1\%, 43.5\% and 26.2\%). The other two tasks behave very differently from one format to the next. Species identification is close to disease with distractors from other datasets (50.0\%), drops to 40.0\% when the distractors are other species from the same dataset, and is the hardest task under OEQ (14.6\%). Many species datasets distinguish closely related species or cultivars, some by abbreviated or coded names, which are hard to tell apart and nearly impossible to name in an open-ended answer. Quality and maturity shows the opposite pattern. It is the hardest task under MCQA across (37.1\%) but the easiest under OEQ (39.5\%), because its classes (e.g., ripe vs.\ unripe) are generic condition words that a one-sentence description naturally contains. The supervised baseline does not share this pattern. Its top-1 accuracy is highest for species identification (79.5\%), the task the VLMs find hardest under OEQ, so the difficulty lies in naming these classes rather than in telling the images apart.

\begin{table*}[!tp]
\centering
\caption{VQA accuracy (\%) by question format, pooled over the questions of all 13 datasets and averaged over generation seeds. $N$ is the number of questions. Open-ended answers are graded by the LLM judge. ROUGE-L (\%) is computed on open-ended answers against the reference answer without the judge. Best value per column in bold. The last row is the chance baseline (Section~\ref{sec:baselines}), which does not answer open-ended questions. $^\dagger$Self-judged on open-ended questions.}
\label{tab:itt}
\renewcommand{\arraystretch}{1.2}
\tablefont
\begin{tabular}{lccccc}
\toprule
\multirow{2}{*}{\textbf{Model}} & \textbf{Multiple choice} & \textbf{Yes/no} & \textbf{Numeric} & \textbf{Open-ended (judged)} & \multirow{2}{*}{\textbf{ROUGE-L}} \\
 & $N=1{,}053$ & $N=204$ & $N=236$ & $N=1{,}604$ & \\
\midrule
Qwen3.6-35B-A3B & $60.4$ & $77.0$ & $17.8$ & $44.4$ & $7.7$ \\
Qwen3.8-27B & $\mathbf{60.6}$ & $73.4$ & $\mathbf{23.4}$ & $51.8$ & $6.2$ \\
Gemma-4-31B$^\dagger$ & $58.3$ & $\mathbf{78.4}$ & $17.7$ & $50.6$ & $8.9$ \\
Gemma-4-12B & $52.5$ & $75.7$ & $17.9$ & $41.4$ & $7.7$ \\
Muse-Glimmer-30B & $58.5$ & $73.7$ & $22.2$ & $\mathbf{64.7}$ & $7.1$ \\
Nemotron-3.5-Bolt-74B-A9B & $53.5$ & $71.9$ & $18.9$ & $42.1$ & $\mathbf{13.0}$ \\
MiniCPM-V-4.7-35B-A3B & $41.1$ & $77.3$ & $10.9$ & $51.6$ & $9.0$ \\
Ministral-3-14B & $41.0$ & $69.6$ & $16.7$ & $33.4$ & $12.7$ \\
\midrule
\textit{Chance (random answer)} & $15.8$ & $51.6$ & $8.2$ & --- & --- \\
\bottomrule
\end{tabular}

\end{table*}

Within VQA (Table~\ref{tab:itt}), yes/no questions are easiest (69.6--78.4\%) and numeric questions hardest (10.9--23.5\%), with multiple choice (41.0--60.6\%) and open-ended questions (33.4--64.7\%) in between. Numeric questions ask for counts or measurements, which current VLMs rarely get right.

\subsection{Recognizing a missing answer}
\label{sec:results_nota}

\takeaway{Models have some sense of whether the correct answer is among the options, but not enough. When it is missing and the other options are plausible (MCQA within), most models guess a wrong one. Making the options implausible (MCQA across) does not sharpen that sense; it only makes models reject the list more often, whether or not the answer is in it.}

\begin{figure}[!tbp]
\centering
\includegraphics[width=0.92\linewidth]{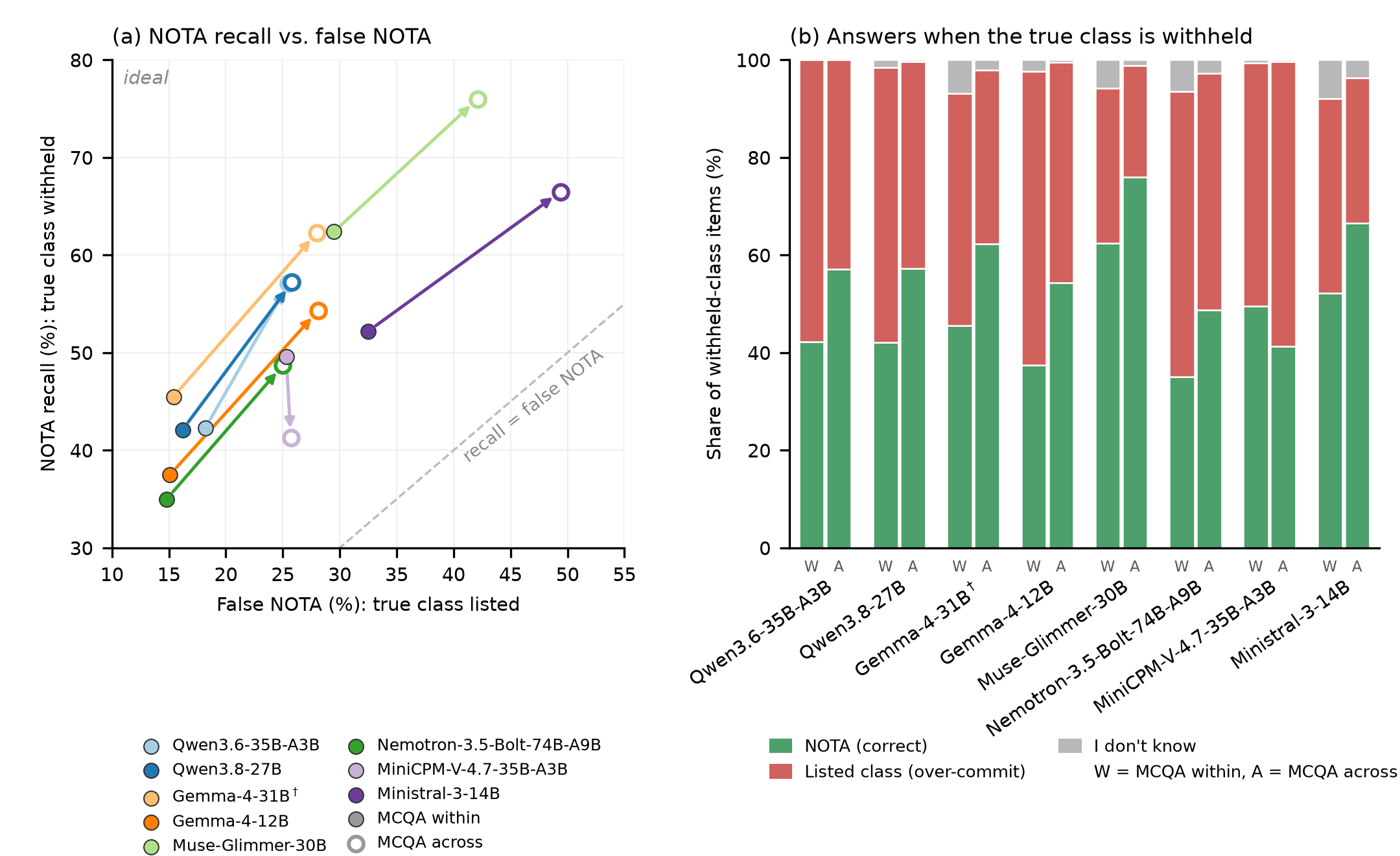}
\caption{What models do when the correct class is not among the options (Experiment~1, seed 0, default 30\% NOTA rate). (a) Whether models can tell a missing answer from a listed one. The y-axis is how often a model answers NOTA when the answer is missing (NOTA recall, also the green share of the bars in panel b), and the x-axis is how often it answers NOTA when the answer is listed (false NOTA). The upper left is ideal, and a model on the dashed diagonal answers NOTA equally often in both cases. Arrows run from plausible options (MCQA within, filled) to implausible options (MCQA across, hollow). (b) Answers on the questions whose true class is withheld, either ``None of the above'' (correct), a wrong listed class (over-commit) or ``I don't know''. Each model has two bars, W with plausible options from the same dataset (MCQA within) and A with implausible options from other datasets (MCQA across). Per-model values are in Table~\ref{tab:behaviour} (Appendix~\ref{app:tables}). $^\dagger$Gemma-4-31B is also the LLM judge (not used for MCQA scoring).}
\label{fig:behaviour}
\end{figure}

In 30\% of MCQA questions the true class is withheld, so the correct answer is ``None of the above'' (NOTA). Figure~\ref{fig:behaviour} asks two linked questions about these items: what do models answer when the correct class is missing (panel b), and can they tell that situation apart from one where it is listed (panel a)?

Figure~\ref{fig:behaviour}b shows that most models guess when the correct class is missing. With plausible options, that is, other classes of the same dataset (MCQA within), models pick a wrong listed class for 31.8--60.2\% of these questions, and five of the eight models do so more often than they answer NOTA. Models almost never choose ``I don't know''. With implausible options from other datasets (MCQA across), the share of NOTA answers grows for every model except MiniCPM.

Figure~\ref{fig:behaviour}a shows that models can partly distinguish the two situations, but that implausible options (MCQA across) do not improve this ability. More NOTA answers are an improvement only if models do not also answer NOTA when the correct class is listed. Figure~\ref{fig:behaviour}a therefore plots the NOTA rate on questions whose correct class is missing against the NOTA rate on questions whose correct class is listed. Every model lies above the diagonal, so each one answers NOTA more often when it is correct than when it is wrong. When the options change from plausible (MCQA within) to implausible (MCQA across), however, the models move up and to the right, roughly parallel to the diagonal, rather than towards the ideal upper-left corner. In other words, they answer NOTA more often in both situations. Because questions whose correct class is listed make up 70\% of all questions, most of these additional NOTA answers are wrong. MiniCPM is the exception, because under MCQA across it mostly picks a distractor from another dataset.

Taken together, the two panels show that plausible options (MCQA within) lead models to guess, whereas implausible options (MCQA across) make them reject the option list more readily but not more selectively. Muse-Glimmer and Ministral, the two models with the largest share of NOTA answers in Figure~\ref{fig:behaviour}b, are also the two furthest to the right in Figure~\ref{fig:behaviour}a, which means that they answer NOTA readily whether or not it is correct. Benchmarks that use implausible distractors, as MCQA across does, therefore overstate how well models recognize that the correct diagnosis is missing. In the field, the alternatives are usually similar conditions of the same crop, and under those conditions most models guess.

Abstention plays almost no role. ``I don't know'' accounts for at most 8.4\% of answers, and it is not aimed at the questions whose answer is missing. Each model uses it about as often whether or not the true class is listed, and more often when the options are similar. It reflects how alike the options look, not whether the correct answer is available.

\subsection{Benchmark design and scoring}
\label{sec:results_design}

\takeaway{The NOTA rate decides how much a model's readiness to answer ``None of the above'' counts, and with it which model ranks first. The IDK option, the position of NOTA and IDK, and the scoring of open-ended answers also shift accuracy by several points and can reorder models.}

\begin{table*}[!tp]
\centering
\caption{NOTA rate (Experiment 3; seed 0, seven models). MCQA accuracy combines NOTA recall, the accuracy on items whose true class is withheld, with listed accuracy, the accuracy on items whose true class is among the options (Section~\ref{sec:prompt_methodology}); both are measured at the default 30\% rate (\%). $\Delta$ at 10\% and $\Delta$ at 50\% give the measured change in accuracy (points) when 10\% or 50\% of items have their true class withheld, relative to the same model at 30\%. In parentheses is the change predicted, dataset by dataset, from NOTA recall and listed accuracy alone, i.e., if the model answered every item as before. $^{*}$95\% paired bootstrap CI over datasets excludes 0.}
\label{tab:ablation_nota}
\renewcommand{\arraystretch}{1.2}
\tablefont
\setlength{\tabcolsep}{4pt}
\begin{tabular}{lcccccccc}
\toprule
\multirow{2}{*}{\textbf{Model}} & \multicolumn{4}{c}{\textbf{MCQA within}} & \multicolumn{4}{c}{\textbf{MCQA across}} \\
\cmidrule(lr){2-5} \cmidrule(lr){6-9}
 & \makecell{NOTA\\recall} & \makecell{Listed\\acc.} & \makecell{$\Delta$ at 10\%\\(pred.)} & \makecell{$\Delta$ at 50\%\\(pred.)} & \makecell{NOTA\\recall} & \makecell{Listed\\acc.} & \makecell{$\Delta$ at 10\%\\(pred.)} & \makecell{$\Delta$ at 50\%\\(pred.)} \\
\midrule
Qwen3.6-35B-A3B & $42.3$ & $49.1$ & $+2.3^{*}$ ($+1.4$) & $-0.5$ ($-1.2$) & $57.1$ & $53.4$ & $-1.8^{*}$ ($-0.9$) & $+0.5$ ($+0.6$) \\
Qwen3.8-27B & $42.1$ & $48.0$ & $+0.7$ ($+1.2$) & $-1.5^{*}$ ($-1.0$) & $57.2$ & $51.9$ & $-2.3^{*}$ ($-0.9$) & $+0.2$ ($+1.2$) \\
Gemma-4-31B$^\dagger$ & $45.5$ & $49.4$ & $+0.6$ ($+0.9$) & $-0.9$ ($-0.6$) & $62.3$ & $51.7$ & $-1.5^{*}$ ($-2.0$) & $+1.7^{*}$ ($+2.2$) \\
Gemma-4-12B & $37.5$ & $46.5$ & $+1.7^{*}$ ($+1.8$) & $-0.4$ ($-1.6$) & $54.3$ & $47.2$ & $-0.7$ ($-1.1$) & $+2.7^{*}$ ($+1.8$) \\
Muse-Glimmer-30B & $62.4$ & $43.3$ & $-3.5^{*}$ ($-4.4$) & $+4.1^{*}$ ($+3.4$) & $76.0$ & $45.3$ & $-5.6^{*}$ ($-6.0$) & $+5.9^{*}$ ($+6.3$) \\
MiniCPM-V-4.7-35B-A3B & $49.6$ & $37.1$ & $-2.2^{*}$ ($-2.8$) & $+3.0^{*}$ ($+2.3$) & $41.3$ & $35.5$ & $-1.4^{*}$ ($-1.1$) & $+1.6^{*}$ ($+1.2$) \\
Ministral-3-14B & $52.2$ & $31.4$ & $-4.2^{*}$ ($-4.1$) & $+4.3^{*}$ ($+4.2$) & $66.5$ & $29.8$ & $-7.2^{*}$ ($-7.4$) & $+7.7^{*}$ ($+7.3$) \\
\bottomrule
\end{tabular}

\end{table*}

Changing the NOTA rate (Experiment~3) moves accuracy mostly because it changes the mix of questions, not because models answer differently. MCQA accuracy combines two scores, each measured on its own group of questions. NOTA recall is the accuracy on the questions whose true class is withheld, and listed accuracy is the accuracy on the questions whose true class is among the options (Section~\ref{sec:prompt_methodology}). With $r$ the share of withheld questions,
\begin{equation}
  \text{Accuracy} = r \cdot \text{NOTA recall} + (1 - r) \cdot \text{Listed accuracy}.
\end{equation}
Raising $r$ gives more weight to NOTA recall, and lowering it gives more weight to listed accuracy. A model whose NOTA recall is well above its listed accuracy therefore gains accuracy as $r$ rises, even if it answers every question exactly as before.

Table~\ref{tab:ablation_nota} shows that this reweighting accounts for the measured changes. For each model and format it lists the two scores at the default 30\% rate, followed by the measured change in accuracy at 10\% and 50\%, with the change predicted from the two scores in parentheses. Across all models, formats and rates, the predicted change differs from the measured change by at most 1.4 points (mean absolute difference 0.6 points), and NOTA recall itself does not change significantly with the rate for any model or format.

The size of the change therefore follows the gap between a model's two scores. Muse-Glimmer and Ministral answer NOTA readily, so their NOTA recall exceeds their listed accuracy by 19.1--36.8 points (the difference between the first two columns of Table~\ref{tab:ablation_nota}), and their accuracy changes the most. Under MCQA across, for example, Ministral has a NOTA recall of 66.5\% and a listed accuracy of 29.8\%, and its accuracy falls by 7.2 points at 10\% and rises by 7.7 points at 50\%. For the other five models this difference is between $-$9.0 and $+$12.5 points, and their measured changes in Table~\ref{tab:ablation_nota} are all within 3.0 points of zero. Model rankings therefore depend on a number that the benchmark designer chooses. At a 50\% rate, Muse-Glimmer (60.3\%) would lead MCQA across, 3.8 points ahead of the next model, whereas at 10\% it would be fourth. Because neither NOTA recall nor listed accuracy depends on the rate, reporting them separately avoids the problem.

\begin{table}[!tbp]
\centering
\caption{Removing ``I don't know'' (Experiment 4; seed 0, seven models). IDK (\%) is the share of answers that are ``I don't know'' when the option is offered, and $\Delta$Acc.\ the change in accuracy (points) when it is removed, compared item by item with the same questions with IDK offered. $^{*}$95\% paired bootstrap CI over datasets excludes 0. VQA MC values are dominated by a single MIRAGE question (Section~\ref{sec:results_stability}).}
\label{tab:ablation_idk}
\renewcommand{\arraystretch}{1.2}
\tablefont
\begin{tabular}{lcccccc}
\toprule
\multirow{2}{*}{\textbf{Model}} & \multicolumn{2}{c}{\textbf{MCQA within}} & \multicolumn{2}{c}{\textbf{MCQA across}} & \multicolumn{2}{c}{\textbf{VQA MC}} \\
\cmidrule(lr){2-3} \cmidrule(lr){4-5} \cmidrule(lr){6-7}
 & \makecell{IDK\\(\%)} & \makecell{$\Delta$Acc.\\(pts)} & \makecell{IDK\\(\%)} & \makecell{$\Delta$Acc.\\(pts)} & \makecell{IDK\\(\%)} & \makecell{$\Delta$Acc.\\(pts)} \\
\midrule
Qwen3.6-35B-A3B & $0.0$ & $+1.6^{*}$ & $0.0$ & $+1.0$ & $0.3$ & $+16.4^{*}$ \\
Qwen3.8-27B & $1.8$ & $+0.5$ & $0.2$ & $-0.9$ & $0.4$ & $-0.6$ \\
Gemma-4-31B$^\dagger$ & $8.4$ & $+2.4^{*}$ & $2.2$ & $+0.9^{*}$ & $0.6$ & $-0.4$ \\
Gemma-4-12B & $2.6$ & $+0.2$ & $0.4$ & $+0.6$ & $0.6$ & $+0.3$ \\
Muse-Glimmer-30B & $6.7$ & $+1.6^{*}$ & $1.4$ & $+0.7$ & $1.6$ & $-0.7$ \\
MiniCPM-V-4.7-35B-A3B & $0.9$ & $-1.4^{*}$ & $0.2$ & $-2.7^{*}$ & $0.1$ & $-2.0^{*}$ \\
Ministral-3-14B & $6.6$ & $+1.5^{*}$ & $4.2$ & $+1.2^{*}$ & $3.2$ & $+0.6$ \\
\bottomrule
\end{tabular}

\end{table}

Removing the IDK option (Experiment~4) changes MCQA accuracy by $-$2.7 to $+$2.4 points compared with the same questions when IDK is offered (Table~\ref{tab:ablation_idk}), and the effect depends on how often a model used the option. Gemma-4-31B, Muse-Glimmer and Ministral, which answered IDK for 6.6--8.4\% of MCQA-within questions, gain 1.5--2.4 points when it is removed. Without IDK, their uncertain answers have to go somewhere else. A few become correct answers, but most become NOTA. On questions whose true class is withheld a NOTA answer is correct, so NOTA recall rises, by 4.9 points for Gemma-4-31B and 4.6 points for Muse-Glimmer. On questions whose true class is listed the same answer is wrong, so the false-NOTA rate rises as well, by 5.6 and 6.1 points. On balance, accuracy rises slightly. MiniCPM is the only model whose accuracy falls in all three formats (by 1.4, 2.7 and 2.0 points), even though it rarely chose IDK (at most 0.9\% of answers), which suggests that the option also affects how it chooses among the remaining answers. An abstention option is therefore not neutral. When it is offered, it absorbs uncertain answers that would otherwise be spread over the listed classes and NOTA.

The position of NOTA and IDK matters as well. When NOTA and IDK are placed at random positions instead of last (Experiment~2), MCQA-within accuracy falls by 1.1--3.2 points for five models (e.g., Muse-Glimmer 49.6\% $\rightarrow$ 46.9\%, Ministral 37.6\% $\rightarrow$ 34.4\%) and changes by at most 0.2 points for Qwen3.6 and MiniCPM.

The scoring of open-ended answers has the largest effect of all. Fuzzy matching credits answers that reproduce part of the class string. Because one of its measures is a partial-string ratio, an answer that mentions one word of a multi-word label, or a fragment of a short coded label, can pass the 0.6 threshold, whereas the judge requires the answer to identify the reference class. Fuzzy accuracy therefore exceeds judged accuracy for every model, by 4.7 (Gemma-4-31B) to 17.4 (MiniCPM) points (Figure~\ref{fig:agtask_chart}, OEQ panel; seed-0 values in Table~\ref{tab:judge}, Appendix~\ref{app:tables}). The gap is largest for disease (4.9--23.7 points), plausibly because disease names share crop and symptom words (``leaf'', ``spot'', ``blight'') that a one-sentence answer is likely to contain. For quality and maturity, in contrast, the judge is more lenient than fuzzy matching for six of eight models (by 2.4--11.3 points), because it accepts paraphrases of generic condition words. The choice of scorer also reorders models, since MiniCPM-V-4.7 is the second-best OEQ model by fuzzy matching (43.6\%) but only sixth by the judge (26.2\%).
Gemma-4-31B has the smallest gap of any model, which may partly reflect self-judging. For open-ended VQA, ROUGE-L does not track judged correctness either (Table~\ref{tab:itt}). Nemotron and Ministral have the highest ROUGE-L (13.0\% and 12.7\%) but are among the three lowest by the judge, while Muse-Glimmer, the best by the judge, has one of the lowest ROUGE-L scores (7.1\%). Many reference answers are long descriptions, so n-gram overlap rewards wording that resembles the reference rather than a correct conclusion.

\subsection{Stability across seeds and option orders}
\label{sec:results_stability}
\label{sec:results_seeds}

\takeaway{Model rankings are not artifacts of sampling noise, but a single answer to a single image can change with the seed or with the order in which options are listed.}

Across the three generation seeds, a model's macro accuracy for MCQA, OEQ, detection and segmentation varies by a standard deviation of at most 1.12 points (Table~\ref{tab:seeds}, Appendix~\ref{app:tables}), far less than most differences between models. Individual answers are much less stable, and only 44--84\% of MCQA items receive the same answer in every seed. OEQ (82--95\%) and localization (85--97\%) items agree more often, but there agreement means the same score, a coarser outcome than the same chosen option. The models with the lowest agreement (Nemotron, Qwen3.6, Gemma-4-12B) are not those with the lowest accuracy, so answer-level variability is a property of a model's sampling, distinct from its average skill.

The two observations are consistent. Answers change mainly on items the model is unsure about. In MCQA, items whose answer differs between seeds are answered correctly 26--37\% of the time, against 41--68\% for items with the same answer in every seed. The changes also run in both directions about equally often. Between seeds 0 and 1, 380--1,018 MCQA items per model and format change from correct to incorrect and 371--975 from incorrect to correct, so over the 9,161 items of each format they largely cancel. The changes arise because each answer is sampled at the temperature of the model's card and because batched GPU inference is not bit-reproducible (Section~\ref{sec:stat_experiments}). With a reasoning trace, a single different early token can lead to a different final answer. Reordering the options changes the prompt itself, so the same uncertain items change even more often across orders.

VQA is the exception. MIRAGE contributes a single multiple-choice question and AgroBench a single open-ended one. When that one question flips between seeds, the macro average over the seven datasets with multiple-choice questions moves by 14.3 points, giving standard deviations of up to 8.9 points. Qwen3.6, for example, scores 50.4\%, 52.0\% and 65.4\% across seeds, but 58.8\%, 60.7\% and 59.7\% without MIRAGE. Table~\ref{tab:itt} therefore pools VQA questions instead of averaging over datasets.

\begin{figure}[!tbp]
\centering
\includegraphics[width=0.9\linewidth]{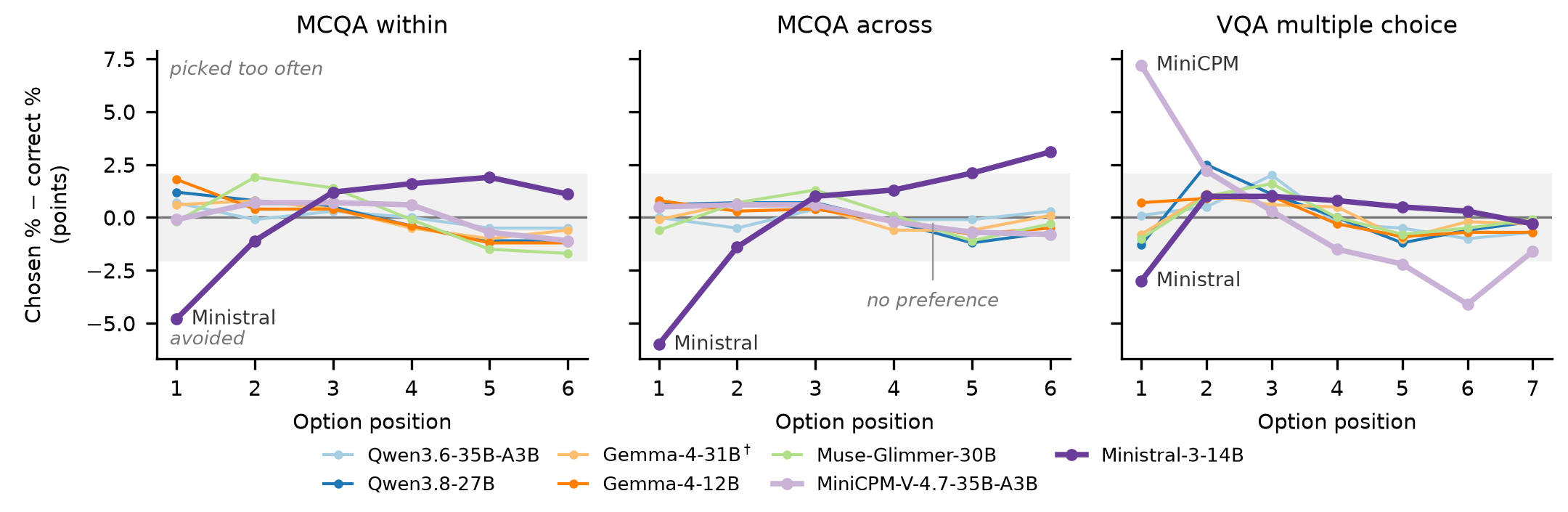}
\caption{Position preference under option-order randomization (Experiment~2, five orders pooled). For each option position, the share of answers at that position minus the share of items whose correct answer is there (points). Zero means the model chooses the position exactly as often as it should; above zero, too often; below, it avoids the position. The grey band marks $\pm$2 points. VQA multiple choice has up to 12 options; positions beyond 7 hold under 1\% of answers and are not shown. Nemotron was not run in Experiment~2.}
\label{fig:position}
\end{figure}

Re-asking every multiple-choice question in five random option orders gives the same picture (Table~\ref{tab:ablation_order}, Appendix~\ref{app:tables}; Figure~\ref{fig:position}). Aggregate accuracy barely moves, with a standard deviation across orders of 0.19--0.61 points on MCQA, comparable to the seed-to-seed variation, although VQA multiple choice again swings more for three models because of the single MIRAGE question. Individual answers, by contrast, change often. Only 27--57\% of MCQA-within items and 30--64\% of MCQA-across items receive the same answer in all five orders, fewer than across seeds, though five orders also give more chances to disagree.

Preferences for particular option positions are small for most models. On MCQA, six of the seven models choose no position more than 1.9 points more or less often than the correct answer appears there, with a mild lean towards early positions. Two models are exceptions. Ministral-3-14B avoids the first option, choosing it for 12.7\% of MCQA-within and 10.7\% of MCQA-across answers although the correct answer is first in 17.5\% and 16.7\% of items. MiniCPM-V-4.7 does the opposite on VQA multiple choice, choosing the first option 7.2 points more often than the correct answer is there.

\section{Discussion}
\label{sec:discussion}

\subsection{Implications for agricultural use}
Across 372 datasets, no open-weight VLM reaches an accuracy that would support unsupervised agricultural decisions. The best models answer about half of the multiple-choice questions correctly, name the correct class in at most about a third of open-ended questions, and localize objects poorly. This agrees with recent agricultural evaluations showing that strong VLMs struggle with fine-grained recognition \cite{arshad_leveraging_2025, shinoda_agrobench_2025}. Off-the-shelf VLMs should therefore be treated as assistive tools rather than standalone diagnostic systems, unless they are paired with domain validation and safeguards.

The results point to four considerations for using and evaluating such systems. First, open-ended accuracy is the realistic measure. Users naturally ask a model what is wrong with a plant rather than choosing from a list, and in that setting the best model names the correct class for at most 35.6\% of images. Multiple-choice accuracy on the same images is 12--28 points higher, but that gain comes from the option list rather than from better perception, so multiple-choice scores should not be read as field performance. Second, when multiple choice is used, for benchmarking or in a constrained interface, its design must be controlled and reported, because the score depends heavily on it. Distractors from the same dataset lower accuracy by 3.2--7.6 points compared with distractors from other datasets, the share of questions whose answer is NOTA moves accuracy by up to 7.7 points and reorders models, removing IDK changes it by up to 2.7 points, and moving NOTA and IDK away from the end of the list lowers it by up to 3.2 points. Models also rarely notice when the correct answer is missing, so neither a benchmark nor a deployed system can rely on them to answer ``none of the above'' or ``I don't know'' (Sections~\ref{sec:results_format}--\ref{sec:results_design}). Third, because a single answer can change with the seed or the option order, several samples or orders should be combined before acting on a diagnosis. Fourth, detection, segmentation and numeric questions are where current models fail most, and dedicated detectors or tools are needed there. A small YOLO11 model trained on 10 labelled images per class already outperforms every VLM on classification, detection and segmentation, so collecting a few labels may be a better investment than prompting a larger VLM.

The confidence intervals in Figure~\ref{fig:main_results_chart}, especially for VQA (13 datasets) and segmentation (34 datasets), reflect the heterogeneity of the AgML collection, whose datasets differ in their number of classes, class imbalance, image quality and similarity between classes. Aggregate accuracies are therefore broad summaries rather than predictions for any single dataset. Because all models are evaluated on the same items, paired comparisons are much tighter than the per-model intervals suggest, so some differences that look small are significant and some that look large are not.

\subsection{Implications for benchmark design}
Several settings that are often treated as implementation details changed our conclusions. Compared with the default setting, accuracy changed by 3.2--7.6 points when distractors came from the same dataset instead of other datasets. It changed by up to 7.7 points when the NOTA rate was 10\% or 50\% instead of 30\%, a pure reweighting of two unchanged scores that nonetheless reorders the top models. It changed by up to 2.7 points when IDK was removed, and by up to 3.2 points when NOTA and IDK were placed at random positions instead of last. Scoring the same open-ended answers by fuzzy matching instead of the judge raised accuracy by up to 17.4 points and reordered models. An LLM judge that requires a specific diagnosis is stricter and more meaningful than fuzzy matching, but it brings its own assumptions (judge model, prompt and the definition of ``specific enough''). We therefore recommend that agricultural VLM benchmarks report the answer format, distractor source, NOTA rate, abstention option and option order with every result. They should report both string-match and judged scores for open-ended answers and validate the judge. They should report NOTA recall and listed accuracy separately, together with over-commitment and false NOTA, because neither NOTA recall nor listed accuracy depends on the NOTA rate, whereas their average does. Finally, they should use paired tests over datasets and avoid macro averages in which a dataset contributes a single question.

\subsection{Next steps}
\label{sec:next_experiments}
Several extensions follow from these results. Closed-source models evaluated under the same protocol, and supervised models trained on full training splits rather than a few images per class, would provide further reference points. Lightweight adaptation, such as few-shot prompting and parameter-efficient fine-tuning (PEFT/LoRA), would quantify how much domain performance can be recovered without losing generality. Controlled context (crop, phenology, region and imaging conditions) would show whether it reduces open-ended ambiguity, especially for species identification. \newtext{Auditing the LLM judge against independent ratings of a random sample of the full benchmark, by several annotators and across all eight models,} and comparing judges would give judged scores, including Gemma-4-31B's self-judged scores, known error rates. Finally, agentic systems that can call detection, retrieval or calculation tools should be evaluated, since localization and numeric VQA are where current VLMs fail most.

\section{Limitations}
\label{sec:limitations}
This study evaluates zero-shot, open-weight VLMs only. It includes no closed-source models, few-shot prompting or fine-tuned VLMs, and its only supervised reference is a YOLO11 model trained on a few images per class. It therefore characterizes off-the-shelf open models rather than the best achievable performance. In particular, the VLMs were not fine-tuned on the labelled images given to YOLO, for example with low-rank adaptation (LoRA) \cite{hu_lora_2022}. The comparison thus pits zero-shot VLMs against a trained model and cannot show whether a VLM adapted with the same few labels would close the gap. That comparison is needed before concluding that supervised models remain the better choice once a few labels are available. Each model was run in a single configuration (one quantization, one reasoning-effort level and the sampling parameters from its model card), and other settings could change individual results. Nemotron-3.5-Bolt in particular was run as a heavily quantized 24\,GB GGUF through llama.cpp, so its results may understate the full-precision model.

Datasets were subsampled (up to five images per class for classification), which keeps 220 classification datasets tractable but makes per-dataset estimates noisy. VQA results rest on 13 datasets, two of which contribute a single question to a question format (Section~\ref{sec:results_stability}). The datasets are not uniformly distributed geographically and may not represent the diversity of global agricultural conditions, imaging environments or crop varieties. All prompts were in English. Detection and segmentation are scored by box matching at a single IoU threshold, and segmentation is reduced to boxes around mask components, which does not measure mask quality.

\newtext{The LLM judge was checked against human labels only on a non-random subsample of one model's answers from an earlier run, by a single annotator, and its rules were written after reviewing errors on the same answers (Appendix~\ref{app:judge}). Its error rate on the full benchmark is therefore unknown.} It is also one of the evaluated models, so Gemma-4-31B's judged scores are self-judged. Six classification datasets name their classes only by numbers (e.g., DeepWeeds and CT-PestNet), so open-ended answers on them can hardly be judged correct, which lowers every model's OEQ average. Experiments 2--4 exclude Nemotron and use a single seed, so their effects are compared against the seed-to-seed variation of Experiment 1 rather than averaged over seeds. Finally, batched GPU inference is not bit-reproducible, so exact outputs cannot be regenerated, although aggregate results are stable.

\section{Conclusion}
\label{sec:conclusion}
We benchmarked eight open-weight VLMs on 372 agricultural datasets spanning classification, detection, segmentation and VQA. Current off-the-shelf open VLMs are not yet reliable for agricultural perception. The strongest models reach only about 55\% on multiple-choice classification, at most 36\% on open-ended classification and at most 25--31\% on detection and segmentation, and no single model is best on every task. How the question is asked matters as much as which model answers it. Listing the options raises accuracy substantially above the open-ended setting in which users actually ask, and that gain depends on how the options are built. Models rarely recognize when the correct option is missing, and changing the option set or the scoring of the same answers moves accuracy by several points relative to the default and can reorder models. Multiple-choice scores should therefore be read with caution and reported with their design. Aggregate results are reproducible across seeds and option orders, but individual answers are not, which matters for a grower asking about a single image. A supervised YOLO11 model trained on only 10 images per class already outperforms the best VLM on every task it covers, so VLMs are not yet a replacement for even a small supervised model. Current VLMs are thus best seen as assistive tools whose open-ended answers still need expert checking. Progress will require domain-adapted evaluation protocols, lightweight adaptation and interfaces that combine visual reasoning with structured agronomic knowledge.

\section*{Code and data availability}
The benchmark code, prompt templates, LLM-judge prompts, model configurations, the exact sampled items and all reported scores and statistics are available at \repo. A maintained version of the benchmark for evaluating new models, \code{agml-bench}, is available at \url{https://github.com/Project-AgML/agml-bench}. The underlying datasets are available through AgML (\agmlurl). Appendix~\ref{app:repro} maps every table, figure and number to the data and commands that produce it.

\section*{Use of AI assistance}
The analysis code, the scripts that produce the tables and figures, and the text of this manuscript were developed with the help of an AI coding assistant (Claude Code, Anthropic), working under the authors' direction on the authors' data and compute. So that the results can be checked independently of how the article was written, every reported number, table and figure is generated by scripts in the repository from the run outputs, and Appendix~\ref{app:repro} gives the command behind each one. The authors take responsibility for the content.

{\small
\bibliographystyle{plain}
\bibliography{main}
}

\appendix
\setcounter{table}{0}
\setcounter{figure}{0}
\renewcommand{\thetable}{A\arabic{table}}
\renewcommand{\thefigure}{A\arabic{figure}}

\section{Supplementary methods}
\label{app:methods}

\subsection{Terms and experiments}
\label{app:terms}
Table~\ref{tab:glossary} defines the formats and metrics used in the paper, and Table~\ref{tab:experiments} summarizes the four experiments.

\begin{table*}[!tbp]
\centering
\caption{Formats and metrics.}
\label{tab:glossary}
\renewcommand{\arraystretch}{1.25}
\tablefont
\begin{tabular}{p{3.3cm} p{12.2cm}}
\toprule
\textbf{Term} & \textbf{Definition} \\
\midrule
MCQA within & Multiple-choice classification with up to four classes from the item's own dataset, followed by NOTA and IDK. \\
MCQA across & As MCQA within, but the distractors come from other datasets with the same agricultural task. \\
OEQ & Open-ended classification question answered in one sentence, graded by the LLM judge. \\
VQA MC / other & VQA multiple-choice questions / yes-no, numeric and open-ended VQA questions. \\
Withheld-class item & MCQA item whose true class is removed from the options (30\% of items by default); NOTA is correct. \\
NOTA recall & Share of withheld-class items answered ``None of the above'' (higher is better). \\
Over-commit rate & Share of withheld-class items answered with one of the listed, all incorrect, classes (lower is better). \\
Listed accuracy & Share of items whose true class is listed that are answered with the true class; any other answer, including NOTA and IDK, is wrong (higher is better). \\
False-NOTA rate & Share of items whose true class is listed that are answered ``None of the above'' (lower is better). \\
IDK rate & Share of items answered ``I don't know''; IDK is always scored as incorrect. \\
Macro accuracy & Mean of per-dataset accuracies, so that every dataset counts equally. \\
Localization accuracy & Per-image $\text{TP}/(\text{TP}+\text{FP}+\text{FN})$ with boxes matched at IoU $\geq 0.5$, averaged over images and then datasets. \\
Agreement (Same) & Share of items with the same chosen option (MCQA, VQA MC) or the same score (other formats) in every seed or option order. \\
Paired bootstrap & Test of the difference between two models. Per-dataset differences in accuracy are resampled over datasets; the p-value is twice the share of resamples in which the mean difference changes sign. \\
Holm correction & Adjustment of the p-values of several comparisons made together (e.g., the 28 model pairs of a format) so that the probability of any false positive stays at 5\%; $p_{\text{Holm}}$ is the adjusted p-value. \\
Top-1 accuracy & Share of images for which the YOLO11 baseline's most probable class, among the dataset's classes, is the true class. \\
Position preference & For each option position, the share of answers at that position minus the share of items whose correct answer is there (points). \\
$^\dagger$ & Self-judged, because Gemma-4-31B is both an evaluated model and the LLM judge. \\
\bottomrule
\end{tabular}
\end{table*}

\begin{table*}[!tbp]
\centering
\caption{Experiments. Each changes one variable from the default setting (30\% NOTA rate, IDK offered, NOTA and IDK last, options in one seeded order) and keeps the evaluation items fixed. Experiments 2--4 are compared item by item with the seed-0 run of Experiment 1.}
\label{tab:experiments}
\renewcommand{\arraystretch}{1.2}
\tablefont
\begin{tabular}{llllll}
\toprule
\textbf{Experiment} & \textbf{Variable} & \textbf{Levels} & \textbf{Models} & \textbf{Seeds} & \textbf{Formats} \\
\midrule
1 & Generation seed & 0, 1, 2 & 8 & 3 & all \\
2 & Option order & \makecell[l]{5 random orders; NOTA and\\IDK at random positions} & 7 & 0 & MCQA within, across; VQA MC \\
3 & NOTA rate & 10\%, 50\% (default 30\%) & 7 & 0 & MCQA within, across; VQA MC \\
4 & IDK option & removed (default offered) & 7 & 0 & MCQA within, across; VQA MC \\
\bottomrule
\end{tabular}
\end{table*}

\subsection{Inference settings}
\label{app:inference}
Every run used one NVIDIA RTX 6000 Ada GPU (48\,GB) with an FP8 key-value cache (8-bit for llama.cpp), a 32K-token context and a limit of 16,384 output tokens per answer. Sampling parameters (temperature, top-$p$, top-$k$) follow each model's card. Every model that supports a reasoning trace was run with it enabled at a low effort setting, meaning the model's own low setting where one exists (Qwen3.8, Muse-Glimmer, Nemotron), and otherwise a 1,024-token cap on the reasoning trace (both Gemma models, Qwen3.6 and MiniCPM). Ministral-3 has no native reasoning mode, so it receives a system prompt asking it to reason inside \code{<think>} tags and to put its final answer inside \code{<answer>} tags. The reasoning trace is removed before the answer is scored. In a preliminary comparison on 120 items per model, accuracy did not differ between off, low, medium and high effort beyond noise, while low effort used about 30\% less GPU time than an uncapped trace.

\subsection{Dataset sampling and agricultural tasks}
\label{app:sampling}
Each dataset was subsampled with a fixed seed, preferring its test split, then its validation split, then its training split, and the number of items kept depends on the task.
\begin{itemize}
    \item Classification datasets keep up to five images per class.
    \item For detection, images are added, rarest class first, until every class appears in at least five selected images.
    \item Segmentation uses the same rule, with the classes present in each mask.
    \item VQA questions are stratified by the dataset's own metadata (e.g., task, question type, crop) and by question format (multiple choice, yes/no, numeric, open-ended), with at least five questions per stratum and at most 250 per dataset.
\end{itemize}
Agricultural tasks are assigned from each dataset's task type and a curated per-class override list, so a dataset whose classes span several tasks counts in each. The crop shown in an image is taken from the sample's metadata, the dataset or the class label. The exact sampled items are listed in the repository (\code{manifests/}).

\subsection{Answer matching and localization scoring}
\label{app:matching}
After any reasoning trace is removed (and the content of an \code{<answer>} tag is used when present), a multiple-choice answer is matched to an option by exact match, then by the longest option named in the answer, then by fuzzy similarity of at least 0.6, where similarity is the maximum of the character-level ratio, the token-set ratio and the partial ratio. The same similarity rule gives the fuzzy-match score of OEQ answers against their reference class. VQA multiple-choice answers are parsed as option letters, yes/no answers as yes or no, and numeric answers as numbers. Answers that are empty or cut off at the token limit are scored as wrong.

For detection and segmentation, a predicted box is a true positive when it matches an unmatched ground-truth box at IoU $\geq 0.5$, with the labels also required to agree when the dataset has more than one class. Segmentation ground-truth boxes are the connected components of each class mask that cover at least 0.02\% of the image. ROUGE-L for open-ended VQA is the F1 of the longest common token subsequence between answer and reference.

\subsection{LLM judge}
\label{app:judge}
\begin{figure*}[!tbp]
    \centering
    \includegraphics[width=0.85\textwidth]{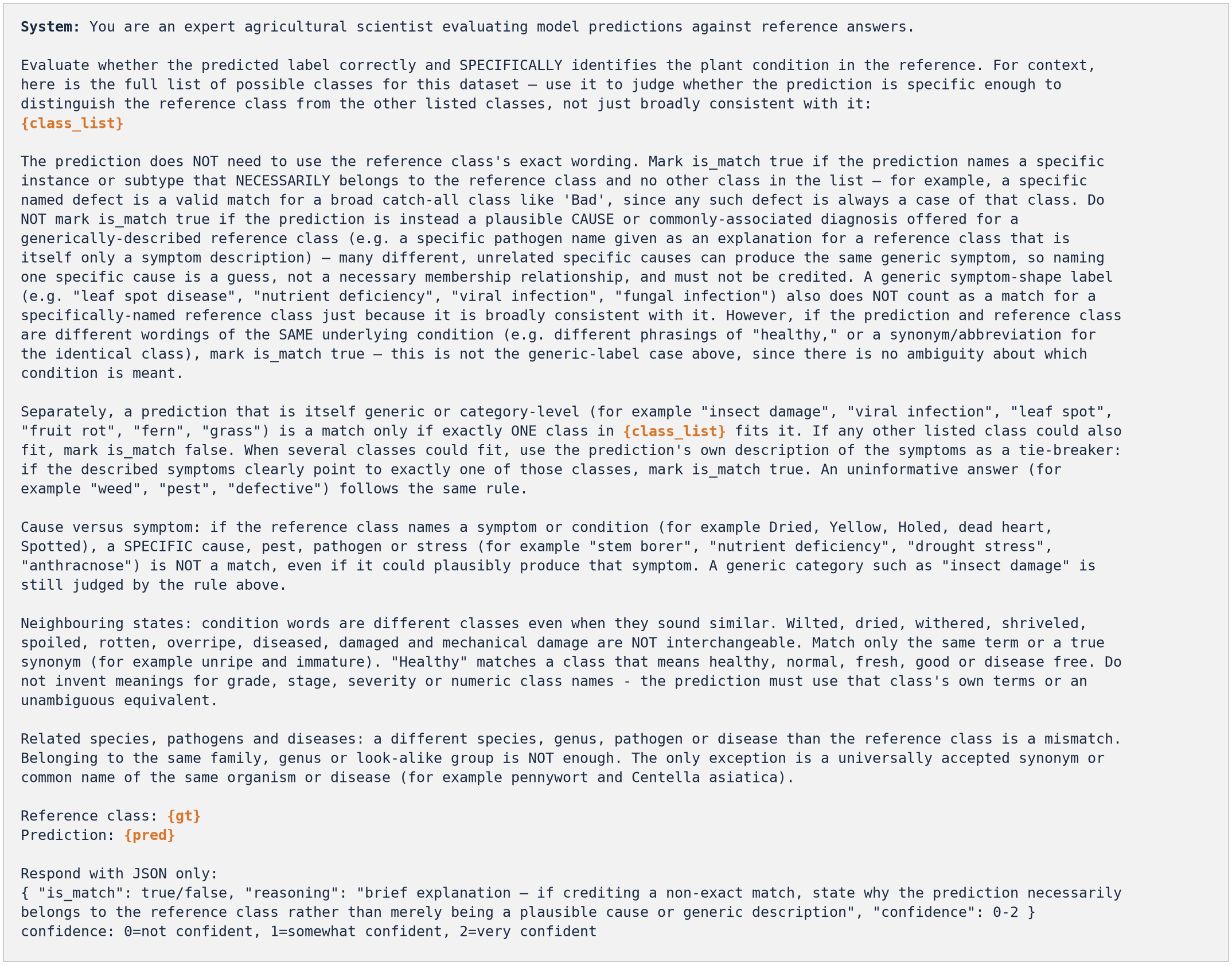}
    \caption{LLM-judge prompt for OEQ answers, made of the system prompt followed by the classification judge prompt. \{class\_list\} is filled with the dataset's full class list, \{gt\} with the reference class and \{pred\} with the model's answer. The judge returns JSON with \code{is\_match}, a brief \code{reasoning} and a 0--2 \code{confidence}. Line breaks between rule groups were added for readability; the text is otherwise verbatim.}
    \label{fig:judge_prompt}
\end{figure*}

The judge is \texttt{\nolinkurl{RedHatAI/gemma-4-31B-it-FP8-dynamic}} with its reasoning trace off and temperature 0. Its system prompt casts it as an expert agricultural scientist, and it returns a JSON verdict (\code{is\_match}, a brief reasoning and a 0--2 confidence). The OEQ judge prompt (Figure~\ref{fig:judge_prompt}) was refined over several iterations until its verdicts matched a set of reference (oracle) judgments. The judge receives the dataset's full class list and marks a prediction correct only if it identifies the reference class specifically enough to rule out every other listed class, and four rules make this concrete.
\begin{itemize}
    \item A specific instance that necessarily belongs to the reference class counts (e.g., a named defect for a catch-all ``Bad'' class), as do synonyms and different wordings of the same condition (e.g., pennywort for \emph{Centella asiatica}; ``healthy'' for normal, fresh or disease-free).
    \item A generic label (e.g., ``insect damage'', ``viral infection'') counts only if exactly one listed class fits it; when several do, the symptoms described in the answer may single one out.
    \item A specific cause, pest or pathogen offered for a class that names a symptom (e.g., ``stem borer'' for ``dead heart'') does not count, because many causes produce the same symptom.
    \item Neighbouring condition states (wilted, dried, shriveled, spoiled, rotten, overripe, diseased, damaged) are not interchangeable, a different species, genus or pathogen is a mismatch even within the same family, and the judge may not infer meanings for grade, stage, severity or numeric class names.
\end{itemize}
For open-ended VQA, the answer's key facts and conclusion must agree with the reference answer without contradicting it, while wording and level of detail may differ.

\newtext{We checked the judge against human labels on a subsample of open-ended classification answers from an earlier run of the benchmark (8,324 answers), all given by one model, Gemma-4-31B, which is also the judge. The subsample is neither the evaluation set of this paper nor random, and it has two parts. The disagreement set holds every answer on which the successive judge prompts (v1 to v3) gave different verdicts, 308 answers of which 282 were labelled. The agreement sample holds 150 answers on which the first two prompts agreed, 75 that both credited and 75 that both rejected. One annotator labelled each answer against the rules above, without seeing the judge's verdicts on the agreement sample. The prompt versions differ as follows. v1 is the original template. v2 also accepts a generic answer when its description points to one class. v3 accepts a generic answer only when exactly one class fits it and forbids inventing meanings for grade or stage classes. v4, the final prompt, adds the rules on causes versus symptoms, neighbouring states and related species or diseases. On the disagreement set the final prompt agrees with the human label on 79\% of answers, against 49\% for v1 and 74\% for v3, and on the agreement sample it agrees on 95\%, with 7 false credits and 1 false rejection (Table~\ref{tab:judge_validation}). These figures describe the prompt on this subsample only. They do not estimate its accuracy on the full benchmark, because the subsample is not random and comes from one model, and because the v4 rules were written after reviewing errors on these same answers, so agreement on them is not an independent test.}

\begin{table}[!tbp]
\centering
{
\caption{Agreement of the LLM-judge prompt versions with human labels on a labelled subsample of Gemma-4-31B's OEQ answers from an earlier run, not the evaluation set of this paper (Appendix~\ref{app:judge}). Cells are the answers on which the judge matches the human label (\% of the row's answers in parentheses). The disagreement set holds the answers on which prompts v1 to v3 disagreed. The agreement sample holds answers on which v1 and v2 agreed, half credited and half rejected, so v1 and v2 match on it by construction. The v4 rules were written after reviewing errors on these answers, so its agreement is not an independent test.}
\label{tab:judge_validation}
\renewcommand{\arraystretch}{1.2}
\tablefont
\begin{tabular}{llrrrrr}
\toprule
\textbf{Set} & \textbf{Task} & \textbf{Answers} & \textbf{v1} & \textbf{v2} & \textbf{v3} & \textbf{v4 (final)} \\
\midrule
Disagreement set & Disease & $65$ & $31$ ($48$) & $35$ ($54$) & $37$ ($57$) & $44$ ($68$) \\
 & Pest/damage & $50$ & $7$ ($14$) & $41$ ($82$) & $42$ ($84$) & $41$ ($82$) \\
 & Species & $81$ & $58$ ($72$) & $19$ ($23$) & $61$ ($75$) & $62$ ($77$) \\
 & Quality/maturity & $86$ & $42$ ($49$) & $6$ ($7$) & $70$ ($81$) & $77$ ($90$) \\
 & \textbf{All} & $\mathbf{282}$ & $\mathbf{138}$ ($\mathbf{49}$) & $\mathbf{101}$ ($\mathbf{36}$) & $\mathbf{210}$ ($\mathbf{74}$) & $\mathbf{224}$ ($\mathbf{79}$) \\
\midrule
Agreement sample & Matches human label & $150$ & $136$ ($91$) & $136$ ($91$) & $141$ ($94$) & $142$ ($95$) \\
 & False credits & & $13$ & $13$ & $8$ & $7$ \\
 & False rejections & & $1$ & $1$ & $1$ & $1$ \\
\bottomrule
\end{tabular}

}
\end{table}

\section{Full results tables}
\label{app:tables}
\setcounter{table}{0}
\renewcommand{\thetable}{B\arabic{table}}
\setcounter{figure}{0}
\renewcommand{\thefigure}{B\arabic{figure}}
This appendix gives the exact values behind the results in Section~\ref{sec:results}, the stability of those results across seeds and option orders, and the few-shot YOLO11 baseline at other training-set sizes.

\subsection{Main results}
\label{app:tables_main}
Tables~\ref{tab:main_results}, \ref{tab:agtask} and~\ref{tab:behaviour} give the exact values plotted in Figures~\ref{fig:main_results_chart}, \ref{fig:agtask_chart} and~\ref{fig:behaviour}, and Table~\ref{tab:judge} gives OEQ accuracy under fuzzy matching and the LLM judge for each model, overall and by task.

\begin{table*}[!tbp]
\centering
\caption{Accuracy (\%) by task, as the mean over generation seeds of the macro average over datasets, with 95\% bootstrap confidence intervals over datasets in brackets. $n$ is the number of datasets. OEQ and open-ended VQA answers are graded by the LLM judge. VQA accuracy is the macro average over datasets across all question formats. Best value per column in bold (VLMs only). The chance baseline uses no image, and the YOLO11 baseline sees only the image, so its classification score is one top-1 accuracy across the three formats (Section~\ref{sec:baselines}). $^\dagger$Self-judged on OEQ and open-ended VQA.}
\label{tab:main_results}
\renewcommand{\arraystretch}{1.2}
\tablefont
\setlength{\tabcolsep}{5pt}
\begin{tabular}{lcccccc}
\toprule
\textbf{Model} & \makecell{\textbf{MCQA across}\\($n=220$)} & \makecell{\textbf{MCQA within}\\($n=220$)} & \makecell{\textbf{OEQ (judged)}\\($n=220$)} & \makecell{\textbf{VQA}\\($n=13$)} & \makecell{\textbf{Detection}\\($n=105$)} & \makecell{\textbf{Segmentation}\\($n=34$)} \\
\midrule
Qwen3.6-35B-A3B & \makecell{$54.7$\\{[52.2, 57.2]}} & \makecell{$47.6$\\{[45.1, 50.2]}} & \makecell{$27.0$\\{[24.1, 29.9]}} & \makecell{$49.7$\\{[40.9, 58.1]}} & \makecell{$23.0$\\{[18.7, 27.3]}} & \makecell{$30.0$\\{[20.6, 40.3]}} \\
Qwen3.8-27B & \makecell{$52.1$\\{[49.7, 54.6]}} & \makecell{$46.4$\\{[43.8, 49.1]}} & \makecell{$28.1$\\{[25.1, 31.0]}} & \makecell{$53.9$\\{[46.1, 61.6]}} & \makecell{$\mathbf{24.7}$\\{[20.6, 28.9]}} & \makecell{$28.6$\\{[20.0, 38.2]}} \\
Gemma-4-31B$^\dagger$ & \makecell{$\mathbf{54.9}$\\{[52.2, 57.6]}} & \makecell{$48.2$\\{[45.3, 51.2]}} & \makecell{$29.1$\\{[26.0, 32.3]}} & \makecell{$52.2$\\{[44.4, 59.9]}} & \makecell{$23.4$\\{[19.0, 28.0]}} & \makecell{$30.5$\\{[21.3, 40.7]}} \\
Gemma-4-12B & \makecell{$49.4$\\{[47.0, 51.9]}} & \makecell{$43.9$\\{[41.5, 46.5]}} & \makecell{$25.0$\\{[22.1, 28.0]}} & \makecell{$45.4$\\{[38.6, 52.0]}} & \makecell{$21.3$\\{[17.2, 25.5]}} & \makecell{$\mathbf{30.6}$\\{[21.4, 40.6]}} \\
Muse-Glimmer-30B & \makecell{$54.4$\\{[51.8, 56.9]}} & \makecell{$\mathbf{49.7}$\\{[46.9, 52.6]}} & \makecell{$\mathbf{35.6}$\\{[32.4, 38.8]}} & \makecell{$\mathbf{59.7}$\\{[49.8, 69.5]}} & \makecell{$15.9$\\{[12.2, 19.8]}} & \makecell{$24.2$\\{[15.5, 33.9]}} \\
Nemotron-3.5-Bolt-74B-A9B & \makecell{$47.7$\\{[45.2, 50.2]}} & \makecell{$40.8$\\{[38.4, 43.3]}} & \makecell{$25.7$\\{[22.8, 28.5]}} & \makecell{$46.1$\\{[38.9, 53.0]}} & \makecell{$21.0$\\{[17.1, 24.9]}} & \makecell{$29.8$\\{[21.1, 39.4]}} \\
MiniCPM-V-4.7-35B-A3B & \makecell{$37.3$\\{[34.9, 39.8]}} & \makecell{$41.4$\\{[39.1, 43.8]}} & \makecell{$25.3$\\{[22.3, 28.5]}} & \makecell{$46.2$\\{[37.7, 54.9]}} & \makecell{$17.5$\\{[13.8, 21.4]}} & \makecell{$25.4$\\{[16.8, 34.9]}} \\
Ministral-3-14B & \makecell{$41.2$\\{[39.1, 43.4]}} & \makecell{$37.6$\\{[35.7, 39.7]}} & \makecell{$14.4$\\{[12.0, 16.9]}} & \makecell{$36.8$\\{[30.0, 43.9]}} & \makecell{$8.5$\\{[5.8, 11.4]}} & \makecell{$12.5$\\{[6.9, 19.0]}} \\
\midrule
\textit{Chance (random answer)} & $16.9$ & $17.7$ & $20.8$ & $9.4$ & $2.0$ & $5.8$ \\
\textit{YOLO11s, 10 images per class} & \multicolumn{3}{c}{\makecell{top-1 $77.3$\\{[75.0, 79.5]}}} & --- & \makecell{$37.5$\\{[33.4, 41.5]}} & \makecell{$51.4$\\{[42.2, 60.5]}} \\
\bottomrule
\end{tabular}

\end{table*}

\begin{table*}[!tbp]
\centering
\caption{Classification accuracy (\%) by agricultural task, as the mean over generation seeds of the macro average over datasets. ``Pest'' stands for pest/damage. The number of datasets per task is in parentheses (a dataset whose classes span several tasks counts in each). Best value per column in bold. The last row is the top-1 accuracy of the YOLO11 baseline trained on 10 images per class (Section~\ref{sec:baselines}), which sees only the image and so has the same value under every format. $^\dagger$Self-judged on OEQ.}
\label{tab:agtask}
\renewcommand{\arraystretch}{1.2}
\tablefont
\setlength{\tabcolsep}{3.5pt}
\begin{tabular}{lcccccccccccc}
\toprule
\multirow{2}{*}{\textbf{Model}} & \multicolumn{4}{c}{\textbf{MCQA across}} & \multicolumn{4}{c}{\textbf{MCQA within}} & \multicolumn{4}{c}{\textbf{OEQ (judged)}} \\
\cmidrule(lr){2-5} \cmidrule(lr){6-9} \cmidrule(lr){10-13}
 & \makecell{Disease\\(124)} & \makecell{Pest\\(62)} & \makecell{Species\\(59)} & \makecell{Quality\\(25)} & \makecell{Disease\\(124)} & \makecell{Pest\\(62)} & \makecell{Species\\(59)} & \makecell{Quality\\(25)} & \makecell{Disease\\(124)} & \makecell{Pest\\(62)} & \makecell{Species\\(59)} & \makecell{Quality\\(25)} \\
\midrule
Qwen3.6-35B-A3B & $57.7$ & $55.1$ & $58.0$ & $38.0$ & $51.4$ & $47.5$ & $45.0$ & $42.0$ & $30.1$ & $30.0$ & $16.9$ & $39.5$ \\
Qwen3.8-27B & $54.9$ & $53.7$ & $54.6$ & $37.1$ & $51.1$ & $43.6$ & $42.1$ & $\mathbf{42.9}$ & $32.7$ & $30.8$ & $14.4$ & $\mathbf{42.8}$ \\
Gemma-4-31B$^\dagger$ & $\mathbf{58.2}$ & $55.9$ & $57.4$ & $\mathbf{38.6}$ & $\mathbf{54.3}$ & $50.3$ & $41.5$ & $41.1$ & $35.8$ & $26.7$ & $14.3$ & $41.1$ \\
Gemma-4-12B & $54.2$ & $47.1$ & $48.5$ & $37.9$ & $50.1$ & $38.9$ & $38.0$ & $39.9$ & $30.9$ & $22.7$ & $10.5$ & $38.9$ \\
Muse-Glimmer-30B & $55.7$ & $\mathbf{57.6}$ & $\mathbf{58.6}$ & $38.5$ & $52.8$ & $\mathbf{55.7}$ & $\mathbf{48.8}$ & $42.2$ & $\mathbf{39.4}$ & $\mathbf{38.3}$ & $\mathbf{27.2}$ & $42.1$ \\
Nemotron-3.5-Bolt-74B-A9B & $52.7$ & $48.1$ & $45.0$ & $36.2$ & $45.2$ & $39.4$ & $36.1$ & $37.2$ & $31.9$ & $25.1$ & $10.6$ & $37.2$ \\
MiniCPM-V-4.7-35B-A3B & $38.0$ & $35.8$ & $40.1$ & $36.1$ & $45.2$ & $44.0$ & $36.0$ & $40.3$ & $28.9$ & $24.7$ & $15.3$ & $39.9$ \\
Ministral-3-14B & $44.6$ & $39.2$ & $38.2$ & $34.3$ & $41.7$ & $28.7$ & $32.7$ & $38.5$ & $14.1$ & $11.3$ & $7.7$ & $34.1$ \\
\midrule
Mean of models & $52.0$ & $49.1$ & $50.0$ & $37.1$ & $49.0$ & $43.5$ & $40.0$ & $40.5$ & $30.5$ & $26.2$ & $14.6$ & $39.5$ \\
\textit{YOLO11s top-1} & $78.0$ & $72.9$ & $79.5$ & $73.9$ & $78.0$ & $72.9$ & $79.5$ & $73.9$ & $78.0$ & $72.9$ & $79.5$ & $73.9$ \\
\bottomrule
\end{tabular}

\end{table*}

\begin{table*}[!tbp]
\centering
\caption{Multiple-choice behaviour at the default options (seed 0), macro over datasets (\%). In 30\% of MCQA items the true class is withheld, so ``None of the above'' (NOTA) is correct; in the other 70\% it is listed. The columns use different item sets. On withheld-class items, NOTA rec.\ (recall, higher is better) is the share answered NOTA and Over-commit (lower is better) the share answered with one of the listed, all incorrect, classes; the two sum to 100\% minus the share of those items answered ``I don't know''. On items whose true class is listed, Listed acc.\ (higher is better) is the share answered with the true class, and False NOTA (lower is better) the share answered NOTA, i.e., a correct option rejected; the remaining items are answered with a wrong class or IDK. Within each dataset, MCQA accuracy is the withheld share times NOTA rec.\ plus the listed share times Listed acc.; for these macro averages that is approximately 0.3 $\times$ NOTA rec.\ $+$ 0.7 $\times$ Listed acc.\ (Section~\ref{sec:results_design}). IDK is the share of all items answered ``I don't know''. A model that answers NOTA more often gains recall but also rejects more correct options. MCQA within draws its distractors from the same dataset and MCQA across from other datasets.}
\label{tab:behaviour}
\renewcommand{\arraystretch}{1.2}
\tablefont
\setlength{\tabcolsep}{4pt}
\begin{tabular}{lcccccccccc}
\toprule
\multirow{2}{*}{\textbf{Model}} & \multicolumn{5}{c}{\textbf{MCQA within}} & \multicolumn{5}{c}{\textbf{MCQA across}} \\
\cmidrule(lr){2-6} \cmidrule(lr){7-11}
 & \makecell{NOTA\\rec.} & \makecell{Over-\\commit} & \makecell{Listed\\acc.} & \makecell{False\\NOTA} & IDK & \makecell{NOTA\\rec.} & \makecell{Over-\\commit} & \makecell{Listed\\acc.} & \makecell{False\\NOTA} & IDK \\
\midrule
Qwen3.6-35B-A3B & $42.3$ & $57.7$ & $49.1$ & $18.2$ & $0.0$ & $57.1$ & $42.9$ & $53.4$ & $25.5$ & $0.0$ \\
Qwen3.8-27B & $42.1$ & $56.4$ & $48.0$ & $16.2$ & $1.8$ & $57.2$ & $42.5$ & $51.9$ & $25.8$ & $0.2$ \\
Gemma-4-31B$^\dagger$ & $45.5$ & $47.7$ & $49.4$ & $15.4$ & $8.4$ & $62.3$ & $35.6$ & $51.7$ & $28.0$ & $2.2$ \\
Gemma-4-12B & $37.5$ & $60.2$ & $46.5$ & $15.1$ & $2.6$ & $54.3$ & $45.2$ & $47.2$ & $28.1$ & $0.4$ \\
Muse-Glimmer-30B & $62.4$ & $31.8$ & $43.3$ & $29.5$ & $6.7$ & $76.0$ & $22.8$ & $45.3$ & $42.1$ & $1.4$ \\
Nemotron-3.5-Bolt-74B-A9B & $35.0$ & $58.5$ & $43.7$ & $14.8$ & $6.3$ & $48.7$ & $48.6$ & $47.6$ & $25.0$ & $2.6$ \\
MiniCPM-V-4.7-35B-A3B & $49.6$ & $49.8$ & $37.1$ & $25.3$ & $0.9$ & $41.3$ & $58.4$ & $35.5$ & $25.7$ & $0.2$ \\
Ministral-3-14B & $52.2$ & $39.9$ & $31.4$ & $32.5$ & $6.6$ & $66.5$ & $29.8$ & $29.8$ & $49.4$ & $4.2$ \\
\bottomrule
\end{tabular}

\end{table*}

\begin{table*}[!tbp]
\centering
\caption{OEQ accuracy (\%, seed 0) scored by fuzzy string matching against the reference class vs.\ by the LLM judge, macro over datasets, overall and by agricultural task. $\Delta$ = Judge $-$ Fuzzy (overall; per task, subtract the two columns). Figure~\ref{fig:agtask_chart} plots the same comparison by task, averaged over generation seeds. $^\dagger$Self-judged.}
\label{tab:judge}
\renewcommand{\arraystretch}{1.2}
\tablefont
\setlength{\tabcolsep}{4pt}
\begin{tabular}{lccccccccccc}
\toprule
\multirow{2}{*}{\textbf{Model}} & \multicolumn{3}{c}{\textbf{All}} & \multicolumn{2}{c}{\textbf{Disease}} & \multicolumn{2}{c}{\textbf{Pest/Damage}} & \multicolumn{2}{c}{\textbf{Species}} & \multicolumn{2}{c}{\textbf{Quality}} \\
\cmidrule(lr){2-4} \cmidrule(lr){5-6} \cmidrule(lr){7-8} \cmidrule(lr){9-10} \cmidrule(lr){11-12}
 & Fuzzy & Judge & $\Delta$ & Fuzzy & Judge & Fuzzy & Judge & Fuzzy & Judge & Fuzzy & Judge \\
\midrule
Qwen3.6-35B-A3B & $41.9$ & $27.7$ & $-14.2$ & $50.9$ & $30.5$ & $41.1$ & $28.6$ & $28.3$ & $16.9$ & $33.4$ & $38.1$ \\
Qwen3.8-27B & $40.6$ & $29.7$ & $-10.9$ & $51.9$ & $33.1$ & $36.5$ & $32.4$ & $24.0$ & $14.0$ & $35.4$ & $44.2$ \\
Gemma-4-31B$^\dagger$ & $34.2$ & $29.5$ & $-4.7$ & $41.4$ & $36.5$ & $33.3$ & $25.6$ & $21.9$ & $13.9$ & $31.1$ & $42.3$ \\
Gemma-4-12B & $31.8$ & $25.0$ & $-6.9$ & $38.3$ & $30.7$ & $31.2$ & $22.3$ & $20.0$ & $10.9$ & $30.7$ & $37.6$ \\
Muse-Glimmer-30B & $45.8$ & $36.4$ & $-9.4$ & $55.4$ & $39.1$ & $42.7$ & $37.8$ & $32.0$ & $26.9$ & $39.8$ & $42.2$ \\
Nemotron-3.5-Bolt-74B-A9B & $39.8$ & $26.3$ & $-13.4$ & $49.8$ & $31.8$ & $38.6$ & $26.6$ & $21.6$ & $10.7$ & $37.0$ & $35.9$ \\
MiniCPM-V-4.7-35B-A3B & $43.6$ & $26.2$ & $-17.4$ & $52.8$ & $29.1$ & $43.9$ & $25.6$ & $28.5$ & $15.1$ & $34.5$ & $40.1$ \\
Ministral-3-14B & $29.3$ & $13.8$ & $-15.6$ & $35.8$ & $14.3$ & $26.8$ & $10.4$ & $16.9$ & $7.8$ & $33.9$ & $33.6$ \\
\bottomrule
\end{tabular}

\end{table*}

\subsection{Stability across seeds and option orders}
\label{app:tables_stability}
Table~\ref{tab:seeds} gives the variation across generation seeds (Section~\ref{sec:results_stability}), and Table~\ref{tab:ablation_order} the results across option orders (Experiment~2).

\begin{table*}[!tbp]
\centering
\caption{Generation-seed stability (Experiment 1, default options, 3 seeds per model). Each cell gives SD / agr., the standard deviation (percentage points) of the macro accuracy across seeds and the share (\%) of items with the same result in every seed. For MCQA and VQA MC the result is the chosen option; for OEQ, VQA other, detection and segmentation it is the item's score, so agreement is not directly comparable between these two groups. VQA MC covers the multiple-choice questions and VQA other the yes/no, numeric and open-ended ones. $^\dagger$Self-judged.}
\label{tab:seeds}
\renewcommand{\arraystretch}{1.2}
\tablefont
\begin{tabular}{lccccccc}
\toprule
\multirow{2}{*}{\textbf{Model}} & \textbf{MCQA across} & \textbf{MCQA within} & \textbf{OEQ} & \textbf{VQA MC} & \textbf{VQA other} & \textbf{Det.} & \textbf{Seg.} \\
 & SD / agr. & SD / agr. & SD / agr. & SD / agr. & SD / agr. & SD / agr. & SD / agr. \\
\midrule
Qwen3.6-35B-A3B & $0.32$ / $62$ & $0.63$ / $56$ & $0.50$ / $82$ & $8.26$ / $64$ & $0.52$ / $76$ & $0.32$ / $85$ & $0.56$ / $89$ \\
Qwen3.8-27B & $1.12$ / $68$ & $0.41$ / $61$ & $0.36$ / $84$ & $0.45$ / $65$ & $0.08$ / $75$ & $0.24$ / $87$ & $0.58$ / $91$ \\
Gemma-4-31B$^\dagger$ & $0.18$ / $78$ & $0.08$ / $70$ & $0.29$ / $90$ & $0.79$ / $71$ & $0.09$ / $80$ & $0.70$ / $92$ & $0.22$ / $92$ \\
Gemma-4-12B & $0.55$ / $64$ & $0.19$ / $57$ & $0.13$ / $87$ & $8.20$ / $59$ & $0.54$ / $74$ & $0.49$ / $92$ & $0.43$ / $95$ \\
Muse-Glimmer-30B & $0.12$ / $84$ & $0.33$ / $78$ & $0.40$ / $88$ & $0.42$ / $77$ & $0.08$ / $85$ & $0.28$ / $95$ & $0.94$ / $94$ \\
Nemotron-3.5-Bolt-74B-A9B & $0.26$ / $50$ & $0.43$ / $44$ & $0.18$ / $85$ & $1.54$ / $55$ & $0.20$ / $72$ & $0.21$ / $89$ & $0.53$ / $85$ \\
MiniCPM-V-4.7-35B-A3B & $0.16$ / $73$ & $0.30$ / $72$ & $0.37$ / $92$ & $0.63$ / $57$ & $0.45$ / $87$ & $0.44$ / $95$ & $0.91$ / $94$ \\
Ministral-3-14B & $0.40$ / $68$ & $0.24$ / $59$ & $0.24$ / $95$ & $8.91$ / $61$ & $5.75$ / $81$ & $0.53$ / $97$ & $0.59$ / $93$ \\
\bottomrule
\end{tabular}

\end{table*}

\begin{table*}[!tbp]
\centering
\caption{Option-order randomization (Experiment 2). Each multiple-choice question is re-asked under five seeded option orders, with ``None of the above'' and ``I don't know'' also at random positions. Acc.\ is the mean macro accuracy (\%) over the five orders, SD the standard deviation across orders (points) and Same the share (\%) of items answered with the same option in all five orders. Lower SD and higher Same mean more stable results. Pos.\ is the position preference (Figure~\ref{fig:position}), the largest difference (points) between the share of answers at a position and the share of items whose correct answer is there, with that position in parentheses; positive means the position is chosen too often. With up to 12 options for VQA multiple choice, positions run to 12.}
\label{tab:ablation_order}
\renewcommand{\arraystretch}{1.2}
\tablefont
\setlength{\tabcolsep}{4pt}
\begin{tabular}{lcccccccccccc}
\toprule
\multirow{2}{*}{\textbf{Model}} & \multicolumn{4}{c}{\textbf{MCQA within}} & \multicolumn{4}{c}{\textbf{MCQA across}} & \multicolumn{4}{c}{\textbf{VQA multiple choice}} \\
\cmidrule(lr){2-5} \cmidrule(lr){6-9} \cmidrule(lr){10-13}
 & \makecell{Acc.\\(\%)} & \makecell{SD\\(pts)} & \makecell{Same\\(\%)} & \makecell{Pos.\\(pts)} & \makecell{Acc.\\(\%)} & \makecell{SD\\(pts)} & \makecell{Same\\(\%)} & \makecell{Pos.\\(pts)} & \makecell{Acc.\\(\%)} & \makecell{SD\\(pts)} & \makecell{Same\\(\%)} & \makecell{Pos.\\(pts)} \\
\midrule
Qwen3.6-35B-A3B & $47.2$ & $0.19$ & $44$ & $+0.7$ (1) & $54.3$ & $0.55$ & $50$ & $-0.5$ (2) & $56.7$ & $7.42$ & $52$ & $+2.0$ (3) \\
Qwen3.8-27B & $44.5$ & $0.28$ & $46$ & $+1.2$ (1) & $51.5$ & $0.30$ & $53$ & $-1.2$ (5) & $66.2$ & $0.58$ & $53$ & $+2.5$ (2) \\
Gemma-4-31B$^\dagger$ & $46.9$ & $0.34$ & $57$ & $-1.0$ (5) & $54.9$ & $0.19$ & $64$ & $+0.7$ (2) & $65.4$ & $0.65$ & $60$ & $+1.1$ (2) \\
Gemma-4-12B & $42.6$ & $0.29$ & $42$ & $+1.8$ (1) & $49.6$ & $0.55$ & $50$ & $+0.8$ (1) & $53.7$ & $8.76$ & $50$ & $+1.0$ (3) \\
Muse-Glimmer-30B & $46.9$ & $0.30$ & $56$ & $+1.9$ (2) & $54.7$ & $0.41$ & $63$ & $+1.3$ (3) & $52.6$ & $6.89$ & $62$ & $+1.6$ (3) \\
MiniCPM-V-4.7-35B-A3B & $41.3$ & $0.61$ & $43$ & $-1.1$ (6) & $39.9$ & $0.39$ & $50$ & $-0.8$ (6) & $35.8$ & $1.28$ & $28$ & $+7.2$ (1) \\
Ministral-3-14B & $34.4$ & $0.45$ & $27$ & $-4.8$ (1) & $38.9$ & $0.29$ & $30$ & $-6.0$ (1) & $50.9$ & $0.67$ & $40$ & $-3.0$ (1) \\
\bottomrule
\end{tabular}

\end{table*}

\subsection{Few-shot YOLO11 at 5, 10 and 15 images per class}
\label{app:yolo}

\begin{figure}[!tbp]
\centering
\includegraphics[width=\textwidth]{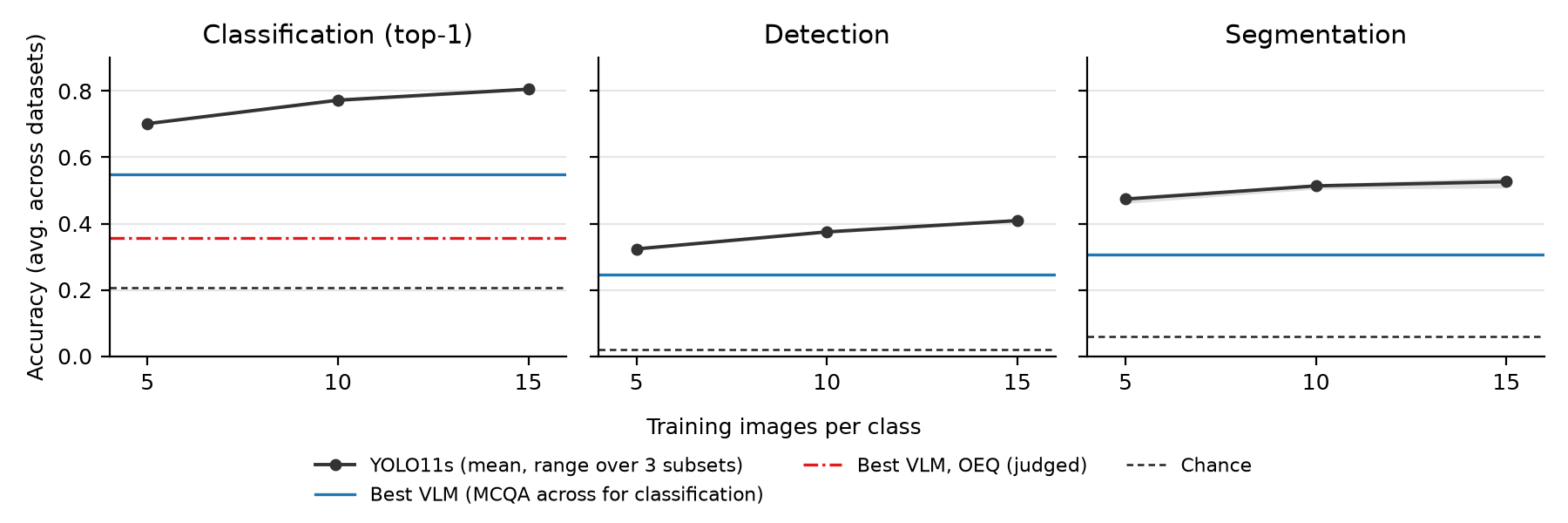}
\caption{Few-shot YOLO11s baseline, one model per task, at 5, 10 and 15 training images per class. Classification is top-1 accuracy (image only). Points are the mean over three training subsets and the shaded band, mostly narrower than the line, is their range. Blue lines mark the best VLM (for classification, under MCQA across) and the red dash-dotted line the best VLM under OEQ (mean over generation seeds). The black dashed line is the chance baseline (Section~\ref{sec:baselines}), for classification that of OEQ, which like top-1 accuracy names a class from the dataset's list. VQA has no supervised baseline.}
\label{fig:yolo}
\end{figure}

YOLO accuracy grows with the number of training images per class, but the gains shrink (Figure~\ref{fig:yolo}). From 5 to 15 images per class, top-1 classification accuracy rises from 70.2\% to 80.6\%, detection from 32.4\% to 40.9\% and segmentation from 47.4\% to 52.6\%, with most of the gain between 5 and 10 images. Even 5 images per class are enough to outperform the best VLM on every task, so the supervised model catches up with the VLMs below the smallest training set tested. The three training subsets give nearly the same accuracy, differing by at most 1.1 points, except for segmentation (up to 3.2 points), which has only 34 datasets.

\section{Drivers of accuracy (regression analysis)}
\label{app:drivers}
\setcounter{table}{0}
\renewcommand{\thetable}{C\arabic{table}}
\setcounter{figure}{0}
\renewcommand{\thefigure}{C\arabic{figure}}
The experiments vary several evaluation settings, each one at a time from the default. To compare their influence on a common scale, we fit an item-level linear probability model of correctness on the main effects of the dataset, the model, the answer format (OEQ vs.\ MCQA), the distractor source (MCQA within vs.\ across), the NOTA rate (10, 30 or 50\%), whether IDK is offered, the option order (default vs.\ randomized, which also randomizes the position of NOTA and IDK) and the generation seed. Items are weighted so that every dataset counts equally within each run and format, as in the macro averages of the main text. The fit pools Experiments 1--4 for the seven models that took part in all of them (Nemotron ran only Experiment 1), giving 49 runs and 1.60 million classification answers over 220 datasets. Multiple-choice VQA, the only VQA format varied by the experiments, is fitted separately (81,081 answers, 7 datasets). Detection and segmentation are not included, because no evaluation setting was varied for them.

Each variable's importance is its Shapley share of the explained variance ($R^2$), i.e., its contribution averaged over all orders in which the variables can be added to the model \cite{lindeman_introduction_1980}. Effects are the model coefficients in accuracy points, with 95\% confidence intervals clustered by dataset (Table~\ref{tab:drivers}).

\begin{table*}[!tbp]
\centering
\caption{Relative importance of evaluation variables. R$^2$ (pts) is each variable's Shapley share of the variance explained by the main-effects model, in percentage points of $R^2$, and Share expresses it as a percentage of the explained variance. The interaction rows give the $R^2$ added to the main-effects model. Effects are changes in accuracy (points) relative to the default setting (MCQA within for the distractor source; MCQA for OEQ), with 95\% confidence intervals clustered by dataset. Seven models, Experiments 1--4.}
\label{tab:drivers}
\renewcommand{\arraystretch}{1.15}
\tablefont
\begin{tabular}{lrrrr}
\toprule
 & \multicolumn{2}{c}{\textbf{Classification}} & \multicolumn{2}{c}{\textbf{VQA multiple choice}} \\
\cmidrule(lr){2-3} \cmidrule(lr){4-5}
\textbf{Variable} & \textbf{R$^2$ (pts)} & \textbf{Share} & \textbf{R$^2$ (pts)} & \textbf{Share} \\
\midrule
Dataset & $10.47$ & $73.4\%$ & $4.87$ & $53.7\%$ \\
Answer format (OEQ vs.\ MCQA) & $1.95$ & $13.7\%$ & -- & -- \\
Model & $1.08$ & $7.6\%$ & $4.09$ & $45.1\%$ \\
Distractor source (across vs.\ within) & $0.52$ & $3.6\%$ & -- & -- \\
NOTA rate & $0.11$ & $0.8\%$ & $0.04$ & $0.5\%$ \\
Generation seed & $0.07$ & $0.5\%$ & $0.04$ & $0.4\%$ \\
IDK offered & $0.05$ & $0.3\%$ & $0.02$ & $0.2\%$ \\
Option order and NOTA/IDK position & $0.03$ & $0.2\%$ & $0.01$ & $0.1\%$ \\
\midrule
Main effects, total & $14.28$ & $100\%$ & $9.07$ & $100\%$ \\
\quad + Model $\times$ prompt variables & $+0.25$ & & $+0.21$ & \\
\quad + Dataset $\times$ answer format and distractor source & $+2.78$ & & -- & \\
\midrule
\multicolumn{5}{l}{\textit{Effects on accuracy (points, 95\% CI clustered by dataset)}} \\
OEQ (vs.\ MCQA) & \multicolumn{2}{c}{$-18.6$ [$-20.4$, $-16.8$]} & \multicolumn{2}{c}{--} \\
MCQA across (vs.\ within) & \multicolumn{2}{c}{$+4.3$ [$+2.5$, $+6.1$]} & \multicolumn{2}{c}{--} \\
NOTA rate 10\% (vs.\ 30\%) & \multicolumn{2}{c}{$-1.9$ [$-2.6$, $-1.1$]} & \multicolumn{2}{c}{$+1.5$ [$-3.9$, $+6.9$]} \\
NOTA rate 50\% (vs.\ 30\%) & \multicolumn{2}{c}{$+2.0$ [$+1.2$, $+2.7$]} & \multicolumn{2}{c}{$-2.3$ [$-11.1$, $+6.4$]} \\
IDK removed & \multicolumn{2}{c}{$+0.4$ [$+0.1$, $+0.8$]} & \multicolumn{2}{c}{$+2.0$ [$-3.1$, $+7.0$]} \\
Randomized order and NOTA/IDK position & \multicolumn{2}{c}{$-0.8$ [$-1.1$, $-0.5$]} & \multicolumn{2}{c}{$+1.2$ [$+0.0$, $+2.4$]} \\
Generation seed 1 (vs.\ 0) & \multicolumn{2}{c}{$+0.0$ [$-0.2$, $+0.2$]} & \multicolumn{2}{c}{$-2.5$ [$-7.5$, $+2.4$]} \\
Generation seed 2 (vs.\ 0) & \multicolumn{2}{c}{$+0.1$ [$-0.2$, $+0.3$]} & \multicolumn{2}{c}{$-0.2$ [$-1.5$, $+1.0$]} \\
Dataset effects, SD across datasets & \multicolumn{2}{c}{$16.1$} & \multicolumn{2}{c}{$11.9$} \\
\bottomrule
\end{tabular}

\end{table*}

For classification the main effects explain 14.3\% of the item-level variance. Most of the rest is variation between images within a dataset and run-to-run randomness of individual answers. Of the explained part, the dataset accounts for 73\%, since datasets differ in difficulty with a standard deviation of 16.1 points. The answer format follows with 14\% (OEQ is 18.6 points below MCQA), then the model with 8\% and the distractor source with 4\% (MCQA across is 4.3 points above MCQA within). The NOTA rate, IDK, option order and generation seed together account for less than 2\%. Their average effects are at most 2 points, although the NOTA rate affects individual models by up to 7.7 points in opposite directions (Section~\ref{sec:results_design}), which an average effect hides. Allowing the effect of the format and distractor source to differ by dataset adds 2.8 points of $R^2$, more than the format's main effect. This is the variation by agricultural task in Figure~\ref{fig:agtask_chart}, where some datasets lose far more than others in the open-ended format. Allowing the prompt variables to differ by model adds 0.25 points.

For multiple-choice VQA, the dataset (54\%) and the model (45\%) explain almost all of the explained variance, and none of the option-set settings has a significant average effect except a 1.2-point gain from randomized option order.

In short, which dataset a model is tested on matters most, followed by how the question is asked (open-ended vs.\ multiple choice) and which model is used. The option-set details matter little on average, but they can change model rankings.

\section{Reproducing the results}
\label{app:repro}
\setcounter{table}{0}
\renewcommand{\thetable}{D\arabic{table}}
\setcounter{figure}{0}
\renewcommand{\thefigure}{D\arabic{figure}}
Every result table and figure in this paper is written by a script from the stored model outputs, and every number quoted in the text is taken from the files these scripts write. Only the descriptive Tables~\ref{tab:models}--\ref{tab:prompts}, \ref{tab:glossary} and~\ref{tab:experiments} are written by hand. This appendix maps each result to its inputs and to the command that produces it, so that a reader can check any claim without trusting how the article was written. All paths are relative to the repository, \repo.

\subsection{What each run records}
Each evaluation run writes one directory. Its \path{run.json} records the model, sampling parameters, reasoning setting, generation seed, option settings (NOTA rate, option order, IDK), judge model and code commit. Its \path{items.jsonl} lists every evaluation item with its image, prompt, options and reference answer, and \path{predictions.jsonl} keeps, for every item, the raw response, the reasoning trace, the parsed answer, whether it is correct and, for judged items, the judge's full verdict and reasoning. Finally, \path{metrics.json} and \path{compute.json} hold the run's scores and its GPU time and memory.
Any single answer quoted in the paper, such as those in Figures~\ref{fig:prompts_overview} and~\ref{fig:judge_samples}, can therefore be traced to its prompt, response and verdict. The scripts that build the paper also write \path{paper/numbers.json}, which holds every aggregate they compute, and the statistics reports \path{reports/stats/seeds/stats.md} and \path{reports/stats/exp24/stats.md} hold every confidence interval and test quoted in the text.

\subsection{Reproduction from run outputs}
Given the run directories, the single command, which needs only a CPU,
{\footnotesize
\begin{verbatim}
AGBENCH_SEEDS=<Exp. 1 runs> AGBENCH_EXP24=<Exp. 2-4 runs> \
  AGBENCH_BASELINES=<baseline runs> AGBENCH_YOLO=<YOLO runs> \
  bash paper/scripts/reproduce.sh
\end{verbatim}
}
\noindent recomputes the statistics from the per-item predictions, regenerates every table and figure and rebuilds this PDF. The runs can be the reader's own, produced with the commands in Section~\ref{app:rerun}; without the variables, the script reads the paths of our runs (Table~\ref{tab:provenance}). Table~\ref{tab:provenance} lists what each step produces. It takes about 7 minutes on a laptop CPU. On our runs, we checked that it regenerates both statistics reports byte for byte, and \path{paper/numbers.json}, \path{paper/drivers.json}, every generated table and every figure identically.

\begin{table*}[!tbp]
\centering
\caption{Provenance of every result. Exp.~1 denotes the Experiment~1 runs (\code{runs/rejudge\_v4/exp/seeds/s0}--\code{s2}) and Exp.~2--4 the runs in \code{runs/runpod\_exp/runs}. YOLO denotes the few-shot runs in \code{runs/yolo11s-shared/<k>shot/s0}--\code{s2}. \emph{Seeds stats} and \emph{exp24 stats} are the reports written by step~1 of \code{reproduce.sh} (\code{agbench eval stats}). Scripts are in \code{paper/scripts/}; \code{make\_behaviour.py} and \code{make\_position.py} read \code{paper/numbers.json}, which \code{make\_assets.py} writes.}
\label{tab:provenance}
\renewcommand{\arraystretch}{1.25}
\tablefont
\begin{tabular}{p{2.9cm} p{5.6cm} p{6.7cm}}
\toprule
\textbf{Result} & \textbf{Inputs} & \textbf{Produced by} \\
\midrule
Table~\ref{tab:models} & \path{src/agbench/eval/models.yaml}; \path{run.json} of each run & (configuration; written by hand) \\
Table~\ref{tab:data}, Figure~\ref{fig:dataset_overview} & sampled items (\path{manifests/}); one Exp.~1 run's \path{items.jsonl} & \code{agbench sample}; \code{make\_overview.py} \\
Table~\ref{tab:prompts}, Figure~\ref{fig:prompts_overview} & \path{src/agbench/eval/prompts.yaml}; Gemma-4-31B, seed~0 & \code{extract\_prompt\_examples.py}, \code{make\_prompts.py} \\
Figures~\ref{fig:judge_samples}, \ref{fig:judge_prompt} & Exp.~1 seed~0 verdicts; \path{prompts.yaml} & \code{make\_judge.py} \\
Figure~\ref{fig:main_results_chart}, Table~\ref{tab:main_results} & Exp.~1 (per-item predictions); model pairs from seeds stats & \code{make\_assets.py} \\
Figure~\ref{fig:agtask_chart}, Tables~\ref{tab:agtask}, \ref{tab:judge} & Exp.~1 (per-item predictions) & \code{make\_assets.py} \\
Table~\ref{tab:itt} & Exp.~1 (per-item predictions) & \code{make\_assets.py} \\
Figure~\ref{fig:behaviour}, Table~\ref{tab:behaviour} & seeds stats (behaviour; within vs.\ across) & \code{make\_assets.py}, \code{make\_behaviour.py} \\
Table~\ref{tab:ablation_nota} & exp24 stats (NOTA rate); Exp.~1 seed-0 predictions (NOTA recall, listed accuracy) & \code{make\_assets.py} \\
Table~\ref{tab:ablation_idk} & exp24 stats (IDK removed) & \code{make\_assets.py} \\
Figure~\ref{fig:position}, Table~\ref{tab:ablation_order} & exp24 stats (position; option order) & \code{make\_assets.py}, \code{make\_position.py} \\
Table~\ref{tab:seeds} & seeds stats (seed stability) & \code{make\_assets.py} \\
Table~\ref{tab:drivers} & Exp.~1 and Exp.~2--4 (per-item predictions) & \code{make\_drivers.py} \\
YOLO rows of Tables~\ref{tab:main_results} and~\ref{tab:agtask}, bars of Figure~\ref{fig:agtask_chart}, Figure~\ref{fig:yolo} & YOLO runs (per-item predictions, \path{compute.json}) & \code{make\_assets.py} \\
Table~\ref{tab:compute} & \path{compute.json} of every run & \code{make\_assets.py} \\
Numbers in the text & \path{paper/numbers.json}; seeds and exp24 stats & the scripts above; audited by \code{check\_numbers.py} \\
\bottomrule
\end{tabular}
\end{table*}

\subsection{Auditing the numbers in the text}
The last step, \path{paper/scripts/check_numbers.py}, extracts every number from the text and captions and checks that each one occurs, at the precision written, among the values the pipeline computed (the files above, the statistics reports, the generated tables, benchmark counts recomputed from the items, and the configuration files). For this version every number is found. The check is strict for integers and counts, since a random integer of the same range would be found 0\% of the time. It is weak for numbers with decimals, because almost every one-decimal value between 0 and 100 occurs somewhere in the outputs; for those it catches only gross errors, and Table~\ref{tab:provenance} is the way to check the exact source. While this appendix was being written, the audit caught one error. Table~\ref{tab:data} listed the number of VQA questions (3,097) as the number of VQA images (3,592).

\subsection{Reproduction from scratch}\label{app:rerun}
Rerunning the models requires a GPU. The exact run configurations are stored as plan files, one line per run (\path{experiments/seeds.plan} for Experiment~1, \path{experiments/exp24.plan} for Experiments~2--4, \path{experiments/yolo_shared.plan} for the YOLO baselines), and the commands below rerun everything from scratch.
{\footnotesize
\begin{verbatim}
uv sync && hf auth login
uv run agbench sample --data-dir $DATA --from-manifest     # the exact items used here
DATA=$DATA scripts/run_plan.sh experiments/seeds.plan       # Exp. 1: 24 runs
DATA=$DATA scripts/run_plan.sh experiments/exp24.plan       # Exp. 2-4: 56 runs
for s in 0 1 2; do uv run agbench eval baseline --data-dir $DATA --kind chance \
  --gen-seed $s --out runs/baselines/chance/s$s; done
uv run agbench eval baseline --data-dir $DATA --kind prior --out runs/baselines/prior
uv run agbench sample-train --data-dir $DATA --from-manifest   # YOLO training images
grep -v ^# experiments/yolo_shared.plan | while read k s; do
  uv run --extra yolo agbench eval yolo-shared --data-dir $DATA --shots $k --subset $s \
    --size s --out runs/yolo11s-shared; done
AGBENCH_SEEDS=runs/exp/seeds AGBENCH_EXP24=runs/exp24 bash paper/scripts/reproduce.sh
\end{verbatim}
}
\noindent A plan line such as
{\footnotesize
\begin{verbatim}
runs/exp24/order/o1 Qwen/Qwen3.8-27B-FP8 --task classification --task image_text_to_text
  --format mcqa_within --format mcqa_across --format multiple_choice --gen-seed 0
  --order-seed 1 --nota-position random
\end{verbatim}
}
\noindent runs \code{uv run agbench eval run --model <model> --data-dir \$DATA --out <dir> <args>}. Table~\ref{tab:compute} gives the GPU time this takes. Because batched GPU inference is not bit-reproducible (Section~\ref{sec:stat_experiments}), re-runs will not reproduce individual answers exactly; they should reproduce the aggregate results within the seed-to-seed variation of Table~\ref{tab:seeds}.

\begin{table*}[!tbp]
\centering
\caption{GPU time to reproduce the runs, one RTX 6000 Ada (48\,GB) per run, in hours including model loading. For Experiment~1 the columns give the mean per seed of generation and of LLM judging (the final judging pass), the total for all three seeds and the peak GPU memory during generation. For Experiments~2--4 they give the total over the model's runs (five option orders, two NOTA rates and IDK removed), with the number of runs in parentheses. $^\dagger$Also the LLM judge.}
\label{tab:compute}
\renewcommand{\arraystretch}{1.2}
\tablefont
\begin{tabular}{lccccc}
\toprule
\multirow{2}{*}{\textbf{Model}} & \multicolumn{4}{c}{\textbf{Experiment 1 (all formats, per seed)}} & \textbf{Exp.\ 2--4} \\
\cmidrule(lr){2-5} \cmidrule(lr){6-6}
 & \makecell{Generation\\(h)} & \makecell{Judge\\(h)} & \makecell{Total, 3 seeds\\(h)} & \makecell{Peak VRAM\\(GB)} & \makecell{Total (h)\\(runs)} \\
\midrule
Qwen3.6-35B-A3B & $5.7$ & $1.0$ & $20.1$ & $49.0$ & $27.0$ (8) \\
Qwen3.8-27B & $8.4$ & $1.1$ & $28.4$ & $47.4$ & $33.3$ (8) \\
Gemma-4-31B$^\dagger$ & $11.0$ & $1.0$ & $36.0$ & $47.8$ & $48.3$ (8) \\
Gemma-4-12B & $4.5$ & $1.0$ & $16.4$ & $47.7$ & $20.8$ (8) \\
Muse-Glimmer-30B & $7.1$ & $1.0$ & $24.5$ & $50.3$ & $33.4$ (8) \\
Nemotron-3.5-Bolt-74B-A9B & $23.7$ & $0.9$ & $73.9$ & $27.3$ & --- \\
MiniCPM-V-4.7-35B-A3B & $2.3$ & $1.0$ & $10.0$ & $50.9$ & $10.9$ (8) \\
Ministral-3-14B & $2.1$ & $0.9$ & $9.1$ & $48.0$ & $10.3$ (8) \\
\midrule
All models & & & $218$ & & $184$ \\
\bottomrule
\end{tabular}

\end{table*}

\subsection{Software versions}
Experiment~1 ran on the UC Davis farm cluster with agbench commit \code{9352706} and vLLM 0.32.0, and its open-ended answers were then re-judged with the final judge prompt (commit \code{524ee35}), which is the default in \path{src/agbench/eval/prompts.yaml}. Experiments~2--4 ran on Runpod with the \code{vllm/vllm-openai:v0.32.0} image; they contain only multiple-choice questions and are not judged. Nemotron ran with llama.cpp built from source. The YOLO11 baseline uses Ultralytics 8.4.210 with PyTorch 2.15 (CUDA 13.2), trained on the farm cluster on one NVIDIA A100 80\,GB and two TITAN RTX GPUs. All VLM runs used one RTX 6000 Ada GPU. Python dependencies are pinned in \path{uv.lock}. The YOLO baseline used agbench commit \code{975ed41}. Its training images are drawn by \code{agbench sample-train}, and each line of \path{experiments/yolo_shared.plan} runs \code{agbench eval yolo-shared --shots <k> --subset <s> --size s}. At 10 images per class, the classification model trains in about 4 GPU-hours, the detection model in about 1 and the segmentation model in under half an hour.

\section{Datasets}
\label{app:datasets}
The 372 datasets, grouped by task. Each name is the dataset's AgML identifier, followed by its number of classes and of evaluated images (for VQA, its number of questions and of images). Per-dataset scores for every model are in the repository (\path{reports/scores/}).

{\footnotesize
\subsection*{Classification (220 datasets; classes, images)}
\begin{multicols}{2}\raggedright
\hangindent=1em ACHENY\_\allowbreak{}variety\_\allowbreak{}classification (30, 150)\par
\hangindent=1em african\_\allowbreak{}plum\_\allowbreak{}grading\_\allowbreak{}classification (6, 30)\par
\hangindent=1em AFruitDB\_\allowbreak{}fruit\_\allowbreak{}grade\_\allowbreak{}classification (3, 15)\par
\hangindent=1em agarwood\_\allowbreak{}leaf\_\allowbreak{}disease\_\allowbreak{}classification (14, 70)\par
\hangindent=1em AgriVision4\_\allowbreak{}disease\_\allowbreak{}classification (22, 110)\par
\hangindent=1em apple\_\allowbreak{}leaf\_\allowbreak{}disease\_\allowbreak{}classification (3, 15)\par
\hangindent=1em appleleaf9\_\allowbreak{}disease\_\allowbreak{}classification (9, 45)\par
\hangindent=1em arabica\_\allowbreak{}coffee\_\allowbreak{}leaf\_\allowbreak{}disease\_\allowbreak{}classification (5, 25)\par
\hangindent=1em ash\_\allowbreak{}gourd\_\allowbreak{}disease\_\allowbreak{}classification (5, 25)\par
\hangindent=1em bacterial\_\allowbreak{}grain\_\allowbreak{}rot\_\allowbreak{}classification (4, 20)\par
\hangindent=1em banana\_\allowbreak{}disease\_\allowbreak{}classification\_\allowbreak{}tanzania (3, 15)\par
\hangindent=1em banana\_\allowbreak{}grade\_\allowbreak{}variety\_\allowbreak{}classification (3, 15)\par
\hangindent=1em banana\_\allowbreak{}guava\_\allowbreak{}quality\_\allowbreak{}classification (3, 15)\par
\hangindent=1em banana\_\allowbreak{}leaf\_\allowbreak{}disease\_\allowbreak{}classification (3, 15)\par
\hangindent=1em banana\_\allowbreak{}leaf\_\allowbreak{}nutrient\_\allowbreak{}classification (9, 45)\par
\hangindent=1em banana\_\allowbreak{}variety\_\allowbreak{}classification (6, 30)\par
\hangindent=1em BananaImageBD\_\allowbreak{}ripeness\_\allowbreak{}classification (4, 20)\par
\hangindent=1em BananaImageBD\_\allowbreak{}variety\_\allowbreak{}classification (4, 20)\par
\hangindent=1em BananaLSD\_\allowbreak{}leaf\_\allowbreak{}disease\_\allowbreak{}classification (4, 20)\par
\hangindent=1em Basmati\_\allowbreak{}rice\_\allowbreak{}seed\_\allowbreak{}varitey\_\allowbreak{}classification (11, 55)\par
\hangindent=1em BAU\_\allowbreak{}insectV2\_\allowbreak{}classification (9, 45)\par
\hangindent=1em bay\_\allowbreak{}leaf\_\allowbreak{}disease\_\allowbreak{}classification (3, 15)\par
\hangindent=1em BDHerbalPlants\_\allowbreak{}variety\_\allowbreak{}classification (8, 40)\par
\hangindent=1em BDMANGO\_\allowbreak{}variety\_\allowbreak{}classification (6, 30)\par
\hangindent=1em BDMediLeaves\_\allowbreak{}variety\_\allowbreak{}classification (10, 50)\par
\hangindent=1em bean\_\allowbreak{}cowpea\_\allowbreak{}leaf\_\allowbreak{}disease\_\allowbreak{}classification (6, 30)\par
\hangindent=1em bean\_\allowbreak{}disease\_\allowbreak{}classification\_\allowbreak{}tanzania (3, 15)\par
\hangindent=1em bean\_\allowbreak{}disease\_\allowbreak{}uganda (3, 15)\par
\hangindent=1em betel\_\allowbreak{}leaf\_\allowbreak{}disease\_\allowbreak{}classification (4, 20)\par
\hangindent=1em betel\_\allowbreak{}leaf\_\allowbreak{}disease\_\allowbreak{}classification\_\allowbreak{}2 (3, 15)\par
\hangindent=1em black\_\allowbreak{}gram\_\allowbreak{}disease\_\allowbreak{}classification (5, 25)\par
\hangindent=1em blackgram\_\allowbreak{}plant\_\allowbreak{}leaf\_\allowbreak{}disease\_\allowbreak{}classification (5, 25)\par
\hangindent=1em BPLD\_\allowbreak{}leaf\_\allowbreak{}disease\_\allowbreak{}classification (5, 25)\par
\hangindent=1em BrinjalFruitX\_\allowbreak{}disease\_\allowbreak{}classification (5, 25)\par
\hangindent=1em bruised\_\allowbreak{}vegetable\_\allowbreak{}classification (3, 15)\par
\hangindent=1em carambola\_\allowbreak{}disease\_\allowbreak{}classification (5, 25)\par
\hangindent=1em cauliflower\_\allowbreak{}leaf\_\allowbreak{}disease\_\allowbreak{}classification (3, 15)\par
\hangindent=1em centella\_\allowbreak{}asiatica\_\allowbreak{}leaves (3, 15)\par
\hangindent=1em chitrak\_\allowbreak{}leaf\_\allowbreak{}disease\_\allowbreak{}classification (3, 15)\par
\hangindent=1em citrus\_\allowbreak{}fruit\_\allowbreak{}leaf\_\allowbreak{}disease\_\allowbreak{}classification (6, 30)\par
\hangindent=1em citrus\_\allowbreak{}fruit\_\allowbreak{}variety\_\allowbreak{}classification (4, 20)\par
\hangindent=1em citrusuat\_\allowbreak{}disease\_\allowbreak{}classification (12, 60)\par
\hangindent=1em coconut\_\allowbreak{}tree\_\allowbreak{}disease\_\allowbreak{}classification (5, 25)\par
\hangindent=1em coffee\_\allowbreak{}bean\_\allowbreak{}quality\_\allowbreak{}classification (9, 45)\par
\hangindent=1em COLD\_\allowbreak{}chili\_\allowbreak{}leaf\_\allowbreak{}disease\_\allowbreak{}classification (5, 25)\par
\hangindent=1em COLD\_\allowbreak{}onion\_\allowbreak{}leaf\_\allowbreak{}disease\_\allowbreak{}classification (4, 20)\par
\hangindent=1em CoLeaf\_\allowbreak{}nutritional\_\allowbreak{}deficiency\_\allowbreak{}classification (9, 45)\par
\hangindent=1em corn\_\allowbreak{}leaf\_\allowbreak{}disease\_\allowbreak{}classification (6, 30)\par
\hangindent=1em corn\_\allowbreak{}leaf\_\allowbreak{}pest\_\allowbreak{}classification (4, 20)\par
\hangindent=1em corn\_\allowbreak{}maize\_\allowbreak{}leaf\_\allowbreak{}disease (4, 20)\par
\hangindent=1em cotton\_\allowbreak{}disease\_\allowbreak{}classification (5, 25)\par
\hangindent=1em cotton\_\allowbreak{}leaf\_\allowbreak{}disease\_\allowbreak{}classification (5, 25)\par
\hangindent=1em cotton\_\allowbreak{}leaf\_\allowbreak{}disease\_\allowbreak{}classification\_\allowbreak{}bangladesh\_\allowbreak{}2 (7, 35)\par
\hangindent=1em CottonWeedID15\_\allowbreak{}classification (15, 75)\par
\hangindent=1em crop\_\allowbreak{}pest\_\allowbreak{}disease\_\allowbreak{}classification (18, 90)\par
\hangindent=1em crop\_\allowbreak{}weeds\_\allowbreak{}greece (4, 20)\par
\hangindent=1em CT\_\allowbreak{}PestNet\_\allowbreak{}crop\_\allowbreak{}pest\_\allowbreak{}classification (102, 510)\par
\hangindent=1em ct\_\allowbreak{}pestnet\_\allowbreak{}crop\_\allowbreak{}pest\_\allowbreak{}classification\_\allowbreak{}2 (40, 200)\par
\hangindent=1em cucumber\_\allowbreak{}disease\_\allowbreak{}classification (8, 40)\par
\hangindent=1em custard\_\allowbreak{}apple\_\allowbreak{}disease\_\allowbreak{}classification (6, 30)\par
\hangindent=1em date\_\allowbreak{}grade\_\allowbreak{}variety\_\allowbreak{}classification (3, 15)\par
\hangindent=1em date\_\allowbreak{}palm\_\allowbreak{}leaf\_\allowbreak{}disease\_\allowbreak{}classification (9, 45)\par
\hangindent=1em date\_\allowbreak{}palm\_\allowbreak{}leaf\_\allowbreak{}disease\_\allowbreak{}classification\_\allowbreak{}iraq (4, 20)\par
\hangindent=1em DeepWeeds\_\allowbreak{}classification (9, 45)\par
\hangindent=1em Dhan-Shomadhan\_\allowbreak{}rice\_\allowbreak{}leaf\_\allowbreak{}disease\_\allowbreak{}classification (8, 40)\par
\hangindent=1em DIMPSAR\_\allowbreak{}medicinal\_\allowbreak{}leaf\_\allowbreak{}classification (80, 400)\par
\hangindent=1em DIMPSAR\_\allowbreak{}medicinal\_\allowbreak{}plant\_\allowbreak{}classification (40, 200)\par
\hangindent=1em dragonfruit\_\allowbreak{}disease\_\allowbreak{}classification (4, 20)\par
\hangindent=1em dragonfruit\_\allowbreak{}maturity\_\allowbreak{}classification (2, 10)\par
\hangindent=1em dragonfruit\_\allowbreak{}quality\_\allowbreak{}classification (2, 10)\par
\hangindent=1em durian\_\allowbreak{}disease\_\allowbreak{}classification (10, 50)\par
\hangindent=1em durian\_\allowbreak{}disease\_\allowbreak{}classification\_\allowbreak{}vietnam (6, 30)\par
\hangindent=1em early-crop-weed\_\allowbreak{}classification (4, 20)\par
\hangindent=1em EfficientMaize\_\allowbreak{}classification (2, 10)\par
\hangindent=1em eggplant\_\allowbreak{}disease\_\allowbreak{}classification (6, 30)\par
\hangindent=1em eggplant\_\allowbreak{}leaf\_\allowbreak{}disease\_\allowbreak{}classification (10, 50)\par
\hangindent=1em fig\_\allowbreak{}leaf\_\allowbreak{}ficus\_\allowbreak{}worm\_\allowbreak{}classification (2, 10)\par
\hangindent=1em fresh\_\allowbreak{}rotten\_\allowbreak{}fruit\_\allowbreak{}classification (2, 10)\par
\hangindent=1em fruit\_\allowbreak{}leaf\_\allowbreak{}variety\_\allowbreak{}classification (10, 50)\par
\hangindent=1em FruitNet\_\allowbreak{}quality\_\allowbreak{}classification (3, 15)\par
\hangindent=1em FruitVision\_\allowbreak{}quality\_\allowbreak{}classification (3, 15)\par
\hangindent=1em grape\_\allowbreak{}leaf\_\allowbreak{}disease\_\allowbreak{}classification (4, 20)\par
\hangindent=1em grape\_\allowbreak{}variety\_\allowbreak{}classification (8, 40)\par
\hangindent=1em grapevine\_\allowbreak{}development\_\allowbreak{}stage\_\allowbreak{}classification (4, 20)\par
\hangindent=1em grapevine\_\allowbreak{}disease\_\allowbreak{}classification (3, 20)\par
\hangindent=1em grapevine\_\allowbreak{}esca\_\allowbreak{}classification (2, 10)\par
\hangindent=1em greenhouse\_\allowbreak{}crop\_\allowbreak{}weed\_\allowbreak{}classification (14, 70)\par
\hangindent=1em groundnut\_\allowbreak{}leaf\_\allowbreak{}disease\_\allowbreak{}classification (5, 25)\par
\hangindent=1em groundnut\_\allowbreak{}leaf\_\allowbreak{}disease\_\allowbreak{}classification\_\allowbreak{}2 (5, 25)\par
\hangindent=1em guava\_\allowbreak{}disease\_\allowbreak{}classification (7, 35)\par
\hangindent=1em guava\_\allowbreak{}disease\_\allowbreak{}classification\_\allowbreak{}bangladesh (5, 25)\par
\hangindent=1em guava\_\allowbreak{}disease\_\allowbreak{}pakistan (4, 20)\par
\hangindent=1em guava\_\allowbreak{}maturity\_\allowbreak{}classification (3, 15)\par
\hangindent=1em Hibiscus\_\allowbreak{}Tea\_\allowbreak{}disease\_\allowbreak{}classification (12, 60)\par
\hangindent=1em hog\_\allowbreak{}plum\_\allowbreak{}leaf\_\allowbreak{}disease\_\allowbreak{}classification (2, 10)\par
\hangindent=1em ICPTC\_\allowbreak{}pistachio\_\allowbreak{}tree\_\allowbreak{}variety\_\allowbreak{}classification (4, 20)\par
\hangindent=1em IDDMSLD\_\allowbreak{}spinach\_\allowbreak{}leaf\_\allowbreak{}disease\_\allowbreak{}classification (5, 25)\par
\hangindent=1em java\_\allowbreak{}plum\_\allowbreak{}leaf\_\allowbreak{}disease\_\allowbreak{}classification (6, 30)\par
\hangindent=1em jujube\_\allowbreak{}bruise\_\allowbreak{}classification (2, 10)\par
\hangindent=1em jute\_\allowbreak{}disease\_\allowbreak{}classification (5, 25)\par
\hangindent=1em LeafNet\_\allowbreak{}plant\_\allowbreak{}disease\_\allowbreak{}classification (2, 10)\par
\hangindent=1em lemon\_\allowbreak{}leaf\_\allowbreak{}disease\_\allowbreak{}classification (9, 45)\par
\hangindent=1em lemongrass\_\allowbreak{}disease\_\allowbreak{}classification (3, 15)\par
\hangindent=1em lentil\_\allowbreak{}disease\_\allowbreak{}classification (4, 20)\par
\hangindent=1em Luffa\_\allowbreak{}disease\_\allowbreak{}classification (5, 25)\par
\hangindent=1em Luffa\_\allowbreak{}quality\_\allowbreak{}classification (2, 10)\par
\hangindent=1em maize\_\allowbreak{}disease\_\allowbreak{}classification (3, 15)\par
\hangindent=1em maize\_\allowbreak{}nitrogen\_\allowbreak{}deficiency\_\allowbreak{}classification (3, 15)\par
\hangindent=1em maize\_\allowbreak{}tomato\_\allowbreak{}weed\_\allowbreak{}classification (12, 60)\par
\hangindent=1em malabar\_\allowbreak{}spinach\_\allowbreak{}disease\_\allowbreak{}classification (3, 15)\par
\hangindent=1em mandarin\_\allowbreak{}leaf\_\allowbreak{}variety\_\allowbreak{}classification (4, 20)\par
\hangindent=1em mangifera2012\_\allowbreak{}variety\_\allowbreak{}classification (10, 50)\par
\hangindent=1em mango\_\allowbreak{}leaf\_\allowbreak{}disease\_\allowbreak{}classification (8, 40)\par
\hangindent=1em mango\_\allowbreak{}pest\_\allowbreak{}original\_\allowbreak{}classification (16, 80)\par
\hangindent=1em MangoClassify-12\_\allowbreak{}variety\_\allowbreak{}classification (12, 60)\par
\hangindent=1em MangoImageBD\_\allowbreak{}classification (15, 75)\par
\hangindent=1em MangoLeafBD\_\allowbreak{}disease\_\allowbreak{}classification (8, 40)\par
\hangindent=1em medicinal\_\allowbreak{}leaf\_\allowbreak{}species\_\allowbreak{}classification (6, 30)\par
\hangindent=1em medicinal\_\allowbreak{}plant\_\allowbreak{}classification\_\allowbreak{}bd (10, 50)\par
\hangindent=1em medicinal\_\allowbreak{}plant\_\allowbreak{}parts\_\allowbreak{}classification (8, 40)\par
\hangindent=1em MedLeafX\_\allowbreak{}disease\_\allowbreak{}classification (6, 30)\par
\hangindent=1em MH\_\allowbreak{}SoyaHealthVision\_\allowbreak{}disease\_\allowbreak{}classification\_\allowbreak{}leaf (6, 30)\par
\hangindent=1em MH\_\allowbreak{}SoyaHealthVision\_\allowbreak{}disease\_\allowbreak{}classification\_\allowbreak{}uav (4, 20)\par
\hangindent=1em MH\_\allowbreak{}Weed16\_\allowbreak{}weed\_\allowbreak{}variety\_\allowbreak{}classification (16, 80)\par
\hangindent=1em mint\_\allowbreak{}leaf\_\allowbreak{}classification (4, 20)\par
\hangindent=1em money\_\allowbreak{}plant\_\allowbreak{}disease\_\allowbreak{}classification (3, 15)\par
\hangindent=1em MoringaLeafNet\_\allowbreak{}disease\_\allowbreak{}classification (4, 20)\par
\hangindent=1em mulberry\_\allowbreak{}leaf\_\allowbreak{}variety\_\allowbreak{}classification (10, 50)\par
\hangindent=1em Musa\_\allowbreak{}banana\_\allowbreak{}tier\_\allowbreak{}classification (4, 20)\par
\hangindent=1em oil\_\allowbreak{}palm\_\allowbreak{}fruit\_\allowbreak{}ripeness\_\allowbreak{}classification (5, 25)\par
\hangindent=1em okra\_\allowbreak{}maturity\_\allowbreak{}classification (3, 15)\par
\hangindent=1em OkraDiseaseNet\_\allowbreak{}disease\_\allowbreak{}classification (6, 30)\par
\hangindent=1em OpenEar\_\allowbreak{}developmental\_\allowbreak{}status\_\allowbreak{}classification (2, 10)\par
\hangindent=1em orange\_\allowbreak{}leaf\_\allowbreak{}disease\_\allowbreak{}classification (11, 55)\par
\hangindent=1em orange\_\allowbreak{}leaf\_\allowbreak{}HLB\_\allowbreak{}classification (2, 10)\par
\hangindent=1em paddy\_\allowbreak{}disease\_\allowbreak{}classification (10, 50)\par
\hangindent=1em PaddyNet\_\allowbreak{}LCC\_\allowbreak{}classification (4, 20)\par
\hangindent=1em PaddyVarietyBD\_\allowbreak{}variety\_\allowbreak{}classification (35, 175)\par
\hangindent=1em papaya\_\allowbreak{}leaf\_\allowbreak{}disease\_\allowbreak{}classification\_\allowbreak{}bangladesh (6, 30)\par
\hangindent=1em papaya\_\allowbreak{}leaf\_\allowbreak{}disease\_\allowbreak{}classification\_\allowbreak{}bd (8, 40)\par
\hangindent=1em pea\_\allowbreak{}seed\_\allowbreak{}viability\_\allowbreak{}classification (2, 10)\par
\hangindent=1em pepper\_\allowbreak{}disease\_\allowbreak{}pest\_\allowbreak{}classification (5, 25)\par
\hangindent=1em PFSD\_\allowbreak{}Musa\_\allowbreak{}banana\_\allowbreak{}disease\_\allowbreak{}classification (8, 40)\par
\hangindent=1em PFSD\_\allowbreak{}Musa\_\allowbreak{}banana\_\allowbreak{}variety\_\allowbreak{}classification (7, 35)\par
\hangindent=1em plant\_\allowbreak{}doc\_\allowbreak{}classification (28, 137)\par
\hangindent=1em plant\_\allowbreak{}leaf\_\allowbreak{}disease\_\allowbreak{}classification (16, 80)\par
\hangindent=1em Plant\_\allowbreak{}leaf\_\allowbreak{}diseases\_\allowbreak{}dataset\_\allowbreak{}classification (39, 195)\par
\hangindent=1em plant\_\allowbreak{}seedlings\_\allowbreak{}aarhus (12, 60)\par
\hangindent=1em plant\_\allowbreak{}village\_\allowbreak{}classification (39, 195)\par
\hangindent=1em plum\_\allowbreak{}leaf\_\allowbreak{}disease\_\allowbreak{}classification (6, 30)\par
\hangindent=1em plum\_\allowbreak{}leaf\_\allowbreak{}fruit\_\allowbreak{}disease\_\allowbreak{}classification (6, 30)\par
\hangindent=1em pomegranate\_\allowbreak{}disease\_\allowbreak{}classification (4, 20)\par
\hangindent=1em pomegranate\_\allowbreak{}disease\_\allowbreak{}classification\_\allowbreak{}india (5, 25)\par
\hangindent=1em pomegranate\_\allowbreak{}quality\_\allowbreak{}classification (3, 15)\par
\hangindent=1em potato\_\allowbreak{}disease\_\allowbreak{}classification (10, 50)\par
\hangindent=1em potato\_\allowbreak{}leaf\_\allowbreak{}disease\_\allowbreak{}classification (7, 35)\par
\hangindent=1em PriBel\_\allowbreak{}betel\_\allowbreak{}leaf\_\allowbreak{}disease\_\allowbreak{}classification (3, 15)\par
\hangindent=1em radish\_\allowbreak{}leaf\_\allowbreak{}disease\_\allowbreak{}classification (5, 25)\par
\hangindent=1em rangeland\_\allowbreak{}weeds\_\allowbreak{}australia (9, 45)\par
\hangindent=1em REMP\_\allowbreak{}plant\_\allowbreak{}classification (16, 80)\par
\hangindent=1em rice\_\allowbreak{}disease\_\allowbreak{}classification\_\allowbreak{}bangladesh (8, 40)\par
\hangindent=1em rice\_\allowbreak{}field\_\allowbreak{}weed\_\allowbreak{}classification (11, 55)\par
\hangindent=1em rice\_\allowbreak{}grain\_\allowbreak{}variety\_\allowbreak{}classification (20, 100)\par
\hangindent=1em rice\_\allowbreak{}leaf\_\allowbreak{}bacterial\_\allowbreak{}fungal (8, 40)\par
\hangindent=1em rice\_\allowbreak{}leaf\_\allowbreak{}disease\_\allowbreak{}classification (6, 30)\par
\hangindent=1em rice\_\allowbreak{}leaf\_\allowbreak{}disease\_\allowbreak{}classification\_\allowbreak{}2 (3, 15)\par
\hangindent=1em rice\_\allowbreak{}leaf\_\allowbreak{}disease\_\allowbreak{}classification\_\allowbreak{}bd (6, 30)\par
\hangindent=1em rice\_\allowbreak{}leaf\_\allowbreak{}disease\_\allowbreak{}classification\_\allowbreak{}india (4, 20)\par
\hangindent=1em rice\_\allowbreak{}seedling\_\allowbreak{}classification (3, 15)\par
\hangindent=1em rice\_\allowbreak{}variety\_\allowbreak{}classification\_\allowbreak{}bangladesh (38, 190)\par
\hangindent=1em riseholme\_\allowbreak{}strawberry\_\allowbreak{}classification\_\allowbreak{}2021 (4, 20)\par
\hangindent=1em RoseLeafInsight\_\allowbreak{}disease\_\allowbreak{}classification (4, 20)\par
\hangindent=1em roseleafset\_\allowbreak{}disease\_\allowbreak{}classification (4, 20)\par
\hangindent=1em RoseNet\_\allowbreak{}leaf\_\allowbreak{}disease\_\allowbreak{}classification (3, 15)\par
\hangindent=1em SapBark\_\allowbreak{}64\_\allowbreak{}variety\_\allowbreak{}classification (64, 320)\par
\hangindent=1em sapota\_\allowbreak{}fruit\_\allowbreak{}size\_\allowbreak{}classification (4, 20)\par
\hangindent=1em seasveg\_\allowbreak{}classification\_\allowbreak{}bd (10, 50)\par
\hangindent=1em SIMPDV1\_\allowbreak{}plant\_\allowbreak{}classification (20, 100)\par
\hangindent=1em sorghum\_\allowbreak{}weed\_\allowbreak{}classification (3, 15)\par
\hangindent=1em soybean\_\allowbreak{}damage\_\allowbreak{}classification (5, 25)\par
\hangindent=1em soybean\_\allowbreak{}insect\_\allowbreak{}classification (3, 15)\par
\hangindent=1em soybean\_\allowbreak{}leaf\_\allowbreak{}disease\_\allowbreak{}classification (6, 30)\par
\hangindent=1em soybean\_\allowbreak{}leaf\_\allowbreak{}disease\_\allowbreak{}classification\_\allowbreak{}brazil (3, 15)\par
\hangindent=1em soybean\_\allowbreak{}leaf\_\allowbreak{}wilt\_\allowbreak{}classification\_\allowbreak{}2 (6, 30)\par
\hangindent=1em soybean\_\allowbreak{}variety\_\allowbreak{}classification (3, 15)\par
\hangindent=1em soybean\_\allowbreak{}weed\_\allowbreak{}uav\_\allowbreak{}brazil (4, 20)\par
\hangindent=1em SoyNet\_\allowbreak{}leaf\_\allowbreak{}health\_\allowbreak{}classification (2, 10)\par
\hangindent=1em sugarcane\_\allowbreak{}damage\_\allowbreak{}usa (6, 30)\par
\hangindent=1em sugarcane\_\allowbreak{}leaf\_\allowbreak{}disease\_\allowbreak{}classification (11, 55)\par
\hangindent=1em sunflower\_\allowbreak{}disease\_\allowbreak{}classification (4, 20)\par
\hangindent=1em taro\_\allowbreak{}blight\_\allowbreak{}stage\_\allowbreak{}classification (4, 20)\par
\hangindent=1em tea\_\allowbreak{}leaf\_\allowbreak{}disease\_\allowbreak{}classification (6, 30)\par
\hangindent=1em tea\_\allowbreak{}leaf\_\allowbreak{}disease\_\allowbreak{}classification\_\allowbreak{}bangladesh (4, 20)\par
\hangindent=1em tea\_\allowbreak{}leaf\_\allowbreak{}disease\_\allowbreak{}classification\_\allowbreak{}bangladesh\_\allowbreak{}2 (4, 20)\par
\hangindent=1em tea\_\allowbreak{}sickness\_\allowbreak{}classification (8, 40)\par
\hangindent=1em three\_\allowbreak{}plant\_\allowbreak{}leaf\_\allowbreak{}disease\_\allowbreak{}classification (9, 45)\par
\hangindent=1em TOM2024\_\allowbreak{}disease\_\allowbreak{}classification (24, 119)\par
\hangindent=1em tomato\_\allowbreak{}leaf\_\allowbreak{}disease (10, 50)\par
\hangindent=1em tomato\_\allowbreak{}maturity\_\allowbreak{}classification (2, 10)\par
\hangindent=1em tomato\_\allowbreak{}quality\_\allowbreak{}classification (2, 10)\par
\hangindent=1em tree\_\allowbreak{}species\_\allowbreak{}classification\_\allowbreak{}CZU\_\allowbreak{}exact\_\allowbreak{}cropped (4, 20)\par
\hangindent=1em tree\_\allowbreak{}species\_\allowbreak{}classification\_\allowbreak{}CZU\_\allowbreak{}normal\_\allowbreak{}cropped (4, 20)\par
\hangindent=1em tree\_\allowbreak{}species\_\allowbreak{}classification\_\allowbreak{}slovak\_\allowbreak{}exact\_\allowbreak{}cropped (4, 20)\par
\hangindent=1em tree\_\allowbreak{}species\_\allowbreak{}classification\_\allowbreak{}slovak\_\allowbreak{}exact\_\allowbreak{}cropped\_\allowbreak{}extended (4, 20)\par
\hangindent=1em tree\_\allowbreak{}species\_\allowbreak{}classification\_\allowbreak{}slovak\_\allowbreak{}normal\_\allowbreak{}cropped (4, 20)\par
\hangindent=1em tropical\_\allowbreak{}flower\_\allowbreak{}variety\_\allowbreak{}classification (7, 35)\par
\hangindent=1em tropical\_\allowbreak{}forest\_\allowbreak{}classification (11, 55)\par
\hangindent=1em turmeric\_\allowbreak{}disease\_\allowbreak{}classification (5, 25)\par
\hangindent=1em turmeric\_\allowbreak{}leaf\_\allowbreak{}disease\_\allowbreak{}classification (4, 20)\par
\hangindent=1em uci\_\allowbreak{}rice\_\allowbreak{}leaf\_\allowbreak{}diseases\_\allowbreak{}classification (3, 15)\par
\hangindent=1em vegetable\_\allowbreak{}classification\_\allowbreak{}bangladesh\_\allowbreak{}classification (12, 60)\par
\hangindent=1em VegNet\_\allowbreak{}cauliflower\_\allowbreak{}disease\_\allowbreak{}classification (4, 20)\par
\hangindent=1em VegNet\_\allowbreak{}quality\_\allowbreak{}classification (5, 25)\par
\hangindent=1em vine\_\allowbreak{}virus\_\allowbreak{}photo\_\allowbreak{}dataset (4, 20)\par
\hangindent=1em WaterHyacinth\_\allowbreak{}variety\_\allowbreak{}classification (4, 20)\par
\hangindent=1em watermelon\_\allowbreak{}disease\_\allowbreak{}classification (4, 20)\par
\hangindent=1em wheat\_\allowbreak{}growth\_\allowbreak{}stage\_\allowbreak{}classification (15, 75)\par
\hangindent=1em wheat\_\allowbreak{}seed\_\allowbreak{}variety\_\allowbreak{}classification (3, 15)\par
\hangindent=1em WPDv2\_\allowbreak{}BBCH\_\allowbreak{}classification (5, 25)\par
\hangindent=1em YOLOv8\_\allowbreak{}PlantDisease\_\allowbreak{}classification (15, 75)\par
\end{multicols}

\subsection*{Detection (105 datasets; classes, images)}
\begin{multicols}{2}\raggedright
\hangindent=1em almond\_\allowbreak{}bloom\_\allowbreak{}2023 (1, 6)\par
\hangindent=1em almond\_\allowbreak{}harvest\_\allowbreak{}2021 (4, 5)\par
\hangindent=1em apple\_\allowbreak{}blotch\_\allowbreak{}detection (1, 10)\par
\hangindent=1em apple\_\allowbreak{}detection\_\allowbreak{}drone\_\allowbreak{}brazil (1, 5)\par
\hangindent=1em apple\_\allowbreak{}detection\_\allowbreak{}spain (1, 5)\par
\hangindent=1em apple\_\allowbreak{}detection\_\allowbreak{}usa (1, 5)\par
\hangindent=1em apple\_\allowbreak{}fruitlet\_\allowbreak{}detection (1, 5)\par
\hangindent=1em banana\_\allowbreak{}bunch\_\allowbreak{}detection (1, 5)\par
\hangindent=1em BGM\_\allowbreak{}YOLO\_\allowbreak{}plant\_\allowbreak{}disease\_\allowbreak{}classification (12, 59)\par
\hangindent=1em blueberry\_\allowbreak{}maturity\_\allowbreak{}detection (4, 5)\par
\hangindent=1em BMSB\_\allowbreak{}stinkbug\_\allowbreak{}detection (1, 10)\par
\hangindent=1em cashew\_\allowbreak{}detection (6, 11)\par
\hangindent=1em Cassava\_\allowbreak{}whitefly\_\allowbreak{}detection (1, 5)\par
\hangindent=1em CCOD\_\allowbreak{}cotton\_\allowbreak{}flower\_\allowbreak{}detection (4, 20)\par
\hangindent=1em chili\_\allowbreak{}leaf\_\allowbreak{}curl\_\allowbreak{}detection (3, 15)\par
\hangindent=1em cocoa\_\allowbreak{}monilia\_\allowbreak{}detection (5, 19)\par
\hangindent=1em cocoa\_\allowbreak{}pod\_\allowbreak{}disease\_\allowbreak{}gh\_\allowbreak{}image\_\allowbreak{}classification (4, 10)\par
\hangindent=1em cocoa\_\allowbreak{}pod\_\allowbreak{}disease\_\allowbreak{}gh\_\allowbreak{}image\_\allowbreak{}classification\_\allowbreak{}2 (3, 10)\par
\hangindent=1em CocoaMFDB\_\allowbreak{}detection (1, 5)\par
\hangindent=1em coffee\_\allowbreak{}detection (5, 14)\par
\hangindent=1em cotton\_\allowbreak{}weed\_\allowbreak{}detection (2, 5)\par
\hangindent=1em crop\_\allowbreak{}weed\_\allowbreak{}detection\_\allowbreak{}latvia (2, 7)\par
\hangindent=1em date\_\allowbreak{}cluster\_\allowbreak{}detection (1, 10)\par
\hangindent=1em date\_\allowbreak{}fruit\_\allowbreak{}maturity\_\allowbreak{}detection (4, 23)\par
\hangindent=1em embrapa\_\allowbreak{}wgisd\_\allowbreak{}grape\_\allowbreak{}detection (1, 5)\par
\hangindent=1em ERWIAM\_\allowbreak{}blight\_\allowbreak{}detection (4, 13)\par
\hangindent=1em fortunella\_\allowbreak{}margarita\_\allowbreak{}growth\_\allowbreak{}detection (4, 10)\par
\hangindent=1em fruit\_\allowbreak{}detection\_\allowbreak{}worldwide (7, 35)\par
\hangindent=1em GEMINI\_\allowbreak{}cowpea\_\allowbreak{}flower\_\allowbreak{}detection (1, 10)\par
\hangindent=1em GEMINI\_\allowbreak{}cowpea\_\allowbreak{}pod\_\allowbreak{}detection (1, 5)\par
\hangindent=1em gemini\_\allowbreak{}flower\_\allowbreak{}detection (1, 5)\par
\hangindent=1em gemini\_\allowbreak{}leaf\_\allowbreak{}detection (1, 5)\par
\hangindent=1em gemini\_\allowbreak{}plant\_\allowbreak{}detection (2, 7)\par
\hangindent=1em gemini\_\allowbreak{}pod\_\allowbreak{}detection (1, 10)\par
\hangindent=1em ghai\_\allowbreak{}broccoli\_\allowbreak{}detection (3, 5)\par
\hangindent=1em ghai\_\allowbreak{}green\_\allowbreak{}cabbage\_\allowbreak{}detection (1, 5)\par
\hangindent=1em ghai\_\allowbreak{}iceberg\_\allowbreak{}lettuce\_\allowbreak{}detection (1, 5)\par
\hangindent=1em ghai\_\allowbreak{}romaine\_\allowbreak{}detection (1, 5)\par
\hangindent=1em ghai\_\allowbreak{}strawberry\_\allowbreak{}fruit\_\allowbreak{}detection (10, 11)\par
\hangindent=1em grape\_\allowbreak{}detection\_\allowbreak{}californiaday (1, 5)\par
\hangindent=1em grape\_\allowbreak{}detection\_\allowbreak{}californianight (1, 5)\par
\hangindent=1em grape\_\allowbreak{}detection\_\allowbreak{}syntheticday (2, 5)\par
\hangindent=1em grapevine\_\allowbreak{}bunch\_\allowbreak{}condition\_\allowbreak{}detection (2, 10)\par
\hangindent=1em grapevine\_\allowbreak{}bunch\_\allowbreak{}detection (1, 5)\par
\hangindent=1em grapevine\_\allowbreak{}growth\_\allowbreak{}detection (1, 5)\par
\hangindent=1em greenhouse\_\allowbreak{}crop\_\allowbreak{}weed\_\allowbreak{}detection (14, 20)\par
\hangindent=1em Greenhouse\_\allowbreak{}Tomato\_\allowbreak{}AgRob\_\allowbreak{}object\_\allowbreak{}detection (4, 15)\par
\hangindent=1em GYMNSA\_\allowbreak{}pear\_\allowbreak{}rust\_\allowbreak{}detection (0, 10)\par
\hangindent=1em horseradish\_\allowbreak{}weed\_\allowbreak{}detection (2, 5)\par
\hangindent=1em image\_\allowbreak{}safari\_\allowbreak{}crop\_\allowbreak{}weed\_\allowbreak{}segmentation (5, 13)\par
\hangindent=1em image\_\allowbreak{}safari\_\allowbreak{}plant\_\allowbreak{}part\_\allowbreak{}segmentation (17, 41)\par
\hangindent=1em ImageWeeds\_\allowbreak{}aerial\_\allowbreak{}weed\_\allowbreak{}detection (5, 12)\par
\hangindent=1em ImageWeeds\_\allowbreak{}weed\_\allowbreak{}detection (5, 20)\par
\hangindent=1em jujube\_\allowbreak{}disease\_\allowbreak{}spot\_\allowbreak{}detection (2, 5)\par
\hangindent=1em larch\_\allowbreak{}casebearer\_\allowbreak{}uav\_\allowbreak{}detection (5, 6)\par
\hangindent=1em litchi\_\allowbreak{}fruit\_\allowbreak{}detection (1, 10)\par
\hangindent=1em maize\_\allowbreak{}individual\_\allowbreak{}detection (2, 5)\par
\hangindent=1em maize\_\allowbreak{}seedling\_\allowbreak{}msdd\_\allowbreak{}detection (3, 10)\par
\hangindent=1em maize\_\allowbreak{}weed\_\allowbreak{}detection (2, 5)\par
\hangindent=1em mango\_\allowbreak{}detection\_\allowbreak{}australia (1, 5)\par
\hangindent=1em mango\_\allowbreak{}fruit\_\allowbreak{}stem\_\allowbreak{}detection (2, 5)\par
\hangindent=1em mango\_\allowbreak{}growth\_\allowbreak{}detection (4, 20)\par
\hangindent=1em Medjool\_\allowbreak{}date\_\allowbreak{}ripeness\_\allowbreak{}detection (2, 10)\par
\hangindent=1em MH\_\allowbreak{}Weed16\_\allowbreak{}weed\_\allowbreak{}detection (15, 36)\par
\hangindent=1em mtl\_\allowbreak{}plotcounter\_\allowbreak{}soybean\_\allowbreak{}detection (1, 5)\par
\hangindent=1em niwo\_\allowbreak{}tree\_\allowbreak{}crown\_\allowbreak{}detection (1, 5)\par
\hangindent=1em olive\_\allowbreak{}tree\_\allowbreak{}crown\_\allowbreak{}detection (0, 10)\par
\hangindent=1em onionfoliageset\_\allowbreak{}detection (3, 10)\par
\hangindent=1em OPPD\_\allowbreak{}plant\_\allowbreak{}detection (46, 235)\par
\hangindent=1em oyster\_\allowbreak{}mushroom\_\allowbreak{}maturity\_\allowbreak{}detection (2, 6)\par
\hangindent=1em paddy\_\allowbreak{}weed\_\allowbreak{}target\_\allowbreak{}detection (1, 5)\par
\hangindent=1em papaya\_\allowbreak{}leaf\_\allowbreak{}disease\_\allowbreak{}detection (5, 25)\par
\hangindent=1em plant\_\allowbreak{}doc\_\allowbreak{}detection (29, 137)\par
\hangindent=1em pomegranate\_\allowbreak{}growth\_\allowbreak{}detection (5, 18)\par
\hangindent=1em QuinceSet\_\allowbreak{}detection (2, 10)\par
\hangindent=1em raspberry\_\allowbreak{}maturity\_\allowbreak{}detection (5, 13)\par
\hangindent=1em rice\_\allowbreak{}panicle\_\allowbreak{}detection (1, 5)\par
\hangindent=1em RoCoLe\_\allowbreak{}disease\_\allowbreak{}detection (6, 30)\par
\hangindent=1em SDAU\_\allowbreak{}wheat\_\allowbreak{}head\_\allowbreak{}detection (1, 5)\par
\hangindent=1em sesame\_\allowbreak{}crop\_\allowbreak{}weed\_\allowbreak{}detection (2, 10)\par
\hangindent=1em Strawberry-DS\_\allowbreak{}strawberry\_\allowbreak{}detection (6, 12)\par
\hangindent=1em strawberry\_\allowbreak{}avocado\_\allowbreak{}ripeness\_\allowbreak{}detection (8, 38)\par
\hangindent=1em strawberry\_\allowbreak{}detection\_\allowbreak{}2022 (7, 8)\par
\hangindent=1em strawberry\_\allowbreak{}detection\_\allowbreak{}2023 (7, 6)\par
\hangindent=1em strawberry\_\allowbreak{}growth\_\allowbreak{}detection (7, 16)\par
\hangindent=1em sugar\_\allowbreak{}beet\_\allowbreak{}weed\_\allowbreak{}detection (2, 10)\par
\hangindent=1em sunflower\_\allowbreak{}detection (2, 5)\par
\hangindent=1em synthetic\_\allowbreak{}cowpea\_\allowbreak{}flower\_\allowbreak{}detection (1, 5)\par
\hangindent=1em synthetic\_\allowbreak{}cowpea\_\allowbreak{}pod\_\allowbreak{}detection (1, 5)\par
\hangindent=1em tea\_\allowbreak{}leaf\_\allowbreak{}disease\_\allowbreak{}pest\_\allowbreak{}detection (8, 39)\par
\hangindent=1em TealeafAgeQuality\_\allowbreak{}detection (4, 25)\par
\hangindent=1em three\_\allowbreak{}season\_\allowbreak{}weed\_\allowbreak{}detection (10, 34)\par
\hangindent=1em tomato\_\allowbreak{}adaptive\_\allowbreak{}yolov8\_\allowbreak{}detection (3, 8)\par
\hangindent=1em tomato\_\allowbreak{}detection\_\allowbreak{}kaggle (1, 5)\par
\hangindent=1em tomato\_\allowbreak{}factory\_\allowbreak{}detection (2, 5)\par
\hangindent=1em tomato\_\allowbreak{}leaf\_\allowbreak{}disease\_\allowbreak{}detection (7, 28)\par
\hangindent=1em tomato\_\allowbreak{}ripeness\_\allowbreak{}detection (6, 13)\par
\hangindent=1em UAV\_\allowbreak{}weed\_\allowbreak{}plant\_\allowbreak{}detection (2, 13)\par
\hangindent=1em USU-Corn-WeedDB\_\allowbreak{}detection (3, 11)\par
\hangindent=1em vegetable\_\allowbreak{}crop\_\allowbreak{}early\_\allowbreak{}detection (6, 20)\par
\hangindent=1em weed\_\allowbreak{}crop\_\allowbreak{}detection (13, 37)\par
\hangindent=1em wGrapeUNIPD-DL\_\allowbreak{}white\_\allowbreak{}grape\_\allowbreak{}bunch\_\allowbreak{}detection (0, 5)\par
\hangindent=1em Wheat\_\allowbreak{}Coccinellid\_\allowbreak{}detection (1, 10)\par
\hangindent=1em wheat\_\allowbreak{}disease\_\allowbreak{}detection (7, 33)\par
\hangindent=1em wheat\_\allowbreak{}head\_\allowbreak{}counting (1, 10)\par
\end{multicols}

\subsection*{Segmentation (34 datasets; classes, images)}
\begin{multicols}{2}\raggedright
\hangindent=1em apple\_\allowbreak{}flower\_\allowbreak{}segmentation (1, 5)\par
\hangindent=1em apple\_\allowbreak{}segmentation\_\allowbreak{}minnesota (1, 5)\par
\hangindent=1em AppleTreeDataset\_\allowbreak{}intertwined\_\allowbreak{}segmentation (1, 5)\par
\hangindent=1em barley\_\allowbreak{}disease\_\allowbreak{}segmentation (1, 11)\par
\hangindent=1em bean\_\allowbreak{}synthetic\_\allowbreak{}earlygrowth\_\allowbreak{}aerial (2, 10)\par
\hangindent=1em cabbage\_\allowbreak{}instance\_\allowbreak{}segmentation (1, 5)\par
\hangindent=1em carrot\_\allowbreak{}weed\_\allowbreak{}segmentation (2, 5)\par
\hangindent=1em carrot\_\allowbreak{}weeds\_\allowbreak{}germany (2, 5)\par
\hangindent=1em chicory\_\allowbreak{}root\_\allowbreak{}segmentation (1, 8)\par
\hangindent=1em CoFly-Weed-DB\_\allowbreak{}weed\_\allowbreak{}segmentation (1, 10)\par
\hangindent=1em CWF788\_\allowbreak{}weed\_\allowbreak{}segmentation (1, 5)\par
\hangindent=1em DeepRootLab\_\allowbreak{}root\_\allowbreak{}segmentation (1, 7)\par
\hangindent=1em fruitseg30\_\allowbreak{}segmentation (1, 5)\par
\hangindent=1em gran\_\allowbreak{}canaria\_\allowbreak{}vegetation\_\allowbreak{}segmentation (12, 30)\par
\hangindent=1em greenhouse\_\allowbreak{}leafy\_\allowbreak{}segmentation (1, 19)\par
\hangindent=1em LSID\_\allowbreak{}bean\_\allowbreak{}segmentation (1, 5)\par
\hangindent=1em merlot\_\allowbreak{}mildew\_\allowbreak{}segmentation (1, 5)\par
\hangindent=1em OccPepSeg\_\allowbreak{}YOLO\_\allowbreak{}occluded\_\allowbreak{}segmentation (1, 5)\par
\hangindent=1em onion\_\allowbreak{}weed\_\allowbreak{}segmentation (2, 5)\par
\hangindent=1em OpenEar\_\allowbreak{}base\_\allowbreak{}segmentation (1, 5)\par
\hangindent=1em OpenEar\_\allowbreak{}Projection\_\allowbreak{}segmentation (1, 5)\par
\hangindent=1em pdsnet\_\allowbreak{}field\_\allowbreak{}plant\_\allowbreak{}disease\_\allowbreak{}segmentation (2, 5)\par
\hangindent=1em PhenoBench\_\allowbreak{}segmentation (2, 5)\par
\hangindent=1em pine\_\allowbreak{}wilt\_\allowbreak{}segmentation (1, 10)\par
\hangindent=1em red\_\allowbreak{}grapes\_\allowbreak{}and\_\allowbreak{}leaves\_\allowbreak{}segmentation (2, 5)\par
\hangindent=1em Sesame\_\allowbreak{}Aerial\_\allowbreak{}weed\_\allowbreak{}segmentation (2, 5)\par
\hangindent=1em sorghum\_\allowbreak{}weed\_\allowbreak{}segmentation (3, 6)\par
\hangindent=1em spittlebug\_\allowbreak{}segmentation (1, 10)\par
\hangindent=1em strawberry\_\allowbreak{}runner\_\allowbreak{}segmentation (1, 5)\par
\hangindent=1em UAV\_\allowbreak{}SaSeg\_\allowbreak{}spartina\_\allowbreak{}segmentation (1, 10)\par
\hangindent=1em vineyard\_\allowbreak{}grape\_\allowbreak{}segmentation (1, 5)\par
\hangindent=1em vineyard\_\allowbreak{}pruning\_\allowbreak{}segmentation (3, 7)\par
\hangindent=1em weed\_\allowbreak{}segmentation\_\allowbreak{}france (2, 5)\par
\hangindent=1em white\_\allowbreak{}grapes\_\allowbreak{}and\_\allowbreak{}leaves\_\allowbreak{}segmentation (2, 5)\par
\end{multicols}

\subsection*{Visual question answering (13 datasets; questions, images)}
\begin{multicols}{2}\raggedright
\hangindent=1em Agri-LLaVA (250, 250)\par
\hangindent=1em AgriBench (248, 248)\par
\hangindent=1em AgroBench (245, 245)\par
\hangindent=1em AgroCoT (208, 349)\par
\hangindent=1em AgroMind (234, 324)\par
\hangindent=1em AgroOmni (250, 318)\par
\hangindent=1em CDDM (250, 250)\par
\hangindent=1em LeafNet (250, 250)\par
\hangindent=1em MIRAGE (183, 379)\par
\hangindent=1em PlantExpertVQA (234, 234)\par
\hangindent=1em PlantMicro (250, 250)\par
\hangindent=1em TomaMMU (245, 245)\par
\hangindent=1em WisWheat (250, 250)\par
\end{multicols}

}

\end{document}